\documentclass[conference]{IEEEtran}
\IEEEoverridecommandlockouts
\usepackage{cite}
\usepackage{amsmath,amssymb,amsfonts}
\usepackage{algorithm}
\usepackage{algpseudocode}
\usepackage{graphicx}
\usepackage{textcomp}
\usepackage{xcolor}
\usepackage{comment}
\usepackage{hyperref}
\usepackage{subcaption} 
\usepackage{multirow}
\usepackage{enumitem}
\usepackage{tabularx}
\usepackage{array}
\usepackage{epstopdf}
\usepackage{pdfpages}
\usepackage{tikz}
\newcommand\copyrighttext{%
  \footnotesize This work has been accepted for publication in the Proceedings of the IEEE International Conference on Data Engineering (ICDE) 2025. © 2025 IEEE. Personal use of this material is permitted. Permission from IEEE must be obtained for all other uses, in any current or future media, including reprinting/republishing this material for advertising or promotional purposes, creating new collective works, for resale or redistribution to servers or lists, or reuse of any copyrighted component of this work in other works.}
\newcommand\copyrightnotice{%
\begin{tikzpicture}[remember picture,overlay]
\node[anchor=south,yshift=10pt] at (current page.south) {\fbox{\parbox{\dimexpr\textwidth-\fboxsep-\fboxrule\relax}{\copyrighttext}}};
\end{tikzpicture}%
}

\begin{document}

\title{Path-based summary explanations for graph recommenders}

\author{\IEEEauthorblockN{Danae Pla Karidi}
\IEEEauthorblockA{\textit{Archimedes/Athena RC, Greece} \\
\textit{danae@athenarc.gr}
\and
\IEEEauthorblockN{Evaggelia Pitoura}
\IEEEauthorblockA{\textit{University of Ioannina and Archimedes/Athena RC, Greece}}
\textit{pitoura@uoi.gr}
}}

\maketitle

\begin{abstract}
Path-based explanations provide intrinsic insights into graph-based recommendation models. However, most previous work has focused on explaining an individual recommendation of an item to a user. In this paper, we propose summary explanations, i.e., explanations that highlight why a user or a group of users receive a set of item recommendations and why an item, or a group of items, is recommended to a set of users as an effective means to provide insights into the collective behavior of the recommender. We also present a novel method to summarize explanations using efficient graph algorithms, specifically the Steiner Tree and the Prize-Collecting Steiner Tree. Our approach reduces the size and complexity of summary explanations while preserving essential information, making explanations more comprehensible for users and more useful to model developers. Evaluations across multiple metrics demonstrate that our summaries outperform baseline explanation methods in most scenarios, in a variety of quality aspects.
\end{abstract}


\section{Introduction}
The interpretability and transparency of machine learning models have emerged as critical areas of research, particularly within the domain of explainable AI \cite{10.1145/3514221.3522564, electronics8080832}.
Although substantial research has been done on explaining recommendations \cite{zhang2020explainable}, most of this research focuses on explaining individual recommendations, hence explanations for each specific recommendation made to a user.  However, users often seek a broader understanding of how the entire system functions towards them \cite{tintarev2012beyond}, while item providers want to comprehend the overall system behavior regarding the items they offer \cite{ricci2021recommender, deldjoo2024fairness}.
For example, imagine a music streaming user 
who receives several recommendations for songs.
Explanations relate the recommended songs to several songs, artists and genres that the user has enjoyed in the past. The user would like a single explanation highlighting the most common reasons for the recommendations received.
To address these needs, we propose summary explanations. 

We introduce two types: user-based summary explanations, which explain the overall behavior of the system to a specific user —explaining why they receive specific item recommendations —and item-based summary recommendations, which explain the global behavior of the system for a specific item based on the users to whom the item is recommended.
By summarizing multiple explanations, users can determine the most relevant aspects and the overall rationale behind the recommendations they receive, avoiding information overload \cite{ghazimatin2020prince} and enabling them to make more informed decisions. 
Item providers can identify key features that appeal to users or require improvement, while model developers can detect underlying patterns in model behavior, facilitating targeted and effective system enhancements. 

In addition, we extend our summary recommendations to groups of users and items.
Analyzing group-based summaries helps identify potential biases affecting specific user groups or item categories. These summaries are also useful for targeted marketing and advertising by enabling marketers to design more effective, personalized strategies, while item providers can tailor their advertising based on insights into the interests of specific user groups. 

We focus on four key summarization scenarios: \textit{user-centric} summarizes the explanations to a user, \textit{item-centric} summarizes the explanations to various users for a specific item, \textit{user-group} summarizes the explanations for a group of users, and \textit{item-group} summarizes the explanations to various users for a group of recommended items. 

Moreover, we introduce efficient algorithms for computing summary explanations tailored for graph-based recommenders, which leverage graph structures to capture complex relationships between users, items, and their interactions, enhancing recommendation accuracy \cite{wu2022graph, wang2020recommendation}. 
Graph-based recommendation systems are used in major platforms. Twitter recommends connections between users and content based on shared interests and activities \cite{sharma2016graphjet}, while Spotify uses a co-listening graph linking podcasts and audiobooks \cite{de2024personalized}. Amazon combines co-purchase and co-view data to build a product recommendation graph \cite{amazon}. Similarly, Facebook \cite{ugander2011anatomy} and Pinterest \cite{ying2018graph} incorporate graph algorithms to personalize content recommendations. A common method for explaining graph-based recommendations is path-based explanations, which trace the paths from users to recommended items using the graph structure \cite{geng2022path}. Path-based approaches (also called meta-path or path-embedding techniques), such as PGPR \cite{xian2019reinforcement} and CAFE \cite{xian2020cafe}, are considered state-of-the-art for both recommendation accuracy and explanation quality \cite{guo2020survey, gao2023survey}. Recent models PLM \cite{plm}, and PEARLM \cite{balloccu2023faithful} enhance these explanations by using pre-trained language models.

\copyrightnotice
We propose a novel approach based on the Steiner Tree problem, aggregating individual path-based explanations into a minimum edge tree that connects recommended items for a user while considering individual explanation paths. Additionally, we introduce a Prize-Collecting Steiner Tree variation to balance summary size and comprehensiveness when summarizing many explanation paths.

Finally, we present experimental results using a variety of metrics to evaluate different aspects of the quality of our summary explanations. We use several recommendation methods for the individual explanation paths: PGPR \cite{xian2019reinforcement}, CAFE \cite{xian2020cafe},  PLM-Rec \cite{plm} and PEARLM \cite{balloccu2023faithful} and evaluate our method using two publicly available datasets: the ML1M  \cite{MovieLensData} and the LFM1M \cite{lfm1b}
dataset. Results show that the Steiner summarization method generally outperforms the baselines in metrics like comprehensibility, actionability, redundancy, and relevance, while the PCST variation performs better in metrics like diversity, privacy, and performance scalability.

The main novelty of our work is that,
to the best of our knowledge, we are
the first to introduce summary recommendations of path-based explanations for users, items, groups of users, and groups of items. We also propose algorithms for summary recommendations based on the well-known Steiner Tree and Prize-Collecting Steiner Tree (PCST)
algorithms. In a nutshell, our contributions are as follows:
\begin{itemize}
\item We introduce summary explanations of different granularities, including individual users, individual items, groups of users, and groups of items.
\item We propose efficient algorithms for computing summary explanations for graph-based recommenders.
\item We demonstrate that summary explanations outperform baseline path-based explanation methods across multiple evaluation metrics.
\end{itemize}

In the remainder of this paper, we review related works in Section \ref{related_works}. We outline the problem and elaborate on the formal problem definition in Section \ref{problem}. Our methodology, detailing the summarization algorithms, is presented in Section \ref{steiner}. Section \ref{experiment}
covers the experimental evaluation. Finally, we conclude with a summary of our contributions and future research directions in Section \ref{conclusion}.

\section{Related Works}
\label{related_works}

Previous research has not specifically addressed the task of explanation summarization in recommendation systems. Our work fills this gap by contributing to the broader field of explainable recommendations, with a particular focus on intrinsic graph-based explainable recommendations. Additionally, our research intersects with techniques in graph summarization.

Explainable recommenders aim to explain why items are recommended to users. Recent machine learning advancements have heightened the focus on explainability \cite{zhang2020explainable}, leading to new models such as Latent Factor Models (LFMs) \cite{tao2019}, representation learning methods \cite{wu2019}, and generative models using large language models (LLMs) for personalized textual explanations \cite{talrec23, gpt23}. 
Explainable models can be classified into interpretable models, with transparent decision mechanisms \cite{wang2023sequential}, and black-box models, which generate explanations post-recommendation \cite{wang2022tower}. Intrinsic explanations \cite{bodria2023benchmarking, adadi2018peeking} applied on interpretable models \cite{bodria2023benchmarking, adadi2018peeking} provide insights directly from the recommendation pipeline itself, aiding user comprehension.

\textbf{\textit{Graph-Based Explanations.}}
Our work focuses on graph-based explanations, leveraging knowledge-enhanced graphs where nodes represent users and items, and edges represent interactions. These graphs inject semantic content and extend the graph beyond a bipartite structure, offering efficient recommendations and meaningful explanations \cite{guo2020survey}. Graph-based explanations can be black-box, using embeddings for historical interaction representations \cite{wu2023generic, wang2018ripplenet, 10.1109/TKDE.2018.2833443, 10.1145/3219819.3219965, 10.1609/aaai.v33i01.33015329}, or intrinsic, using graph reasoning and link prediction \cite{seq2021}. 
Our method summarizes explanation paths over knowledge-enhanced graphs and is suitable for intrinsic models that naturally provide explanation paths. Importantly, our approach is compatible with any recommendation method that outputs explanation paths, regardless of the specific reasoning algorithms it uses. Additionally, for methods that do not output paths but provide recommended items and access to underlying graph data, our approach can generate new path explanations based on the graph structure. We benchmark against several intrinsic path-based methods: PGPR \cite{xian2019reinforcement}, which uses reinforcement learning for path reasoning, and CAFE \cite{xian2020cafe}, which uses historical patterns and adversarial training to guide path-finding. Additionally, we provide experimental results against PLM-Rec \cite{plm} and PEARLM \cite{balloccu2023faithful,balloccu2022recency}. Both methods employ language models over knowledge graphs to generate explainable recommendations. PLM-Rec generates novel paths beyond the static KG topology, while PEARLM improves upon this by ensuring that generated paths faithfully adhere to valid KG connections. 

\textbf{\textit{Graph Summarization Techniques.}}
Summarizing explanation paths can be viewed through the lens of the general problem of graph summarization \cite{liu2018graph}, which often involves clustering nodes into super-nodes \cite{lefevre2010grass, toivonen2011compression} or merging edges \cite{maccioni2016scalable}.  
Other graph-summarization methods involve selecting a subset of “important” nodes or edges, resulting in a sparsified graph \cite{shen2006visual, mathioudakis2011sparsification}. For instance, some of the work in \cite{steiner_sum} creates a summarization by identifying the most important nodes using centrality measures and then finding the Steiner Tree that connects them \cite{steiner_sum}.

Existing graph summarization techniques 
are primarily designed for general applications and focus on reducing graph complexity and achieving storage efficiency. However, in the context of recommendation systems, these methods often lead to the loss of crucial local structures and detailed information that are vital for understanding user-item interactions. Moreover, node and edge aggregation methods prioritize overall structure over detailed interactions. Hence, they can obscure the specific paths and relationships that explain why certain items are recommended to users. Additionally, the optimization functions used are application-dependent, limiting their generalizability across different types of graphs. 
Our work, instead, focuses on summarizing recommendation paths by directly including recommended items as terminal nodes and leveraging information from individual explanation paths.  We apply a prize-collecting Steiner Tree method to optimize the inclusion of key nodes and maintain computational efficiency, particularly for larger graphs. This approach ensures that recommendation-relevant paths are prioritized in the summary.

\textbf{\textit{Group Recommendations and Summaries.}}
Another line of research considers group recommendations \cite{group_recs, fair_recs_18, stratigi2017fairness, guo2020group, deng2021knowledge}, which focus on recommending items to groups by aggregating individual preferences into a unified group profile. This approach aims to balance the various preferences within the group to generate satisfying recommendations for all members. In contrast, our work on group summaries tackles a different problem. Instead of creating a group profile, we summarize explanation paths for recommendations given to individual users within a group.
While group recommendation systems unify preferences, our group summaries explain how individual recommendations collectively form a cohesive understanding for the group. 

\section{Problem Definition}
\label{problem}

\label{sec:pre}
Let \( U = \{ u_1, u_2, \ldots, u_n \} \) be a set of users and \( I = \{ i_1, i_2, \ldots, i_m \} \) be a set of items. The \( n \times m \) rating matrix \( M \) is defined such that \( M[u, i] = (r, t) \), where \( r \) is the positive rating and \( t \) is the timestamp of the rating, if \( u \in U \) has rated \( i \in I \), and \( M[u, i] = (0, 0) \) otherwise (indicating no rating). As many graph recommenders \cite{balloccu2023reinforcement, xian2019reinforcement, xian2020cafe}, we construct from \(  M \), a directed weighted graph \( G_M(V_M, E_M, w_M) \), where \( V_M = U \cup I \) is the set of nodes corresponding to users and items. \( E_M \subseteq U \times I \), meaning there is an edge from \( u \) to \( i \) if \( M[u, i] \neq (0, 0) \) \cite{balloccu2023reinforcement, xian2019reinforcement, xian2020cafe}.

In recommendation systems, it is essential to consider both the history and recency of user-item interactions. Older interactions may be less indicative of current preferences, while higher ratings reflect stronger user preferences. To address this, the weight function \( w_M \) is defined on \( E_M \) and maps to \( \mathbb{R} \), combining the rating \( r \) and the timestamp \( t \) from the rating matrix \( M \). For an edge \( (u, i) \in E_M \), where \( M[u, i] = (r, t) \): \( w_M(u, i) = \beta_1 \cdot r + \beta_2 \cdot f(t) \). Here, \( \beta_1 \) adjusts the importance of rating, \( \beta_2 \) adjusts the importance of recency, and \( f(t_{ui}) \) is the recency function that assigns higher values to more recent interactions. The dependency on the rating ensures that interactions with higher ratings have a greater impact on the summarization process. The recency function is defined as: \( f(t_{ui}) = e^{-\gamma \cdot (t_0 - t_{ui})} \), where \( t_0 \) is the current time, and \( \gamma \) is a decay parameter controlling how quickly the weight decreases with time. The exponential decay function is chosen to reflect the natural diminishing influence of older interactions. This combined use of recency and rating scores ensures that the model prioritizes recent and highly rated interactions, which are more likely to align with the user’s current preferences.

We extend \( G_M \) with a set of external nodes \( V_A \) that represent additional information such as item categories, user attributes, etc., and edges \( E_A  \subseteq V_M \times V_A \) that connect users and items with the external nodes. The weight of these edges is defined by \( w_A \), a weight function \( E_A \to \mathbb{R} \). The weight \( w_A(v, a) \) represents the relevance score of the external node \( a \in V_A\) to the user or item \( v \in V\).

The extended graph \( G(V, E, w) \) of \( G_M(V_M, E_M, w_M) \), called the knowledge-based graph, is a directed weighted graph where \( V = V_M \cup V_A \), \( E = E_M \cup E_A \), and a weight function \( w: E \to \mathbb{R} \), where \( w(e) = w_M(e) \) if \( e \in E_M \), and \( w(e) = w_A(e) \) if \( e \in E_A \).

The knowledge-based graph is used to generate recommendations, where a recommendation is a set of items for each user and the explanation is a path \cite{xian2019reinforcement}. This means that the explanation \( E(u, i) \) for the recommendation of item \( i \) to user \( u \) is a path, starting from a user node \( u \in U \) and ending at an item node \( i \in R_u \): \( E(u, i) = (u, v_1, v_2, \ldots, v_k, i) \), where \( u, v_1, v_2, \ldots, v_k, i \in V \) and \( (u, v_1), (v_1, v_2), \ldots, (v_k, i) \in E \).

We use \( R_u \) to denote the set of recommended items to user \( u \):
\( R_u = \{ i \mid i \in I \text{ and } i \text{ is recommended to } u \}.\) We use \( E_u \) to denote the corresponding explanation paths leading from \( u \) to the items in \( R_u \): \( E_u = \{ E(u, i) \mid i \in R_u \}\). 

We use \( C_i \) to denote the set of users to whom the item \( i \) is recommended:
\( C_i = \{ u \mid u \in U \text{ and } i \in R_u \}\). Analogously, we use \( E_i \) to denote the corresponding explanation paths leading from users in \( C_i  \) to item \( i \): \(E_i = \{ E(u, i) \mid u \in C_i \}\).

In this paper, we propose explanations for a set of recommendations as opposed to individual recommendations. While individual explanations are paths of \( G \), summary explanations are subgraphs of \( G \). We start by defining summary explanations for users and items and then generalize our definitions for groups of users and items.

\subsubsection{User-centric and item-centric summary explanations}

Summary user-centric explanations summarize the explanation paths leading from a user to their recommended items. An aggregated view of the individual recommendation paths enables users to determine more effectively the key aspects of the recommendations they get. 

The straightforward definition of summary user-centric explanation is as a union graph that includes the explanation paths \( E_u \) of the items in \( R_u \). However, this simplistic consolidation can lead to information overload, obscuring the overarching rationale and systems behavior towards the user. Users may be overwhelmed by the volume of information and struggle to discern the key reasons behind their recommendations. For explanations to be interpretable, they must be concise and manageable in size \cite{Zhou2021EvaluatingTQ} to highlight the most relevant paths and connections and enhance user comprehension.  

\begin{figure}[ht]
    \centering
    \includegraphics[width=0.5\textwidth]{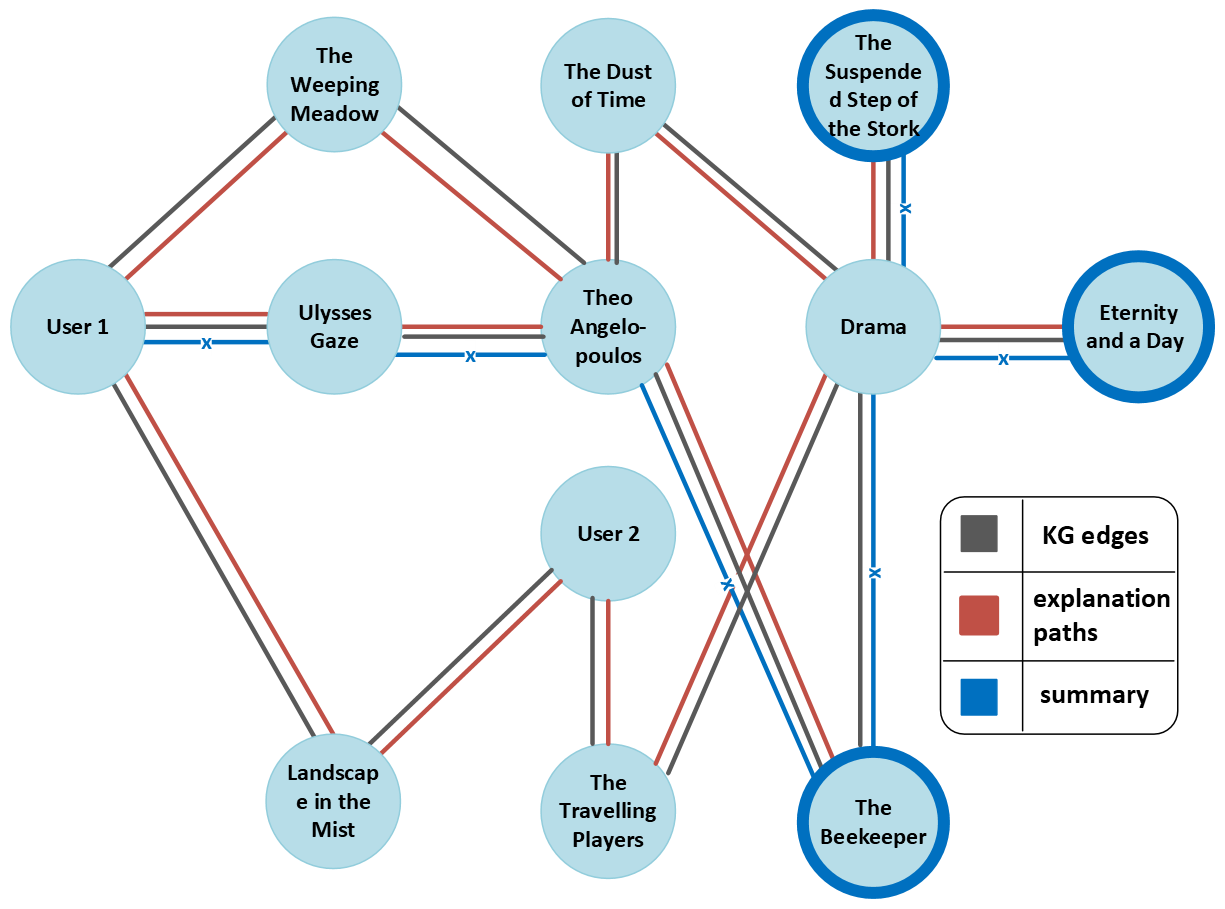}
    \caption{Summarized explanation}
    \label{fig:steiner_tree}
\end{figure}

For example, consider User 1, who receives recommendations for the movies \textit{Eternity and a Day} (Item A), \textit{The Beekeeper} (Item B), and \textit{The Suspended Step of the Stork} (Item C). Table \ref{t:ex} shows the explanation paths and their summarization, while Figure \ref{fig:steiner_tree} depicts the individual explanation paths (red edges) and the summarized explanation (green edges) on the knowledge graph (grey edges). While the original explanations had a total length of 13, the summarization achieves a length of 6 edges, significantly increasing the comprehensibility of the explanation. In addition, before the summarization, User 1 needed to understand how she connected to each movie through multiple separate and convoluted paths. Specifically, each individual path includes intermediary steps that, while part of the explanation, may not all be crucial for understanding the main reasons behind the recommendations. As shown in the example, in \(P_{1,A}\), steps like ``Landscape in the Mist'' and ``User 2'' may not be crucial for understanding the main reason ``User 1'' is connected to ``Eternity and a Day''. In \(P_{1,B}\), steps like ``Ulysses' Gaze'' and ``Theo Angelopoulos'' are necessary, since they are repeated in other paths, indicating their central role, while in \(P_{1,C}\), steps like ``The Weeping Meadow'' and ``The Dust of Time'' add to the clutter without significantly enhancing the understanding of the main connection. By focusing on the repeated and central nodes, the summary path eliminates redundant and less relevant steps: ``Drama'' and ``Theo Angelopoulos'' are key nodes that appear in multiple paths and are thus central to understanding the recommendations. Hence, by summarizing, we not only reduce the total explanation length from 13 to 6 edges, but we also retain only the most crucial connections (``Theo Angelopoulos'', and ``Drama''), which are directly related to the recommendations. 
Though not shown in this example, users often encounter repeated edges across explanation paths, making summaries even more effective.

\begin{table}[ht]
\centering
\caption{Summarized explanation paths for User 1}
\footnotesize 
\renewcommand{\arraystretch}{1.3} 
\begin{tabularx}{\columnwidth}{|p{1cm}|X|}
\hline
\textbf{} & \textbf{Explanation} \\
\hline
\textbf{Item A} &
\(P_{1,A} = (\text{User 1}, \text{Landscape in the Mist}, \text{User 2}, \) \newline
\(\text{The Travelling Players}, \text{Drama}, \text{Eternity and a Day})\) \newline
\textit{``User 1 is connected to Eternity and a Day through Landscape in the Mist, User 2, The Travelling Players, and Drama.''}
\\
\hline
\textbf{Item B} &
\(P_{1,B} = (\text{User 1}, \text{Ulysses Gaze}, \text{Theo Angelopoulos},\) \newline
\(\text{The Beekeeper})\) \newline
\textit{``User 1 is connected to The Beekeeper through Ulysses' Gaze and Theo Angelopoulos.''}
\\
\hline
\textbf{Item C} &
\(P_{1,C} = (\text{User 1}, \text{The Weeping Meadow}, \) \newline
\(\text{Theo Angelopoulos}, \text{The Dust of Time}, \text{Drama}, \) \newline
\(\text{The Suspended Step of the Stork})\) \newline
\textit{``User 1 is connected to The Suspended Step of the Stork through The Weeping Meadow, Theo Angelopoulos, The Dust of Time, and Drama.''}
\\
\hline
\textbf{Summary} &
\( V_S = (\text{User 1}, \text{Ulysses Gaze}, \text{Theo Angelopoulos},\) \newline
\(\text{The Beekeeper}, \text{Drama}, \text{Eternity and a Day}, \) \newline
\(\text{The Suspended Step of the Stork})\) \newline
\textit{``User 1 is connected to Eternity and a Day, The Beekeeper, and The Suspended Step of the Stork through Ulysses' Gaze, Theo Angelopoulos, and Drama.''}
\\
\hline
\end{tabularx}
\label{t:ex}
\end{table}

\begin{description}
    \item[\textbf{Problem [User-Centric Summary]:}]
    Given a knowledge-based graph \( G = (V, E, w) \), a set of items \( R_u \), and a set of explanation paths \( E_u \) the problem definition of user-centric summary \( S \) is as follows:
    
    \textbf{Find}: a weakly connected subgraph \( S = (V_S, E_S, w) \) of \( G \) such that:
    \begin{itemize}[label={-}]
        \item \( R_u \subseteq V_S \)
        \item \( |E_S| \) is minimized
        \item \( \sum_{e \in E_S} w(e) \) is maximized
    \end{itemize}
\end{description}

 Summary item-centric explanations summarize the explanation paths leading from various users to a specific recommended item. A consolidated view of the individual recommendation paths \( E_i \) can help item providers understand why the model recommends an item, understand the collective reasons behind the item's recommendations, and what the key features appeal to users or features that need improvement. The problem definition for the item-centric summary is the same as the user-centric summary, with the difference that we focus on items recommended to users. 

\begin{description}

    \item[\textbf{Problem [Item-Centric Summary]:}]
    Given a knowledge-based graph \( G \), a set of users \( C_i \), and a set of explanation paths \( E_i \):
    
    \textbf{Find} a weakly connected subgraph \( S = (V_S, E_S, w) \) of \( G \) such that:
    \begin{itemize}[label={-}]
        \item \( C_i \subseteq V_S \)
        \item \( |E_S| \) is minimized
        \item \( \sum_{e \in E_S} w(e) \) is maximized
    \end{itemize}
\end{description}

\subsubsection{Summary Explanations for Groups}
Given a group of users \( D \subseteq U \), let \( R_D \) be a set of items recommended to users in \( D\) and \(E_D\) the corresponding explanation paths: \(E_D = \{ E(u, i) \mid u \in D, i \in R_D \}\). Summary user-group explanations summarize the explanation paths leading from a group of users to their recommended items. The problem definition for user-group summary is the same as the user-centric summary with the difference that we now focus on a group of users: the nodes that must be included in the subgraph are \( R_D \) instead of \( R_u \) and the explanation paths are \(E_D\) instead of \(E_u\).

An aggregated view of the individual recommendation paths of users can help in understanding the overall recommendation behavior towards the group. These summaries apply to any group of users, whether defined manually (for example, based on demographics) or identified through machine learning techniques (for example, by clustering behavioral patterns).  Using user group summaries, model developers detect underlying regularities in model behavior and performance and identify potential model biases that may affect specific user groups. Moreover, marketers can use them to tailor group-specific marketing strategies.

Analogously, given a group of items \( F \subseteq I \), let \( C_F \) be a set of users to whom the items in \( F \) were recommended and \(E_F\) the corresponding explanation paths: \(E_F = \{ E(u, i) \mid u \in C_F, i \in F \}
\). Summary item-group explanations summarize the explanation paths leading from various users to a group of items. The item-group summary problem definition is the same as the item-centric summary with the difference that the nodes that must be included in the subgraph are \( C_F \) instead of \( C_i \) and the explanation paths \( E_F\) instead of \(E_i\).

Item-group summaries can be generated for any collection of items, defined either manually for example based on specifications given by the item providers, or through automated grouping methods. This aggregated view of the individual recommendation paths to items can help item providers tailor their advertising efforts based on the most appealing aspects of their items. Moreover, model developers can spot biases that may affect specific item categories.

\section{Computing Summary Explanations}
\label{steiner}
In this section, we present efficient algorithms for computing summary explanations for graph-based recommendations. We employ two graph-based approaches: the Steiner Tree and the Prize Collecting Steiner Tree.

\subsection{Steiner-Tree (ST) Summary Explanations}
\label{sec:weights}
To solve the problem of summarizing explanation paths, we leverage the concept of the Steiner Tree. The Steiner Tree problem, in its classical form, seeks the shortest tree that spans a given set of terminal nodes in a graph, potentially including additional intermediate nodes (Steiner nodes) to minimize the total edge weight. However, our objective is to maximize the total weight of the subgraph while minimizing the number of edges, effectively balancing between weight maximization and edge minimization. To adapt the problem for the Steiner Tree solution, we align the edge weights with the Steiner Tree's minimization objective by multiplying all edge weights by -1, converting the problem into one that seeks to maximize the weights, consistent with the Steiner Tree approach. The set of terminal nodes for user-centric summary is \( T = u \cup R_u \), for item-centric summary \( T = i \cup C_i \), for user group summary \( T = D \cup R_D \), and for item-group summary  \( T = F \cup C_F \).
Algorithm \ref{al:steiner} outlines the steps for constructing a Steiner Tree to summarize explanation paths using a Minimum Spanning Tree approximation. The ST-based algorithm has time complexity \( O(|T| (|E| + |V| \log |V|)) \) under typical conditions, where \( T \) (terminals) is small relative to large \( V \) (nodes), and \( E \) (edges). Its approximation ratio to the optimal Steiner Tree solution is at most 2 \cite{zelikovsky1993faster}.

\begin{algorithm}
\small
\caption{ST-based summary-explanations}
\label{al:steiner}
\begin{algorithmic}[1]
\Require Graph $G = (V, E, w)$, set of terminal nodes $T \subseteq V$
\Ensure Steiner Tree $S = (V_S, E_S)$

\State Initialize $V_S \leftarrow T$, $E_S \leftarrow \emptyset$
\State Compute shortest paths between all pairs of terminal nodes in $T$ using edge weights $w(e)$
\State Construct a complete graph $G_c = (T, E_c, w_c)$ where each edge $(u, v) \in E_c$ represents the shortest path distance between $u$ and $v$ in $G$
\ForAll{edges $(u, v) \in E_c$}
    \State $w_c(u, v) \leftarrow$ sum of edge weights along the shortest path between $u$ and $v$ in $G$
\EndFor
\State Compute the Minimum Spanning Tree (MST) of $G_c$ using edge weights $w_c$, denoted as $MST_c$
\State Initialize $S \leftarrow MST_c$
\ForAll{edges $(u, v) \in MST_c$}
    \State Replace $(u, v)$ in $MST_c$ with the shortest path between $u$ and $v$ in $G$
    \ForAll{edges $(x, y)$ in the shortest path between $u$ and $v$}
        \State Add $(x, y)$ to $E_S$ and $x, y$ to $V_S$
    \EndFor
\EndFor
\State \textbf{return} $S = (V_S, E_S)$
\end{algorithmic}
\end{algorithm}

The initial weights of edges in the graph \(G_M\) are determined based on the user-item interaction matrix \(M\). While this captures the historical interactions between users and items, it does not account for the specific explanation paths that are generated for individual recommendations. The problem with this approach is that without adjusting the weights to reflect the importance of edges in the explanation paths, the summarization algorithms may not prioritize the most relevant paths. This could lead to summaries that merely take into account the individual explanations and create entirely new explanations instead of summarizing the individual ones. To address this, we propose a solution where the weight function \( w(e) \) increases the initial weights of edges of \(G_M\) based on their inclusion in the individual explanation paths. This adjustment is controlled by the parameter \(\lambda\), which allows us to balance the influence of the initial weights and the frequency of edges in the explanation paths. By incorporating both the initial user-item interaction matrix \(M\) and the presence of edges in the explanation paths, our weight function ensures that the summarization algorithms prioritize paths that are part of the individual explanation paths, effectively summarizing them.

The weight function \( w(e) \) is defined as:
\begin{align}
\label{eq:we}
w(e) = w_M(e) \left(1 + \lambda \cdot \frac{\sum_{x \in \mathcal{S}} \mathbf{1}_{e \in P}}{|\mathcal{S}|} \right)
\end{align}
where:
\begin{itemize}[label={-}]
    \item \( w_M(e) \) is the initial weight assigned to edge \( e \) in \( G_M \).
    \item \( P \) varies based on summarization type: user-centric: \( P = E_u \), item-centric: \( P = E_i \), user-group: \( P = E_D \), item-group: \( P = E_F \).
    \item \( \lambda \) is a positive scaling factor adjusting the impact of the input explanation path frequency to the summarization. If \(\lambda\) is set to zero, the summarization algorithm generates a new explanation, as the impact of the existing explanation paths is effectively nullified. Consequently, the algorithm creates a new explanation that includes \(S\).
    \item \( \mathbf{1}_{e \in P} \) is an indicator function that returns 1 if edge \( e \) is part of the explanation paths \( P \) and 0 otherwise.
    \item \(\ S \) varies based on summarization type: user-centric: \( S = R_u \), item-centric: \( S = C_i \), user-group: \( S = R_D \), item-group: \( S = C_F \).
\end{itemize}

\subsection{Prize-Collecting ST Summary Explanations}

To address the problem of a significant increase in the number of terminal nodes when creating group summaries, we leverage the concept of the Prize-Collecting Steiner Tree (PCST) to balance maximizing the total weight of the subgraph with minimizing the number of edges.

The PCST is a generalization of the classical Steiner tree problem. Given a weighted graph and a set of terminal nodes, the Steiner tree problem seeks to find a minimum-cost spanning tree of the terminal nodes. The prize-collecting variant relaxes the connectivity constraint by assigning a prize to each terminal node. Instead of connecting a terminal node to our spanning tree, we can choose to forego the prize of the omitted terminal. The PCST problem aims to find a subgraph that minimizes the total cost, which is the sum of the weights of the included edges minus the prizes of the included nodes. This formulation captures the trade-off between the costs of connecting a terminal and omitting it from the solution. Hence, in the context of group-centric summarization, this approach allows us to prioritize important nodes and edges while reducing the overall complexity of the explanation paths.

To this end, we normalize the edge weights \( w(e) \) to set the prizes \( \alpha \) and \( \beta \) as follows:
\(\alpha = \max(w(e)) \quad \text{and} \quad \beta = \min(w(e))\). Following, we assign node prizes \( p(v) \) such that \( p(v) = \alpha \) for nodes in the set of terminal nodes \( T \), and \( p(v) = \beta \) for nodes not in \( T \). 
High prizes for terminal nodes (\( \alpha \)) prioritize their inclusion, while low prizes for non-terminal nodes (\( \beta \)) discourage their inclusion unless they are essential for connecting terminals. This approach balances the inclusion of terminal nodes and edge costs, ensuring non-terminal nodes are added only if they significantly reduce connection costs.

The total cost to be minimized is: \(C(S) = \sum_{e \in E_S} w'(e) - \sum_{v \in V_S} p(v)\), where \( C(S) \) represents the total cost of the subgraph \( S \), \( w'(e) \) is the modified edge weight for edge \( e \in E_S \), and \( p(v) \) is the prize for node \( v \in V_S \).
Algorithm \ref{alg:pcst} outlines the steps for constructing the Prize-Collecting Steiner Tree to summarize explanation paths using a 2-approximation \cite{goemans1995general} and runs in \( O((|V| + |E|) \log |V|) \) time.

\begin{algorithm}
\small
\caption{PCST-based summary-explanations}
\label{alg:pcst}
\begin{algorithmic}[1]
\Require Graph \( G = (V, E, w) \), node prizes \( p(v) \)
\Ensure Prize-Collecting Steiner Tree \( S = (V_S, E_S) \)

\State Initialize \( V_S \leftarrow \emptyset \), \( E_S \leftarrow \emptyset \)
\State Initialize priority queue \( Q \) and disjoint set \( D \)
\State Initialize \( total\_cost \leftarrow 0 \)

\ForAll{nodes \( v \in V \)}
    \State \( D.make\_set(v) \)
    \State Insert \( v \) into \( Q \) with priority \( -p(v) \)
\EndFor

\While{\( Q \) is not empty}
    \State Extract node \( u \) with highest priority from \( Q \)
    \State \( total\_cost \leftarrow total\_cost - p(u) \)
    \State \( V_S \leftarrow V_S \cup \{u\} \)
    \ForAll{edges \( (u, v) \in E \)}
        \State \( u\_set \leftarrow D.find(u) \)
        \State \( v\_set \leftarrow D.find(v) \)
        \If{\( u\_set \neq v\_set \)}
            \State \( cost \leftarrow w(u, v) \)
            \If{\( cost < Q[v] \)}
                \State \( Q[v] \leftarrow cost \)
                \State \( D.union(u, v) \)
                \State \( E_S \leftarrow E_S \cup \{(u, v)\} \)
                \State \( total\_cost \leftarrow total\_cost + w(u, v) \)
            \EndIf
        \EndIf
    \EndFor
\EndWhile

\State \textbf{return} \( S = (V_S, E_S) \)
\end{algorithmic}
\end{algorithm}

\section{Experimental evaluation}
\label{experiment}
In this section, we present experimental results that evaluate the quality of our summaries and the performance of our algorithms.
We primarily use the MovieLens 1M (ML1M) dataset \cite{MovieLensData} and the DBpedia Knowledge Graph (KG) \cite{cao2018unifying}. 
To enrich ML1M, we extract contextual information from DBpedia dumps \cite{Zhao-DI-2019}, integrating attributes such as director, actors, genre, composers, and other relevant properties for movies. We construct the knowledge-based graph by integrating user-item interactions from ML1M with DBpedia. In these graphs, nodes represent users, movies, and DBpedia entities, while edges connect users to movies based on interactions, with weights indicating user ratings (see Section \ref{sec:weights}). 
Table \ref{tab:knowledge_graph_properties} 
details the properties of the ML1M graph.
We also present experiments using the LFM1M dataset \cite{lfm1b} enriched with relevant information about songs and their properties from
DBpedia \cite{Zhao-DI-2019} at the end of this section.

\begin{table}[ht]
    \centering
    \caption{ML1M Knowledge-Based Graph Statistics}
    \Large
    \resizebox{0.48\textwidth}{!}{%
    \begin{tabular}{|l|c|c|c|c|}
        \hline
        \textbf{Property} & \textbf{User} & \textbf{Item} & \textbf{External Knowledge} & \textbf{Total} \\ \hline
        Number of nodes & 6,040 & 3,883 & 10,820 & 19,844 \\ \hline
        Number of edges & \begin{tabular}[c]{@{}c@{}}932,293 \\ (to items)\end{tabular} & \begin{tabular}[c]{@{}c@{}}178,461 \\ (to external)\end{tabular} & - & 1,125,631 \\ \hline
        Average Degree & 154.35 & \begin{tabular}[c]{@{}c@{}}240.10 \\ (from users) \\ 45.96 (to external)\end{tabular} & 17.99 & 113.45 \\ \hline
        Density & \multicolumn{4}{c|}{0.0057} \\ \hline
        Average Path Length & \multicolumn{4}{c|}{3.20} \\ \hline
        Diameter & \multicolumn{4}{c|}{6.0} \\ \hline
    \end{tabular}
    }
    \label{tab:knowledge_graph_properties}
\end{table}

\subsection{Experiment setup}
\label{setup}
Using the Steiner and PCST algorithms we created summaries for each of the following scenarios: user-centric, item-centric, user-group, and item-group summaries.

\textit{\textbf{User and Item Sampling.}} For user-centric summarization, we selected 100 male and 100 female users, preserving the original rating distribution to reduce bias. For item-centric summarization, we chose 100 items, split equally between the 50 most and 50 least popular items, ensuring both popular and less popular items were evaluated. These sampled subsets also form the user-group and item-group scenarios. 
We also conducted an additional experiment for user-group and item-group scenarios with varying group sizes to evaluate scalability.

\textit{\textbf{Baseline explanation paths.}} The main experiments utilize individual explanation paths generated with the PGPR \cite{xian2019reinforcement} and CAFE \cite{xian2020cafe} methods for a sampled dataset of 200 users. These paths explain the top-10 item recommendations, each reaching the recommended item within a maximum of three edges. The preprocessing of these paths involved generating an incremental set of top-k recommendation paths for \(k=1\) to 10 for each user. We also included PLM \cite{plm} and PLMR \cite{balloccu2023faithful} baselines for additional comparison, with results presented in Figure \ref{fig:perf_cmpr_more} and Figure \ref{fig:perf_div_morebaselines}.

\textit{\textbf{Parameters and Simplifications.}} In our experiment, we conduct summarization experiments across \( k \in [1, 10] \) that represents the number of top recommendations provided by the recommender. For the Steiner summaries, we used three values of \(\lambda\) to vary the influence of the individual explanation paths: \(\lambda = 0.01\), \(\lambda = 1\), and \(\lambda = 100\) to represent low, moderate, and high influence, respectively. In practice, we found that using edge weights in the PCST summarization led to excessively large summaries, making the \(\lambda\) parameter irrelevant. Preliminary experiments showed that this approach did not perform well. As a result, we opted to ignore the edge weights for the PCST summaries. To create PCST summaries, we assigned node prizes \( p(v) \) such that \( p(v) = 1 \) for nodes in the set of terminal nodes \( T \), and \( p(v) = 0 \) for nodes not in \( T \). As in previous works and for our results to be directly comparable with baseline methods, we set \( w_A = 0 \) \cite{xian2019reinforcement,xian2020cafe,balloccu2023faithful}, and did not consider rating recency by setting \( \beta_2 = 0 \). 
We also conducted an additional experiment, shown in Figure \ref{fig:radar_chart}, to assess the impact of recency on summary explanations by adjusting parameters \( \beta_1 \) and \( \beta_2 \), which control the balance between rating score and recency in recommendations.

\textit{\textbf{Implementation.}} The experiments were conducted on a 32 GB RAM, 256 GB SSD, Linux-based PC equipped with a 6-core CPU. Our summarization implementations and necessary baseline data are available in our  GitHub repository\footnote{\url{https://github.com/xsum-rec/xsum}}.

\subsection {Evaluation Metrics and Experimental Results}
\label{section:eval}

For evaluating our summary explanations, we propose a suite of metrics aiming at evaluating different aspects of the quality of the explanations. The proposed metrics are based on related metrics used in evaluating explainable artificial intelligence, in general, \cite{10.1007/978-3-031-20319-0_30, zytek2024llmsxaifuturedirections, rosenfeld2021better, Zhou2021EvaluatingTQ, 10.1007/978-3-031-40878-6_12, nguyen2020quantitative, yang2018towards} and the few path-based specific explanation evaluation metrics \cite{markchom2023explainable, balloccu2022post, 10.1007/978-3-031-28241-6_1} found in the literature.
These definitions are generalized to be applicable to general subgraphs besides paths. We propose appropriate variations of these metrics adopted for the novel problem of subgraph-based summary explanations. The following subsections present the metrics and results given an explanation subgraph \( S = (V_S, E_S, w) \).

\subsubsection {Comprehensibility}
\begin{figure}[ht]
    \centering   
    \begin{subfigure}{0.24\textwidth}
        \centering
        \includegraphics[width=\textwidth]{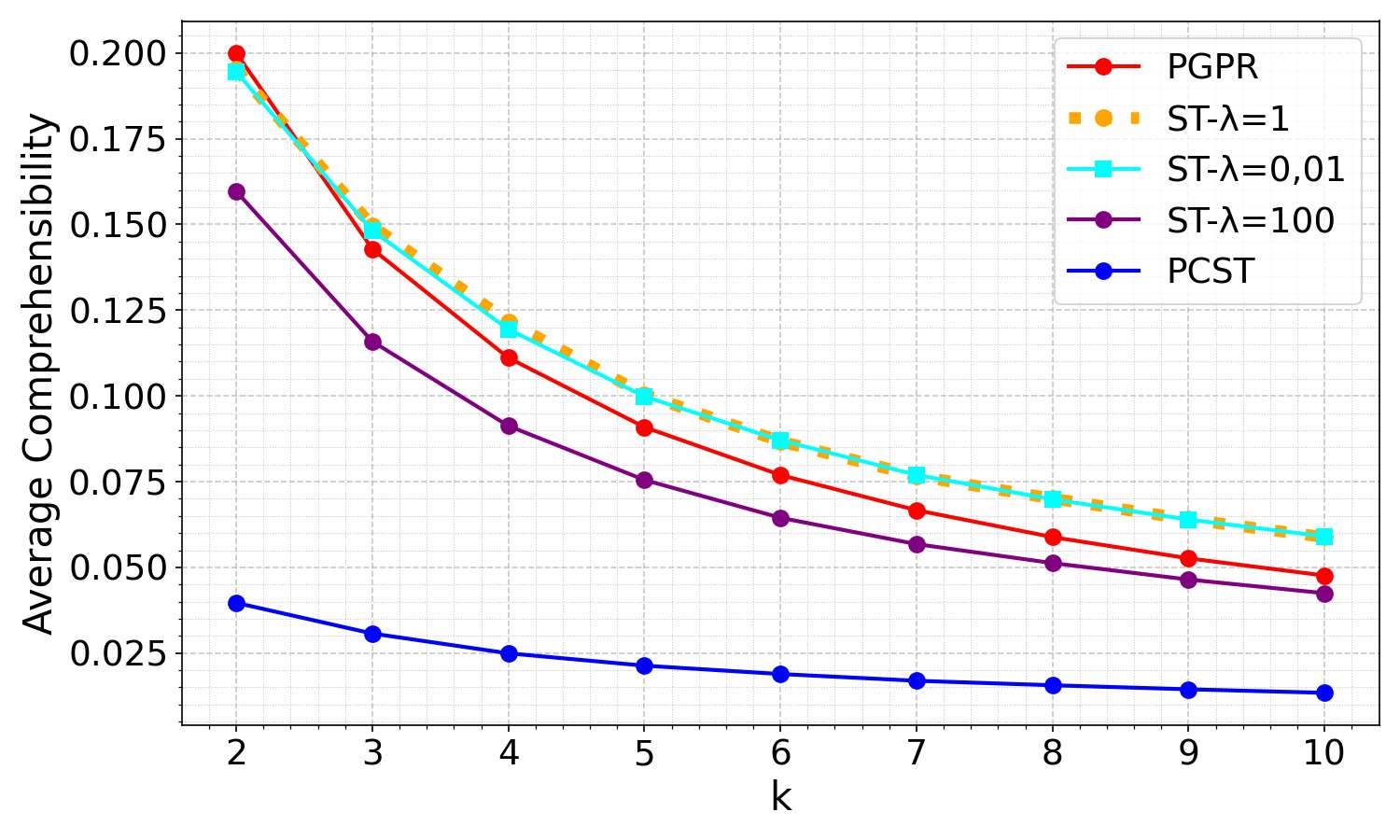}
        \caption{User-centric PGPR}
        \label{fig:cmpr_pgpr_user}
    \end{subfigure}
    \hfill
    \begin{subfigure}{0.24\textwidth}
        \centering
        \includegraphics[width=\textwidth]{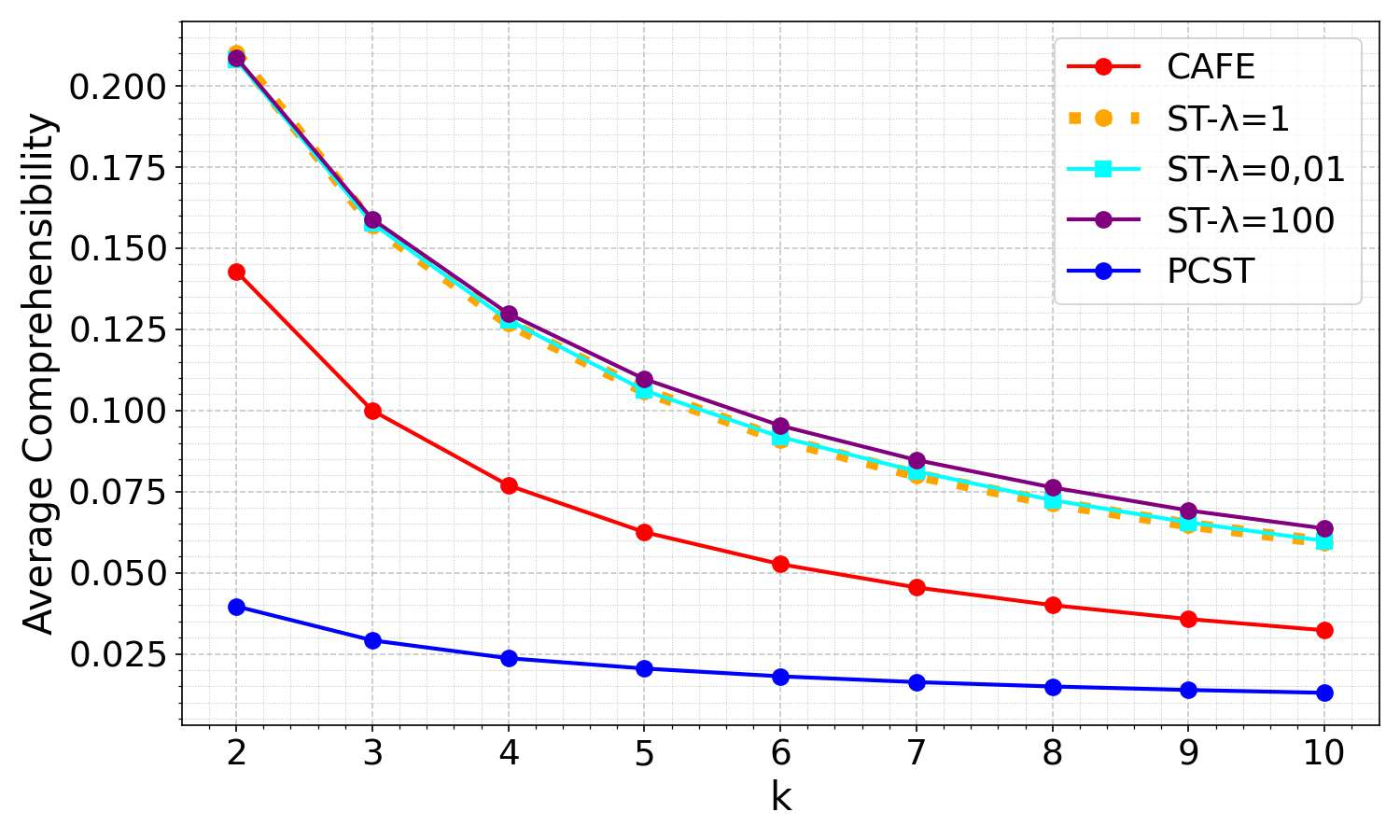}
        \caption{User-centric CAFE}
        \label{fig:cmpr_cafe_user}
    \end{subfigure}
    \begin{subfigure}{0.24\textwidth} 
        \centering
        \includegraphics[width=\textwidth]{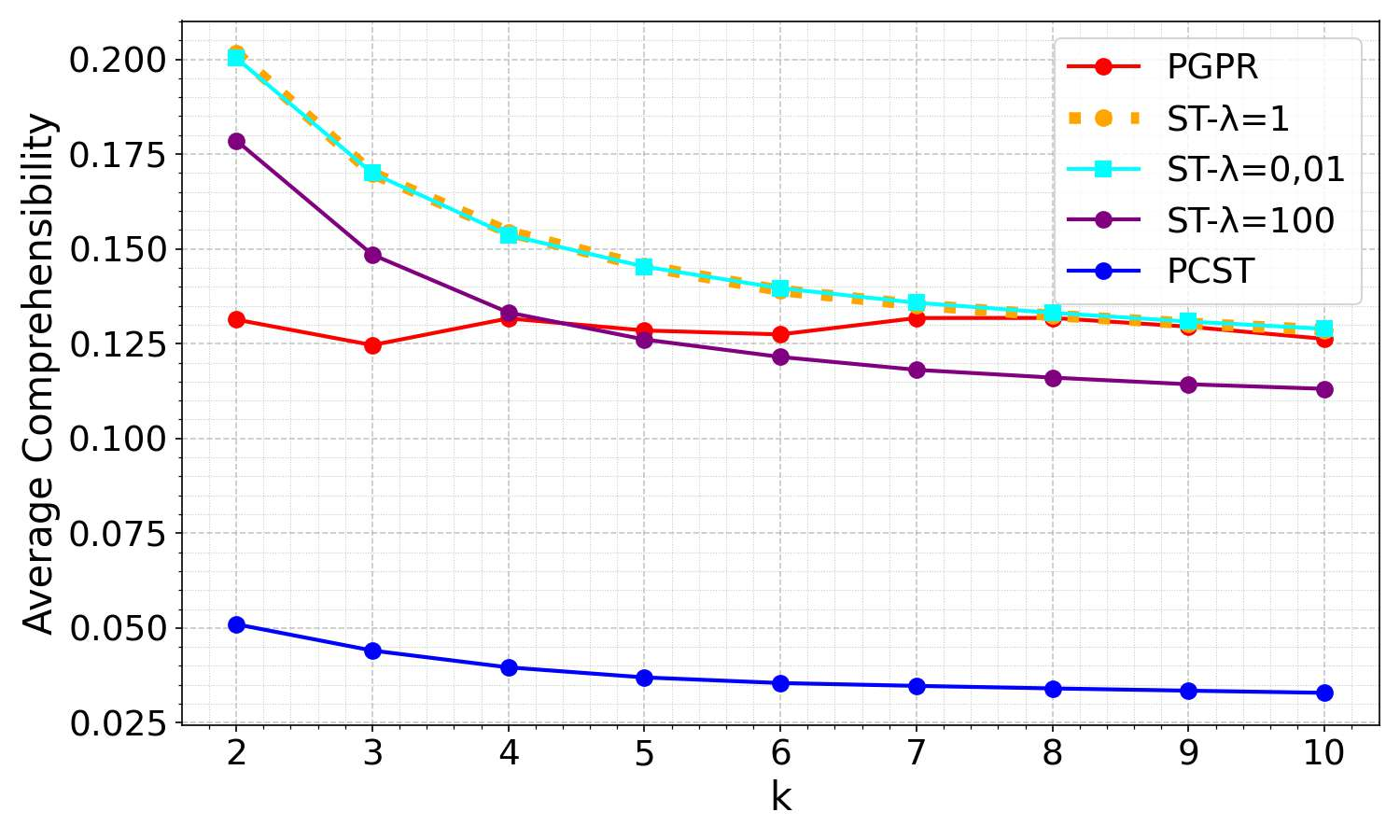}
        \caption{Item-centric PGPR}
        \label{fig:cmpr_pgpr_item}
    \end{subfigure}
    \hfill
    \begin{subfigure}{0.24\textwidth}
        \centering
        \includegraphics[width=\textwidth]{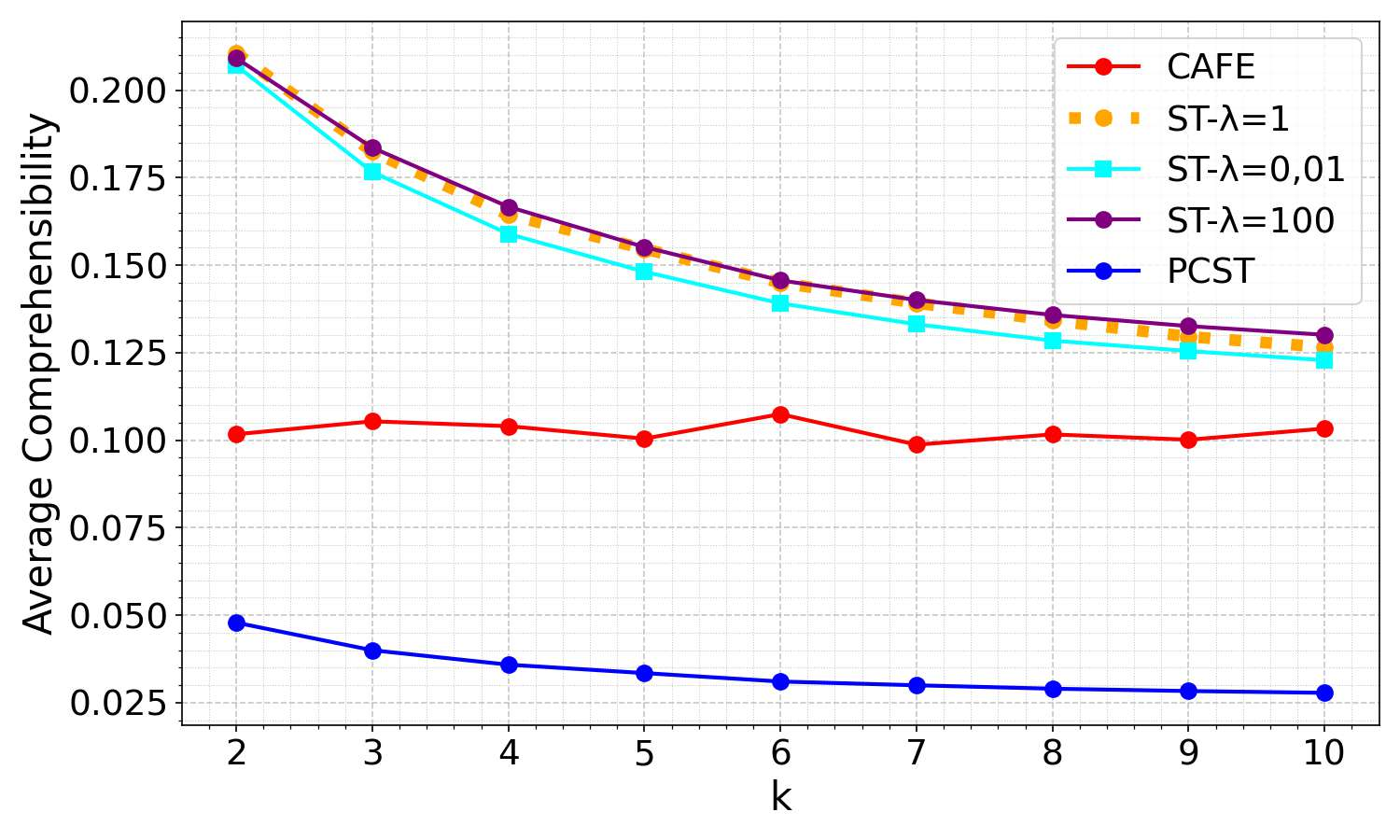}
        \caption{Item-centric CAFE}
        \label{fig:cmpr_cafe_item}
    \end{subfigure}
    \begin{subfigure}{0.24\textwidth} 
        \centering
        \includegraphics[width=\textwidth]{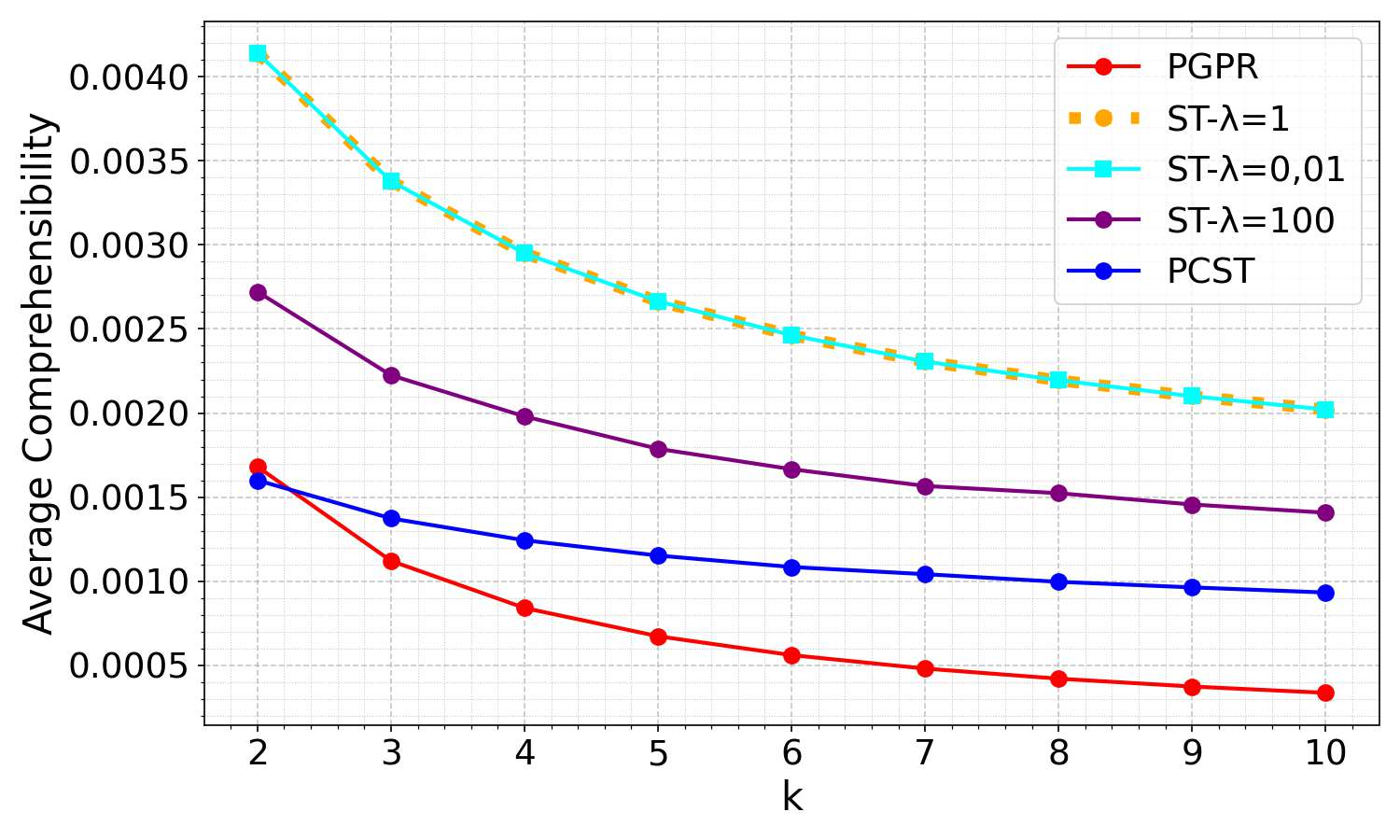}
        \caption{User-group PGPR}
        \label{fig:cmpr_pgpr_user_group}
    \end{subfigure}
    \hfill
    \begin{subfigure}{0.24\textwidth} 
        \centering
        \includegraphics[width=\textwidth]{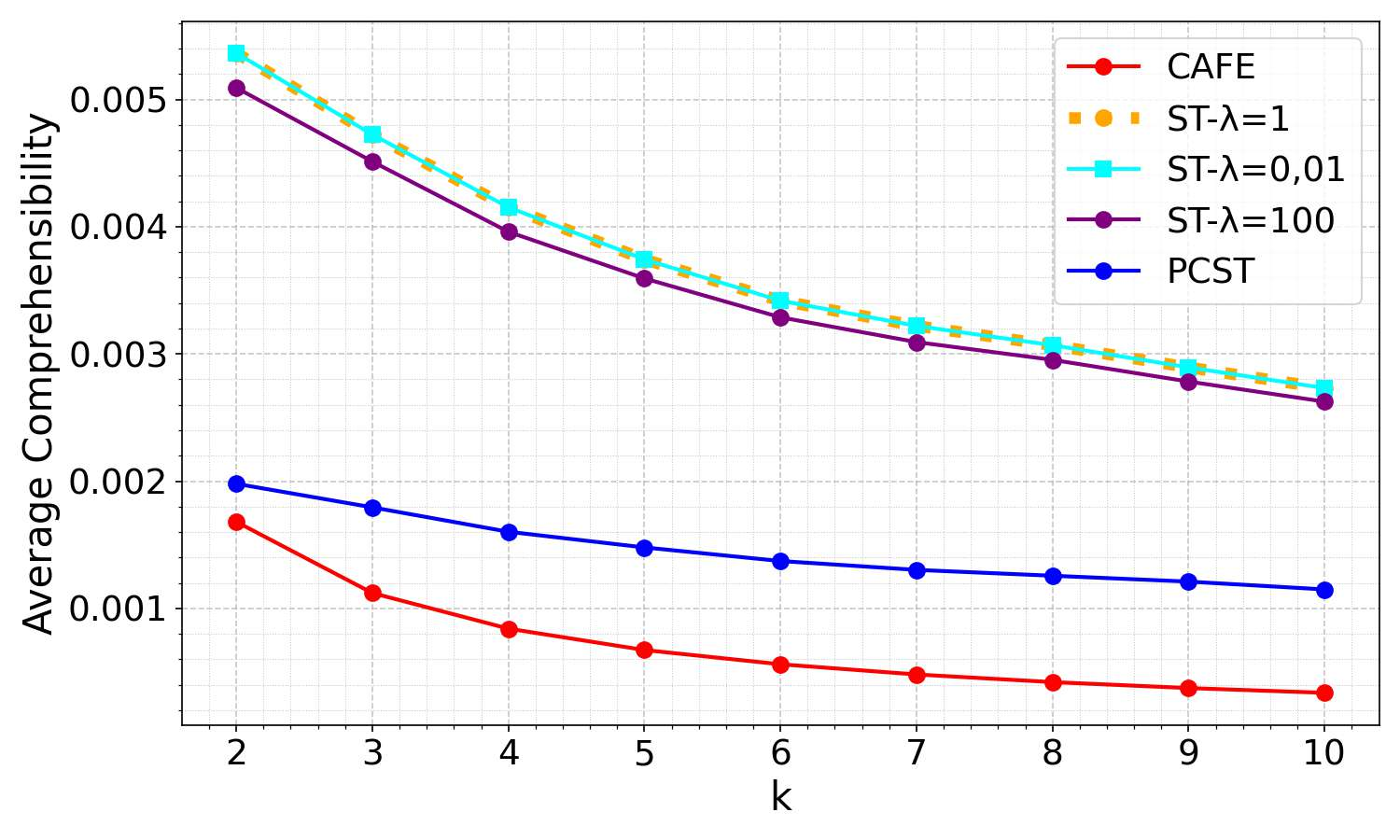}
        \caption{User-group CAFE}
        \label{fig:cmpr_cafe_user_group}
    \end{subfigure}
    \begin{subfigure}{0.24\textwidth}
        \centering
        \includegraphics[width=\textwidth]{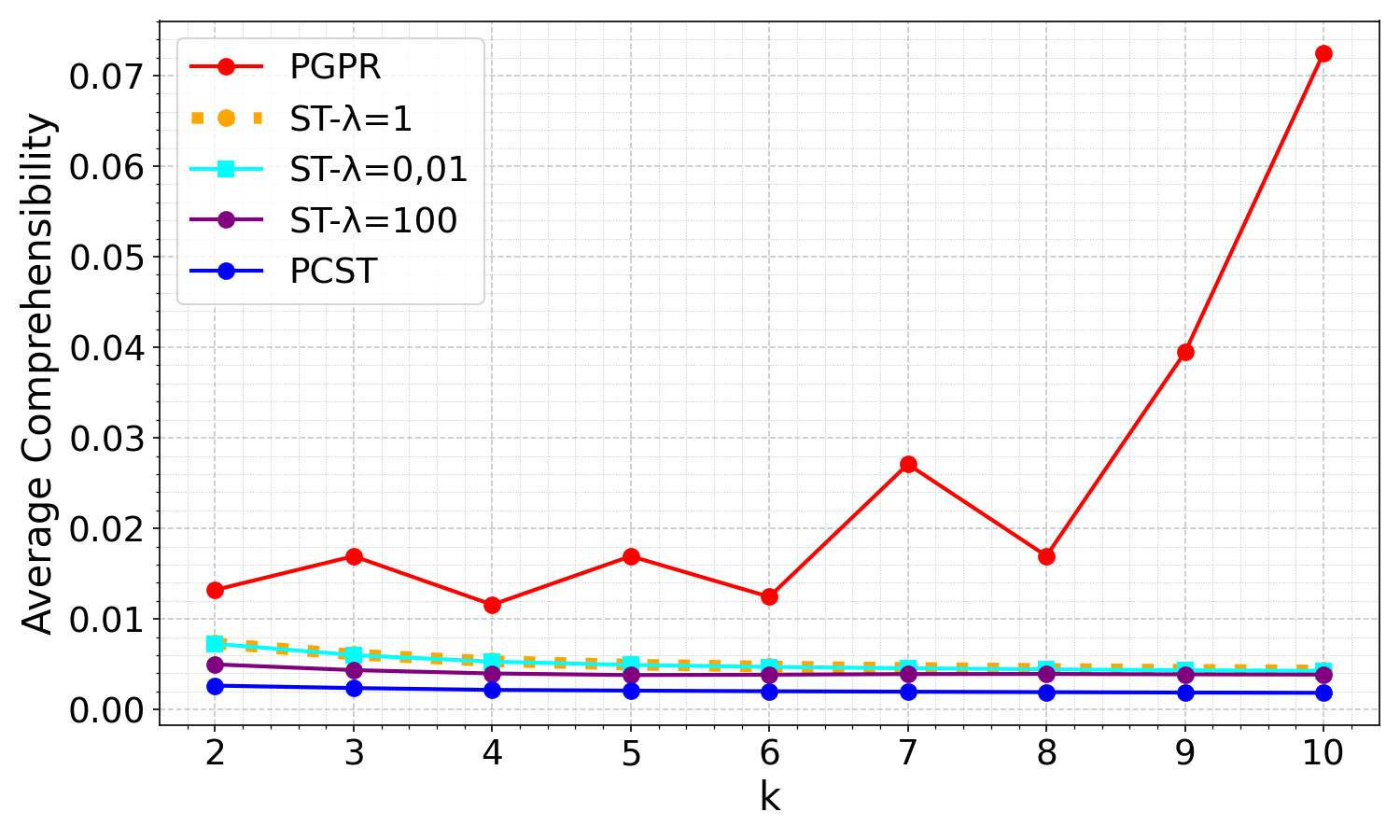}
        \caption{Item-group PGPR}
        \label{fig:cmpr_pgpr_item_group}
    \end{subfigure}
    \hfill
    \begin{subfigure}{0.24\textwidth}
        \centering
        \includegraphics[width=\textwidth]{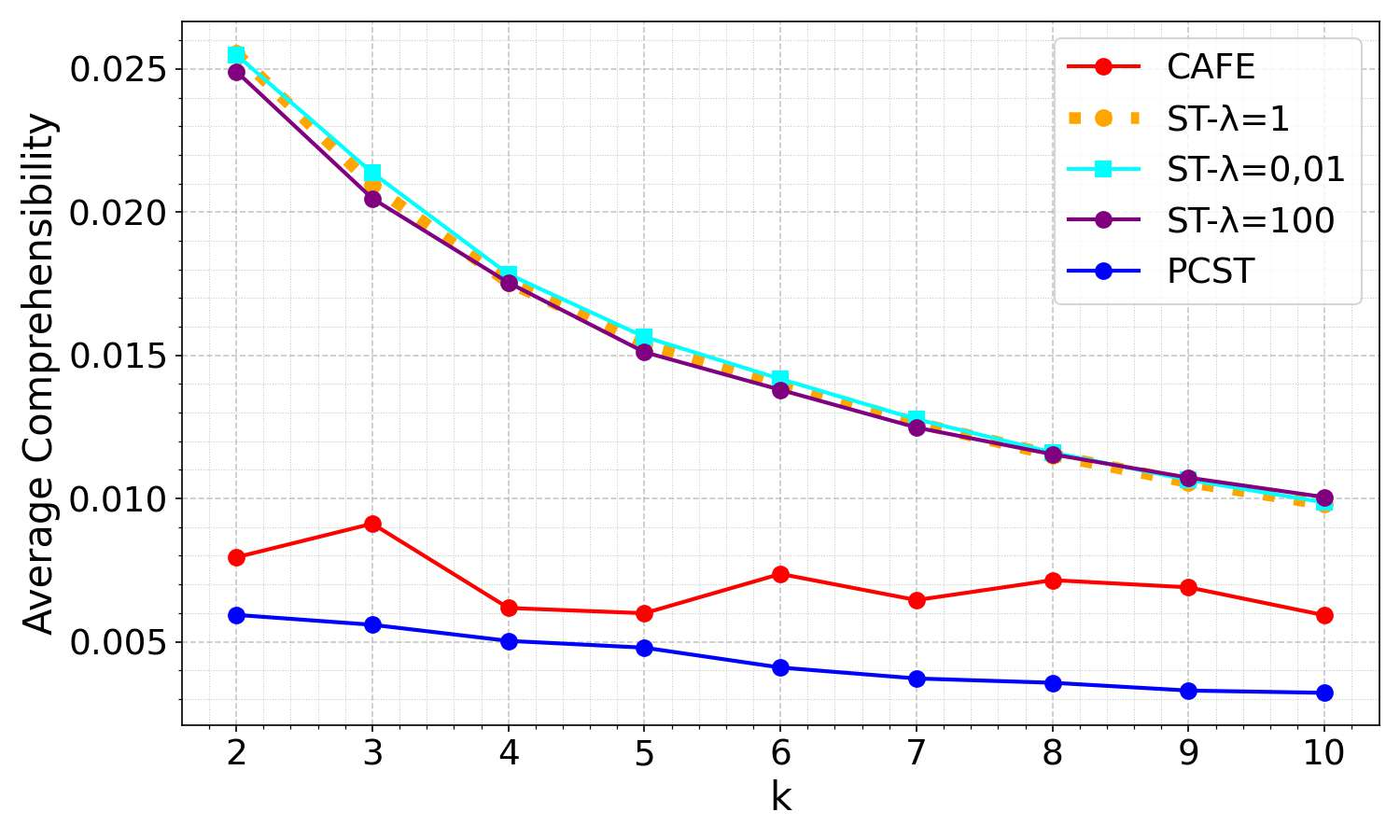}
        \caption{Item-group CAFE}
        \label{fig:cmpr_cafe_item_group}
    \end{subfigure}
    \caption{Comprehensibility}
    \label{fig:cmp}
\end{figure}

The comprehensibility metric, $C(S)$, aims to quantify how easily users can understand the explanation paths.
A higher comprehensibility score corresponds to a briefer and more understandable explanation. The original metric is inversely proportional to the length of the explanation paths, in summary explanations, it is inversely related to the explanation subgraph size: $C(S) = \frac{1}{|E_S|}$. 
Figure \ref{fig:cmp} shows that the ST method outperforms all methods because it connects terminal nodes in a single tree and creates a more compact summary than baselines, which use separate 3-hop paths for each item. PCST outperforms baselines only in user-group scenarios and creates larger trees than ST because, without edge weights to guide path minimization, it focuses solely on connecting high-prize nodes (see \ref{setup}), often including additional nodes to ensure connectivity.
 
\subsubsection{Actionability}
\begin{figure}[ht]
    \centering
    \begin{subfigure}{0.24\textwidth} 
        \centering
         \includegraphics[width=\textwidth]{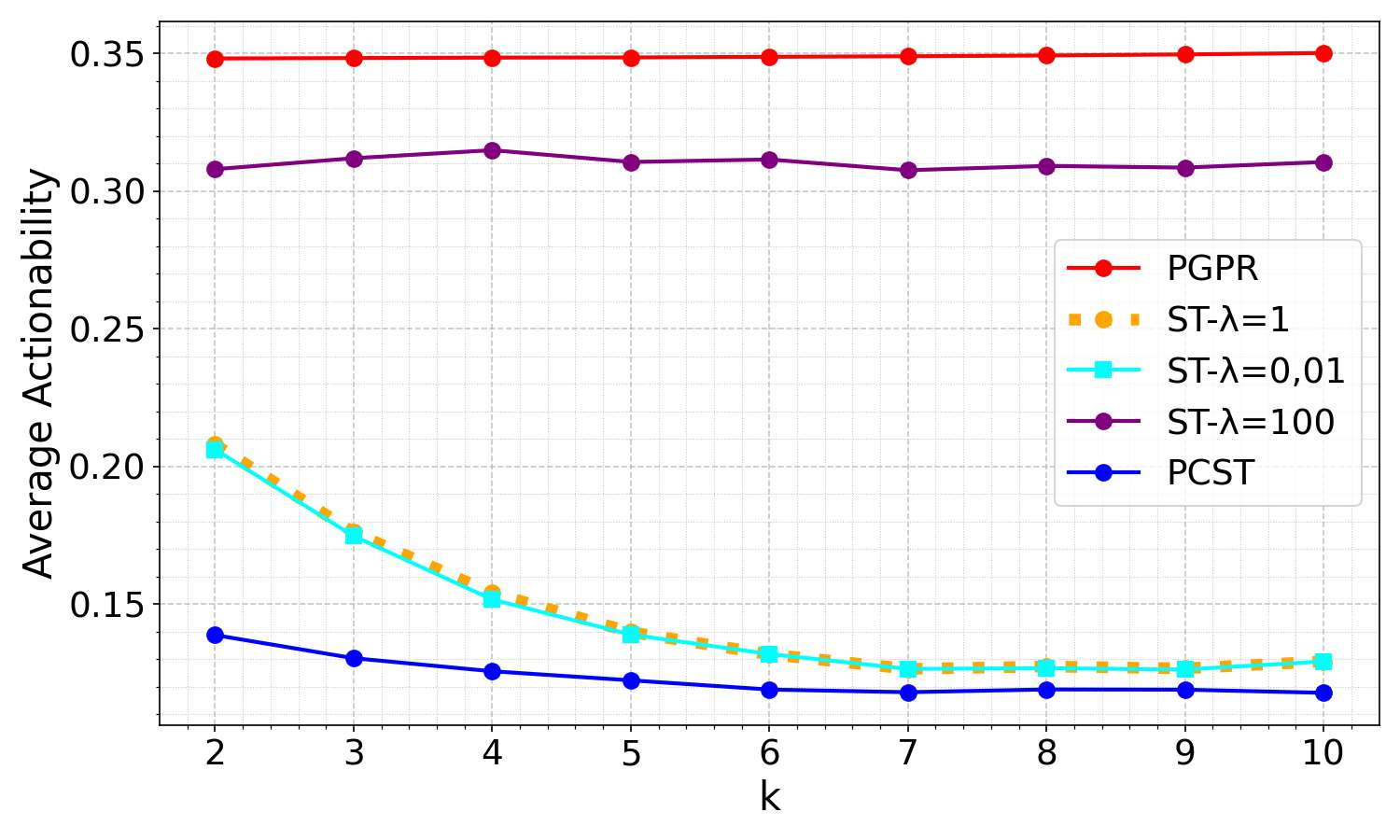}

        \caption{User-centric PGPR}
        \label{fig:act_pgpr_user}
    \end{subfigure}
    \hfill
    \begin{subfigure}{0.24\textwidth} 
        \centering
        \includegraphics[width=\textwidth]{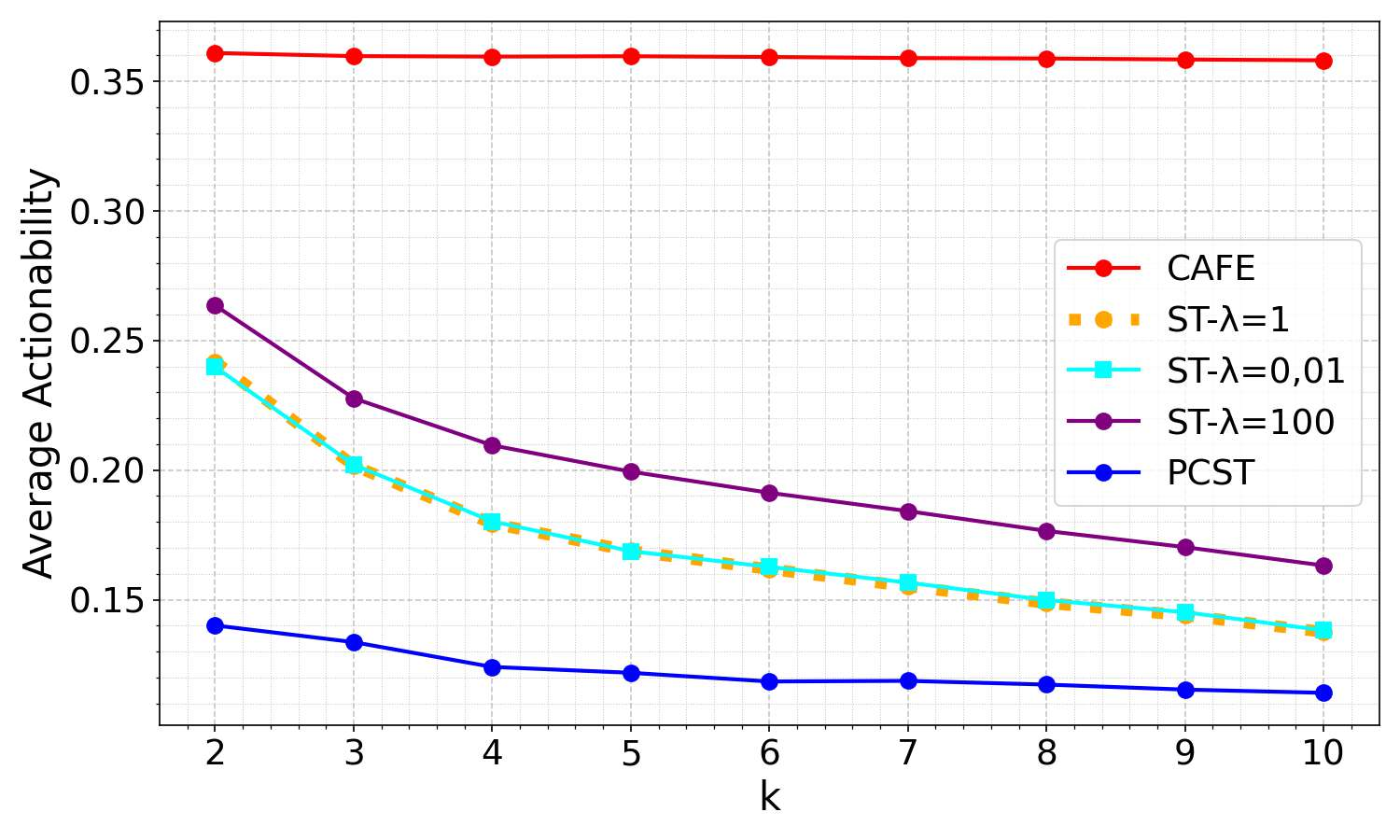}
        \caption{User-centric CAFE}
        \label{fig:act_cafe_user}
    \end{subfigure}


    \begin{subfigure}{0.24\textwidth} 
        \centering
        \includegraphics[width=\textwidth]{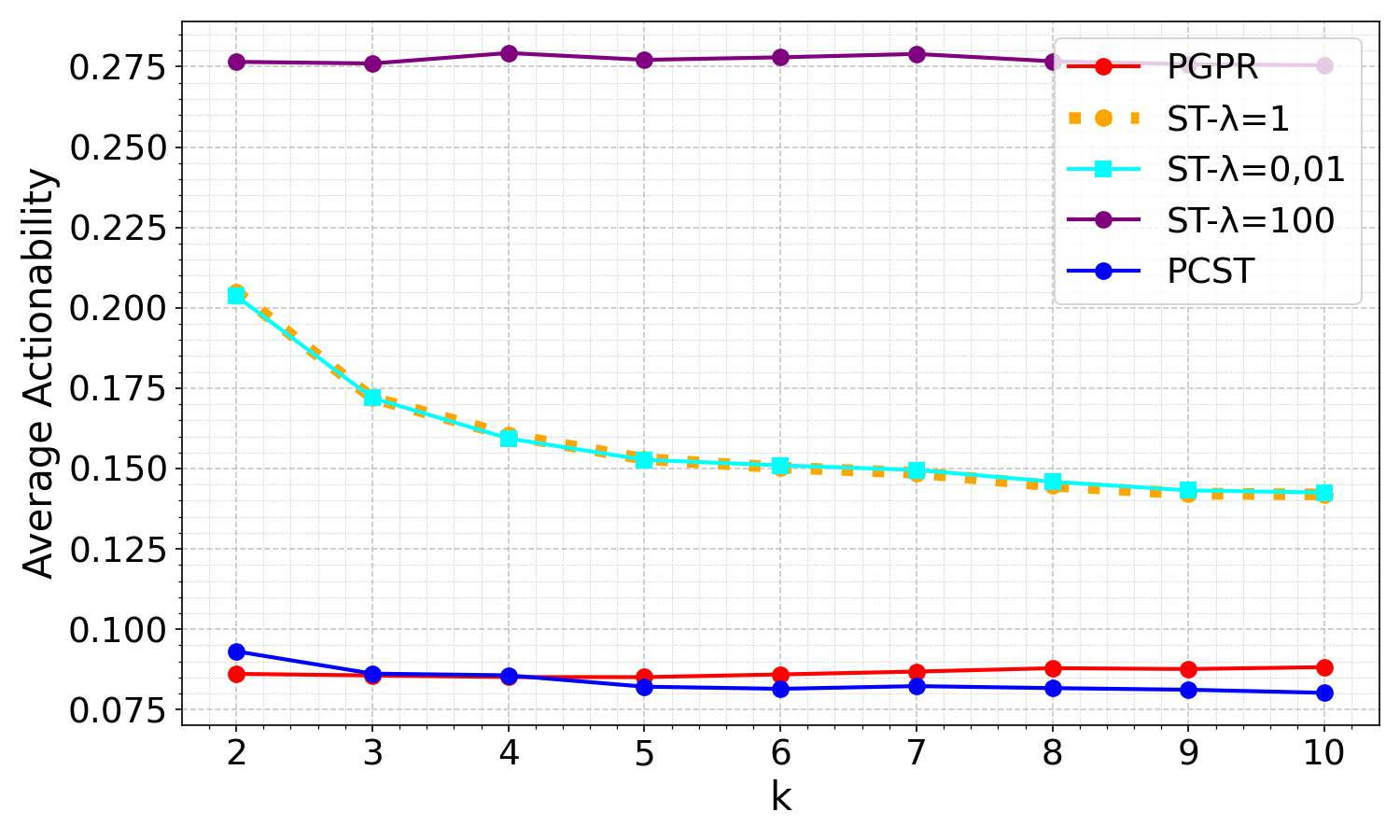}
        \caption{Item-centric PGPR}
        \label{fig:act_pgpr_item}
    \end{subfigure}
    \hfill
    \begin{subfigure}{0.24\textwidth} 
        \centering
        \includegraphics[width=\textwidth]{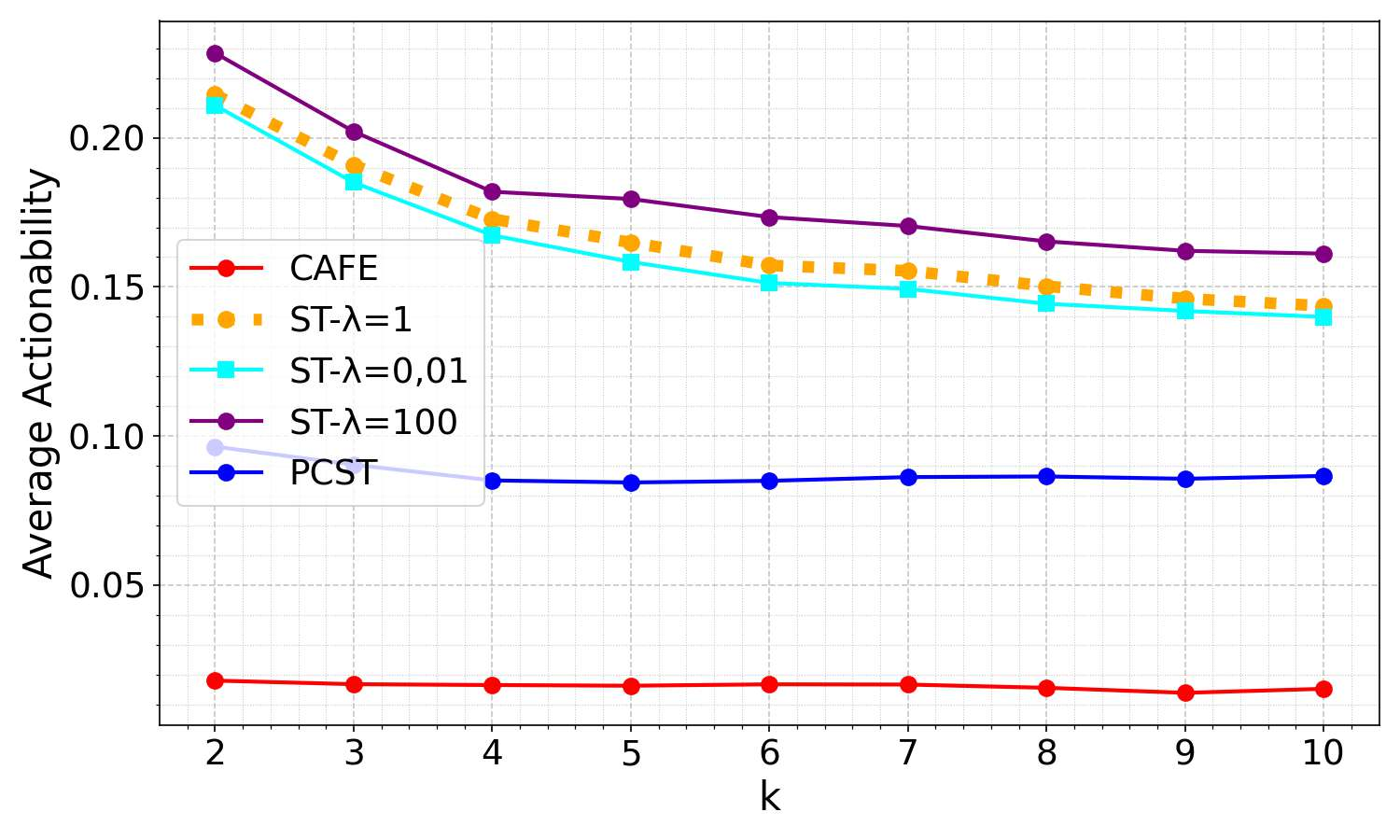}
        \caption{Item-centric CAFE}
        \label{fig:act_cafe_item}
    \end{subfigure}


    \begin{subfigure}{0.24\textwidth} 
        \centering
        \includegraphics[width=\textwidth]{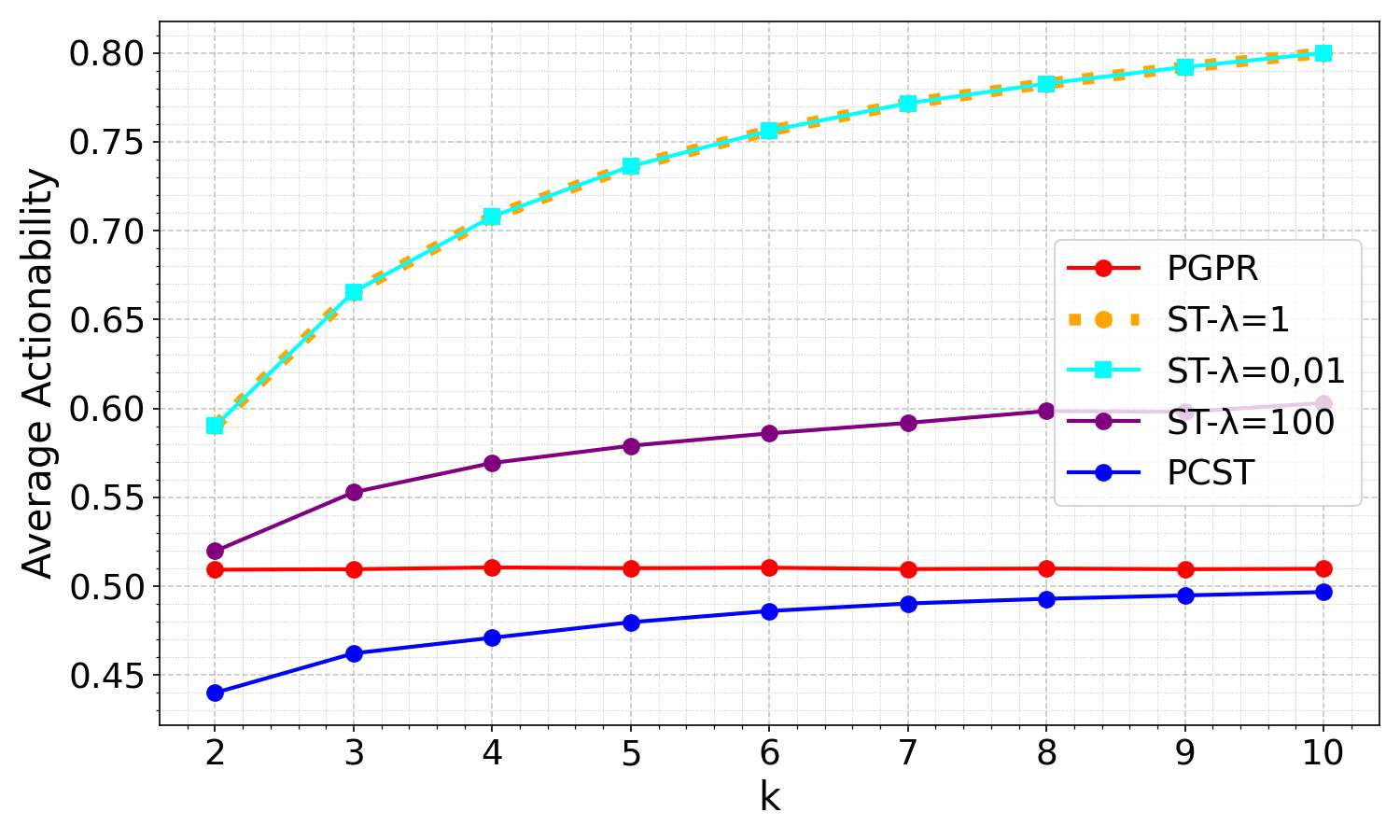}
        \caption{User-group PGPR}
        \label{fig:act_pgpr_user_group}
    \end{subfigure}
    \hfill
    \begin{subfigure}{0.24\textwidth} 
        \centering
        \includegraphics[width=\textwidth]{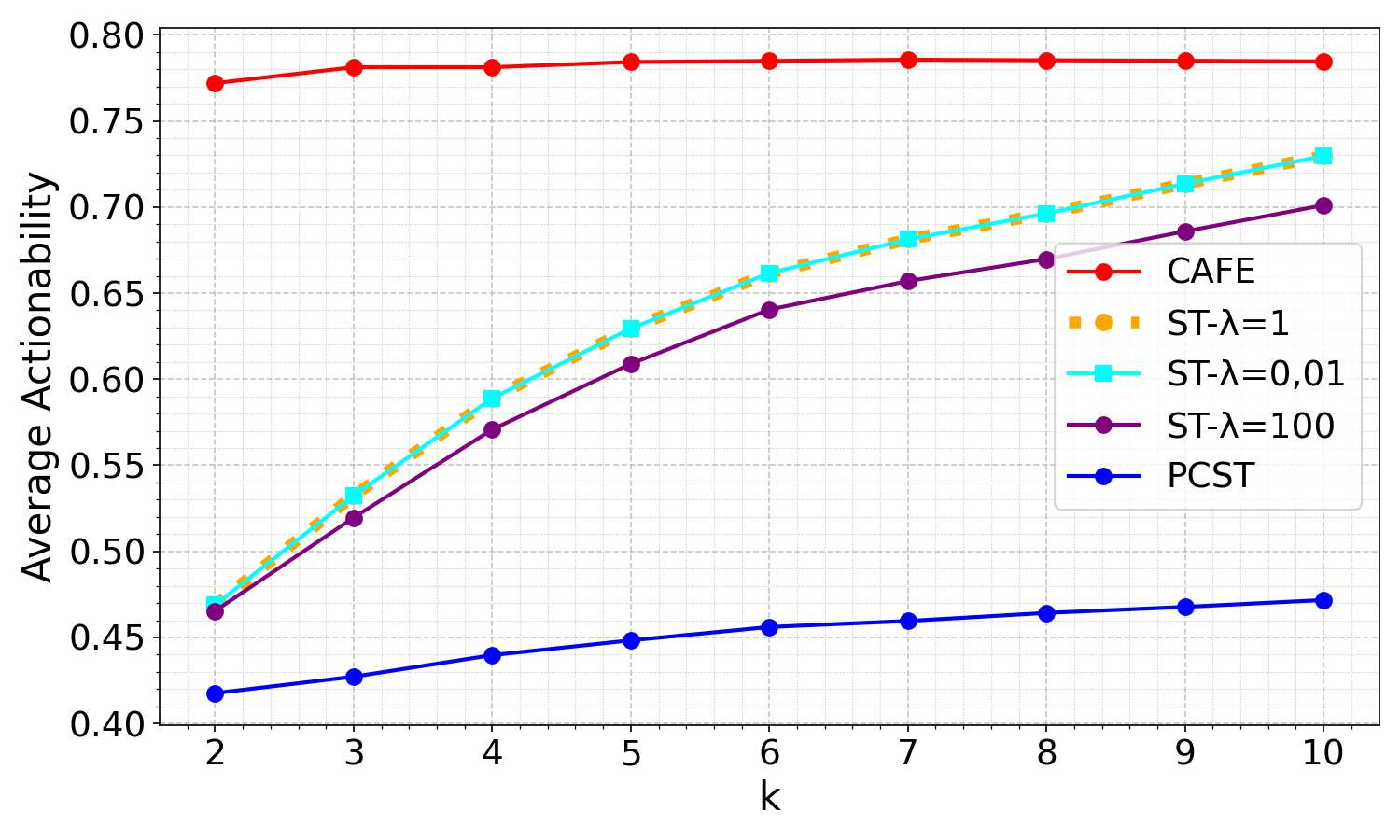}
        \caption{User-group CAFE}
        \label{fig:act_cafe_user_group}
    \end{subfigure}


    \begin{subfigure}{0.24\textwidth} 
        \centering
        \includegraphics[width=\textwidth]{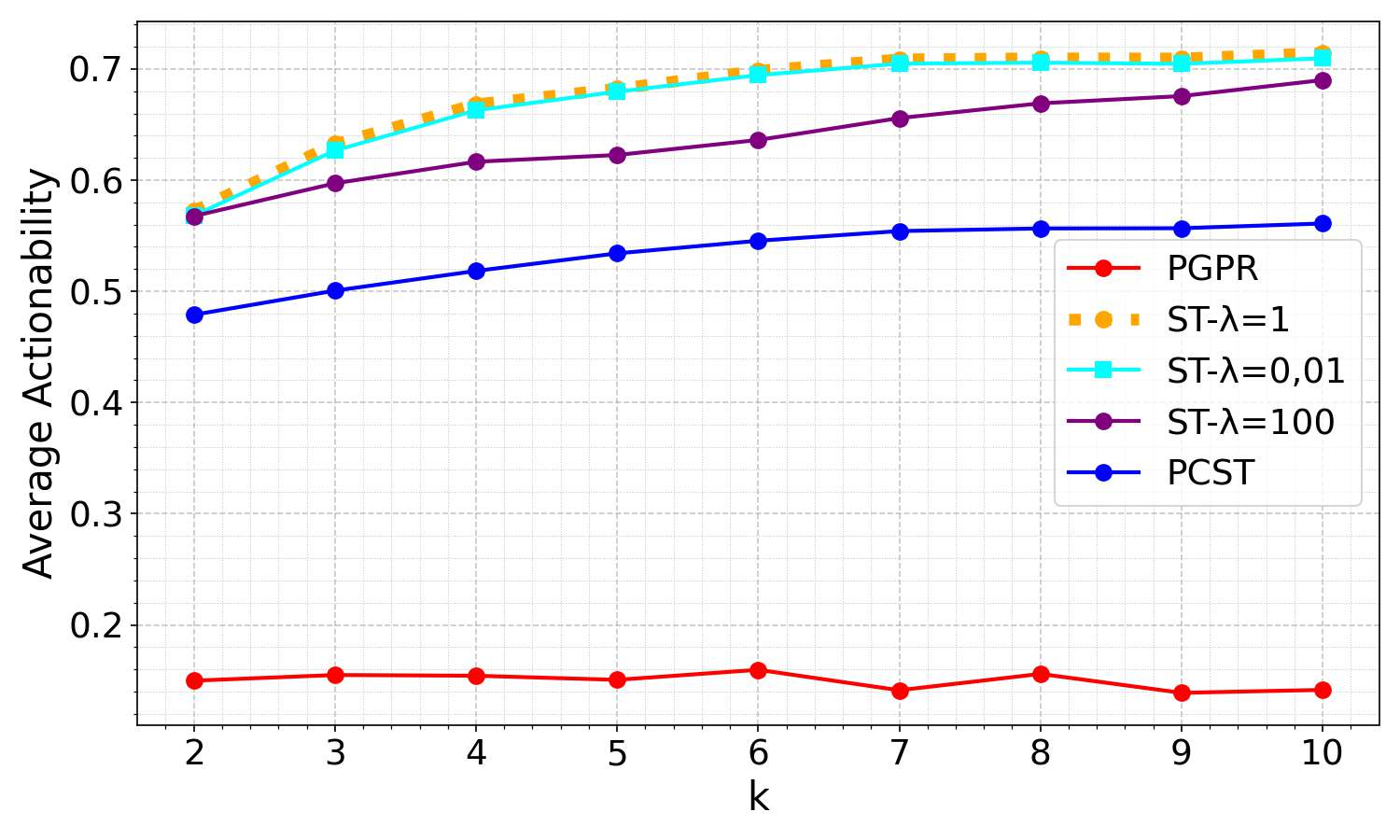}
        \caption{Item-group PGPR}
        \label{fig:act_pgpr_item_group}
    \end{subfigure}
    \hfill
    \begin{subfigure}{0.24\textwidth} 
        \centering
        \includegraphics[width=\textwidth]{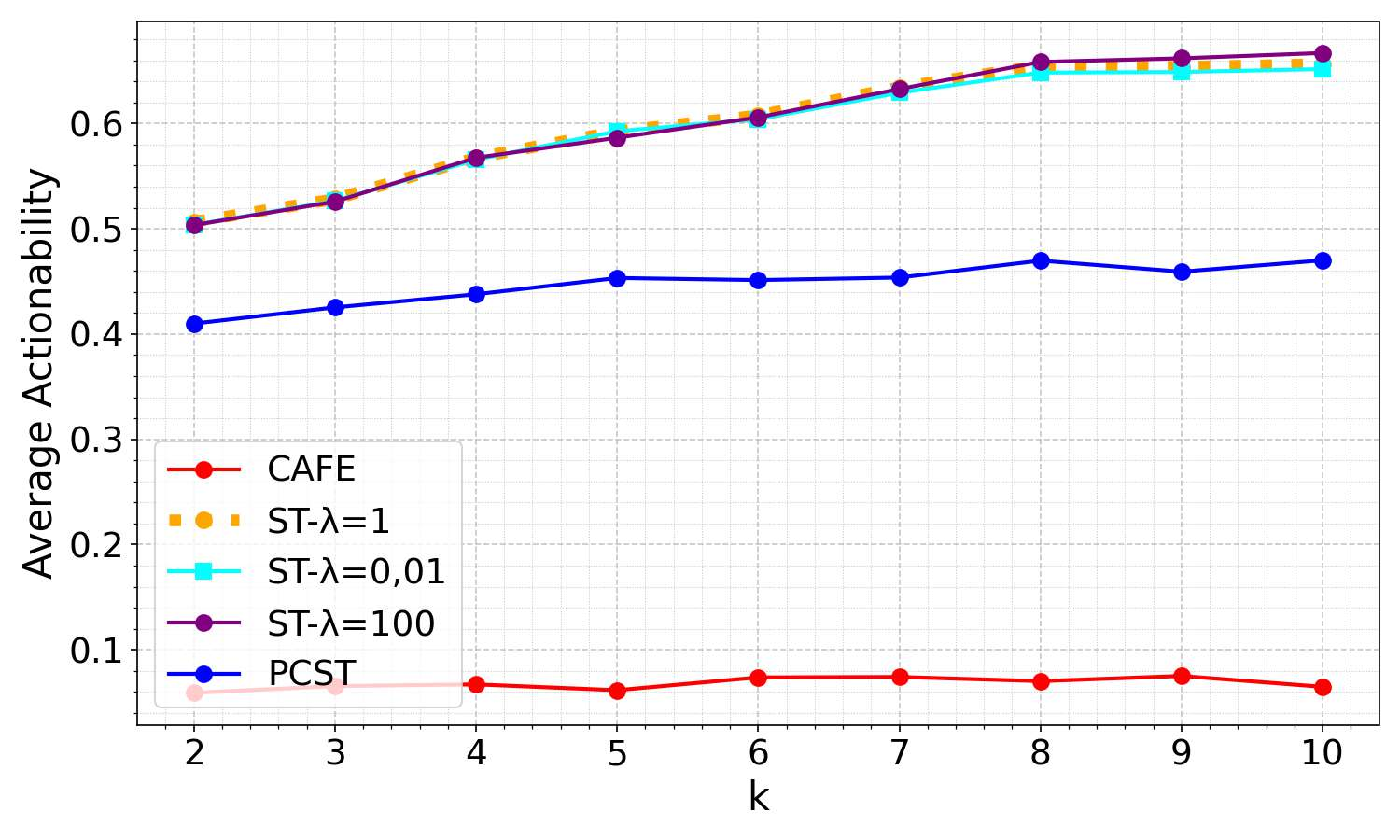}
        \caption{Item-group CAFE}
        \label{fig:act_cafe_item_group}
    \end{subfigure}
    \caption{Actionability}
\label{fig:act}
\end{figure}

The actionability metric, $A(S)$, evaluates the usefulness of explanation paths based on actionability, i.e., the ability of users to modify their recommendations. We assume that item—nodes in \(I\) are actionable, since users can interact with the recommendation system and affect recommendation results by modifying the ratings to items. In contrast, user nodes (\(U\)) and knowledge nodes (\(A\)) are non-actionable, as users cannot modify them in the recommendation process. A higher actionability score indicates more user control over recommendations. 
In the case of paths, actionability is calculated as the proportion of actionable items in the explanation path. Analogously, in summary subgraphs, actionability is the ratio of item nodes in subgraph \(S\) to the total nodes: $A(S) = \frac{\text{Number of item nodes in } S}{|V_S|}$. 
As shown in Figure \ref{fig:act}, the baseline paths maintain high and nearly stable actionability in user-centric and user-group scenarios, since at least one item node (the recommended item) is included in each path. ST with $\lambda = 100$ achieves the highest actionability scores across most summary scenarios since it prioritizes item nodes rated by the user, thus increasing the actionable items in the summary and leading to higher actionability scores. PCST is the least effective, because it is not optimized for item inclusion; this could improve with a node-prize assignment that prioritizes items.

\subsubsection{Diversity Metric}
The diversity metric, $D(S)$, measures the variety among explanation paths, with higher diversity scores indicating a broader range of explanations.
Specifically, diversity reflects the uniqueness of nodes (items, users, or external entities) within the edges of explanation paths, ensuring users receive varied information. For summary subgraphs, given two edges \( e_i \) and \( e_j \) within \( S \), diversity is calculated by comparing unique nodes using the Jaccard similarity: $J(e_i, e_j) = \frac{|V_{e_i} \cap V_{e_j}|}{|V_{e_i} \cup V_{e_j}|}$, where \( V_{e_i} \) and \( V_{e_j} \) are the sets of nodes connected by edges \( e_i \) and \( e_j \), and \( i, j \) are indices referring to any pair of edges in \( S \). The diversity metric \( D(S) \) is defined as \( D(S) = \frac{1}{\binom{|E|}{2}} \sum_{e_i, e_j \in E_S} \left( 1 - J(e_i, e_j) \right)\).

\begin{figure}[ht]
    \centering
    
    \begin{subfigure}{0.24\textwidth} 
        \centering
        \includegraphics[width=\textwidth]{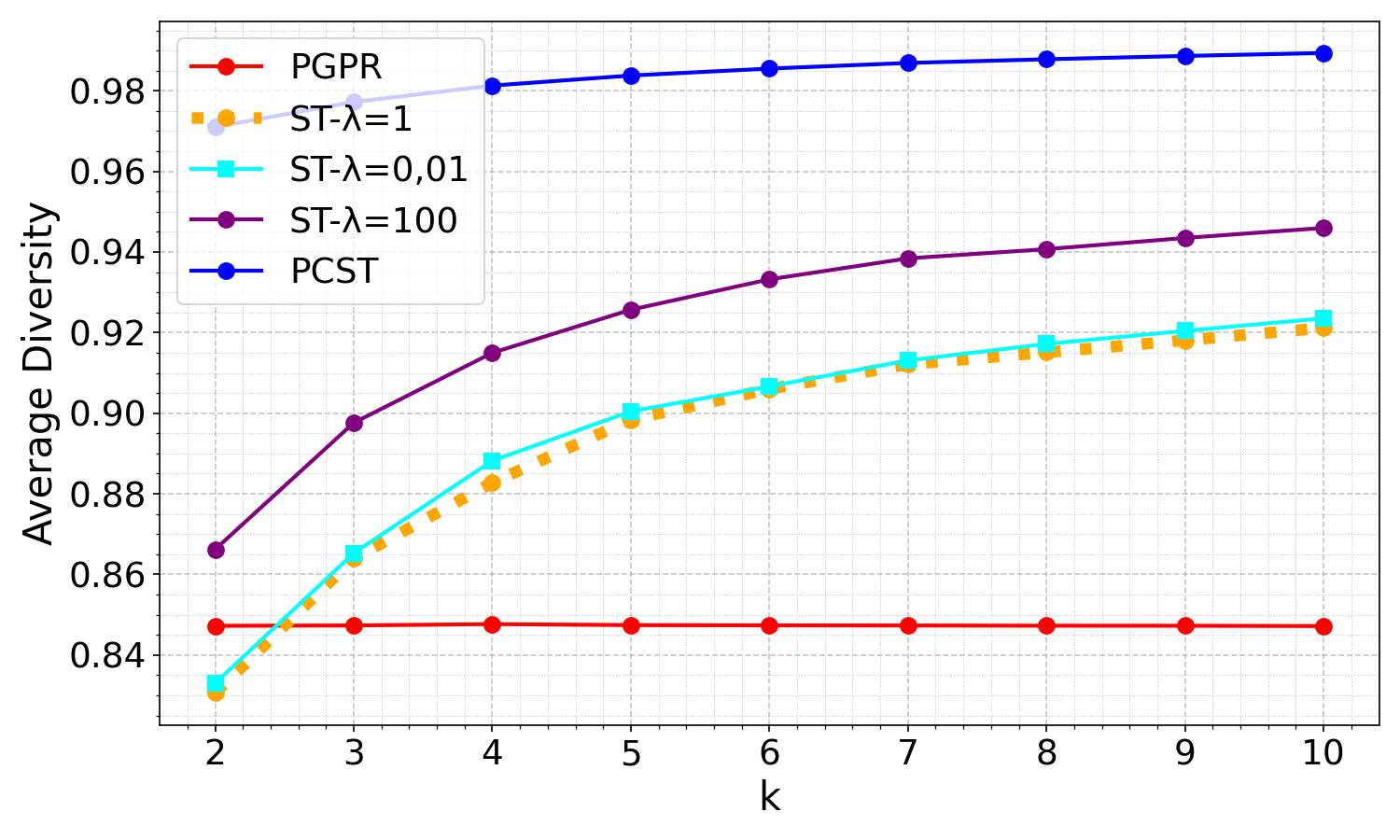}
        \caption{User-centric PGPR}
        \label{fig:div_pgpr_user}
    \end{subfigure}
    \hfill
    \begin{subfigure}{0.24\textwidth} 
        \centering
        \includegraphics[width=\textwidth]{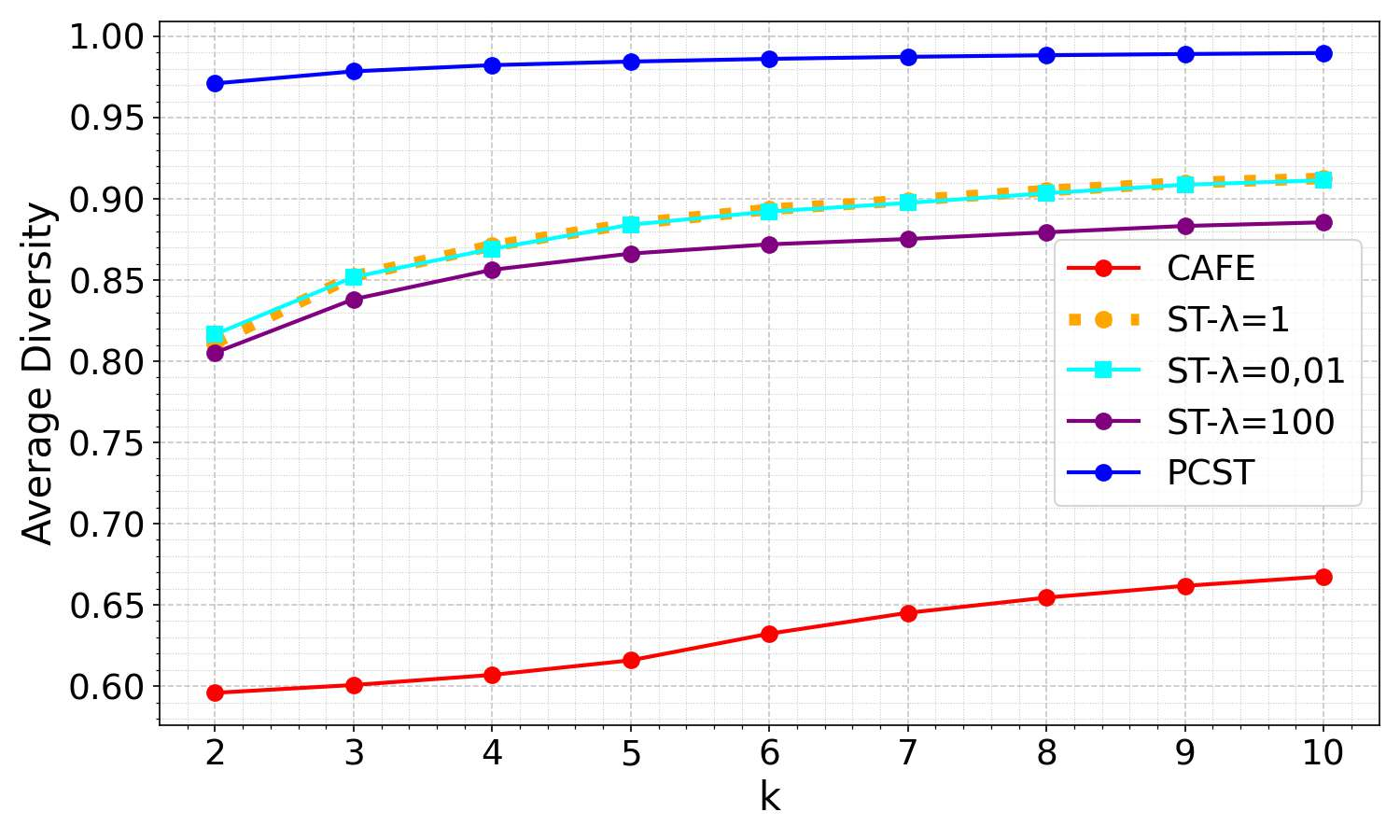}
        \caption{User-centric CAFE}
        \label{fig:div_cafe_user}
    \end{subfigure}


    \begin{subfigure}{0.24\textwidth} 
        \centering
        \includegraphics[width=\textwidth]{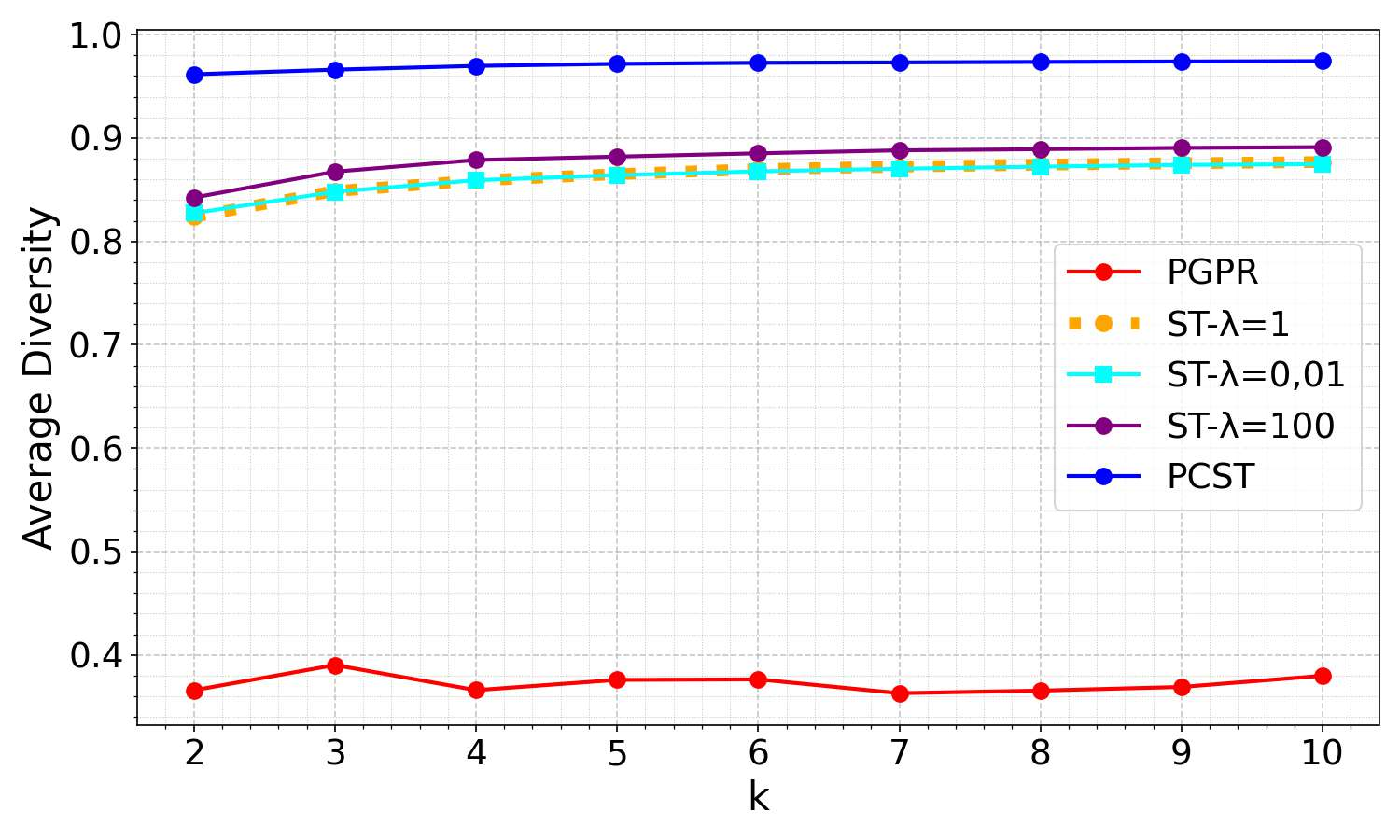}
        \caption{Item-centric PGPR}
        \label{fig:div_pgpr_item}
    \end{subfigure}
    \hfill
    \begin{subfigure}{0.24\textwidth} 
        \centering
        \includegraphics[width=\textwidth]{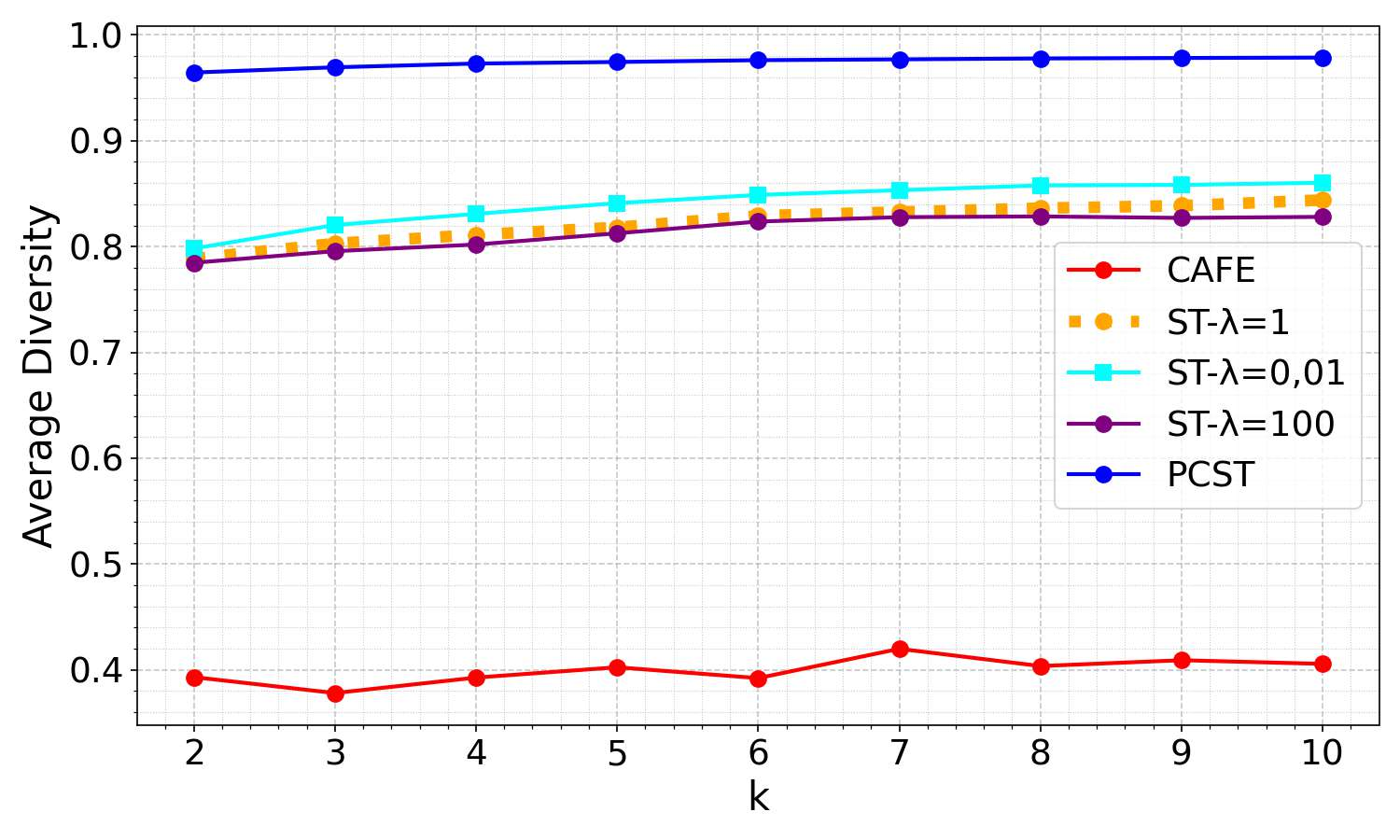}
        \caption{Item-centric CAFE}
        \label{fig:div_cafe_item}
    \end{subfigure}


    \begin{subfigure}{0.24\textwidth} 
        \centering
         \includegraphics[width=\textwidth]{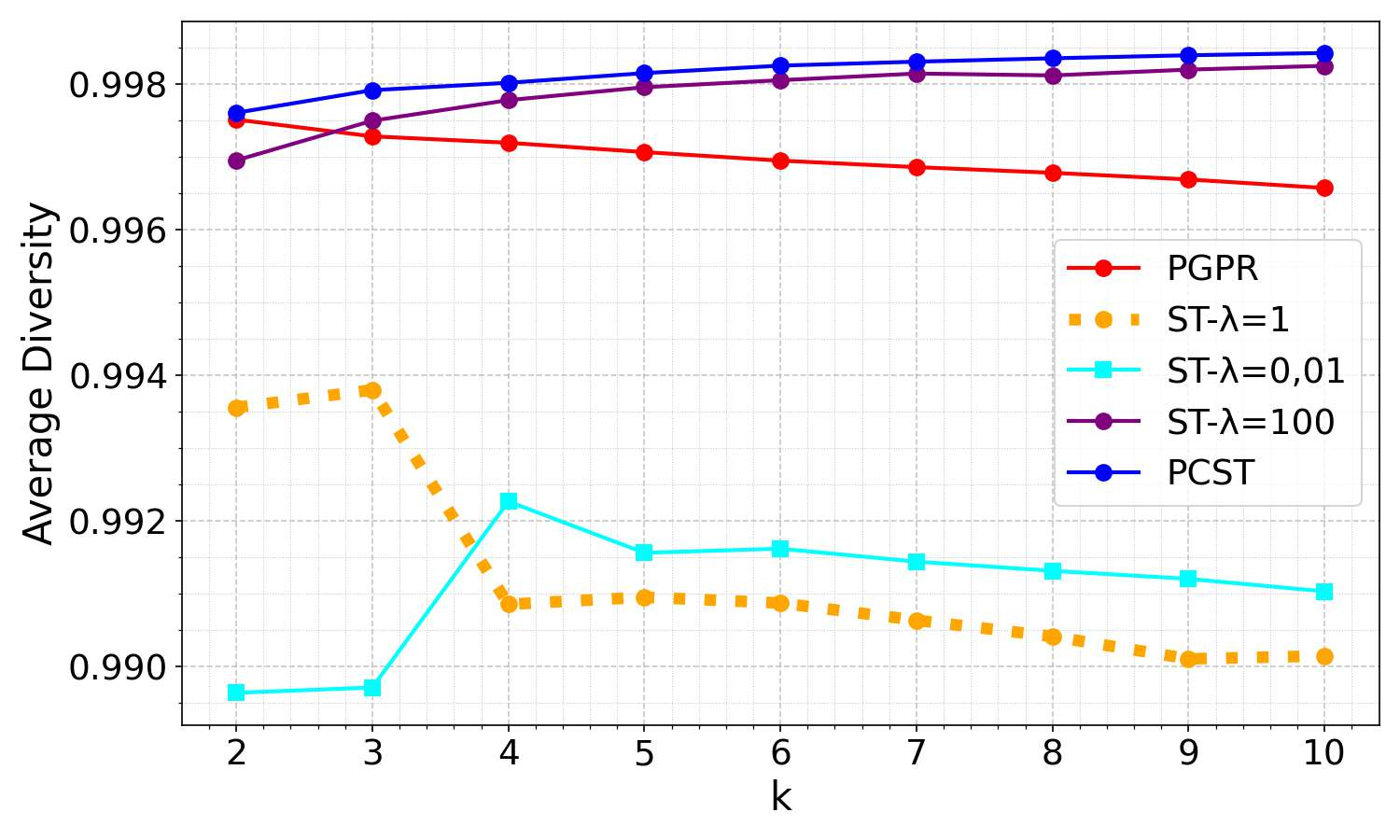}
        \caption{User-group PGPR}
        \label{fig:div_pgpr_user_group}
    \end{subfigure}
    \hfill
    \begin{subfigure}{0.24\textwidth} 
        \centering
        \includegraphics[width=\textwidth]{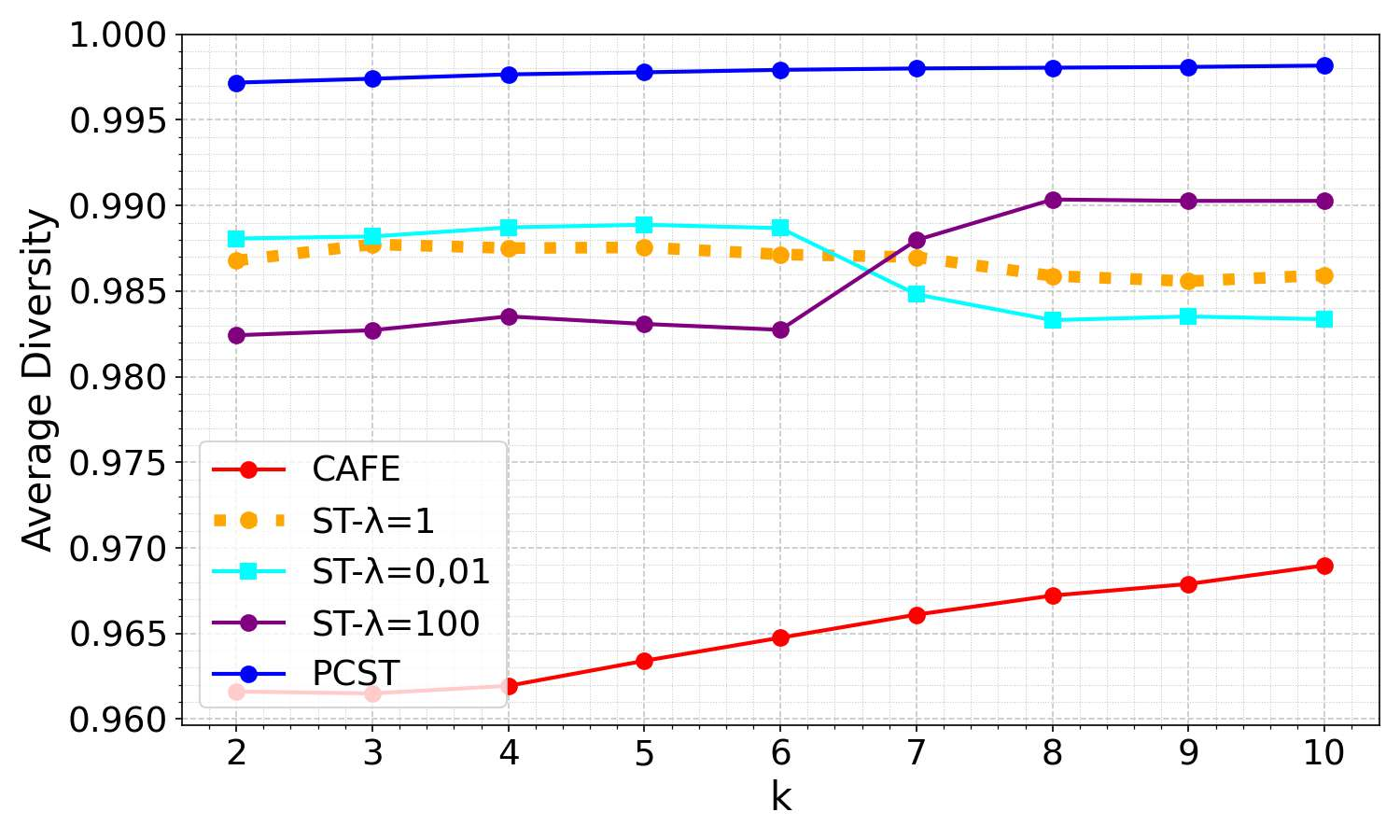}
        \caption{User-group CAFE}
        \label{fig:div_cafe_user_group}
    \end{subfigure}


    \begin{subfigure}{0.24\textwidth} 
        \centering
        \includegraphics[width=\textwidth]{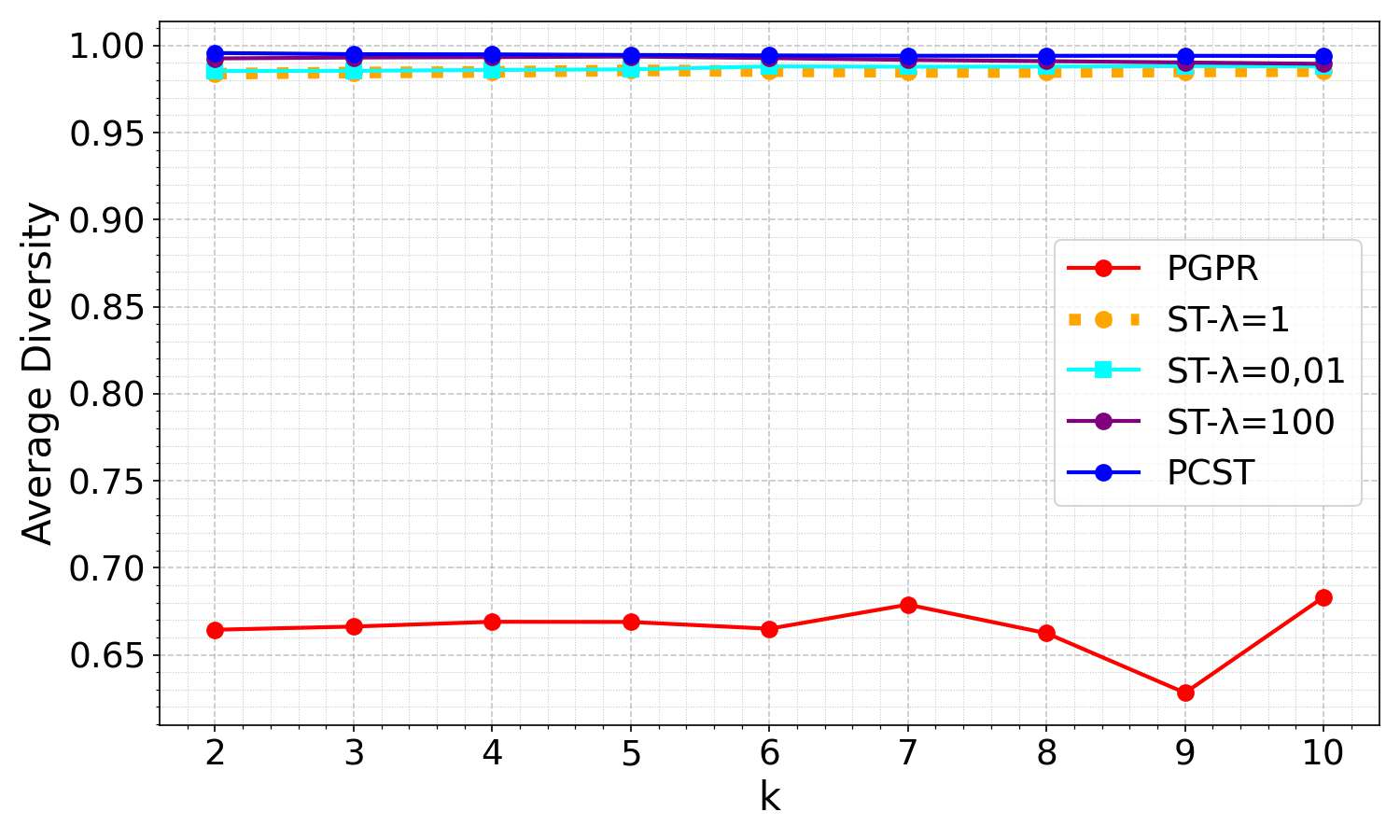}
        \caption{Item-group PGPR}
        \label{fig:div_pgpr_item_group}
    \end{subfigure}
    \hfill
    \begin{subfigure}{0.24\textwidth} 
        \centering
        \includegraphics[width=\textwidth]{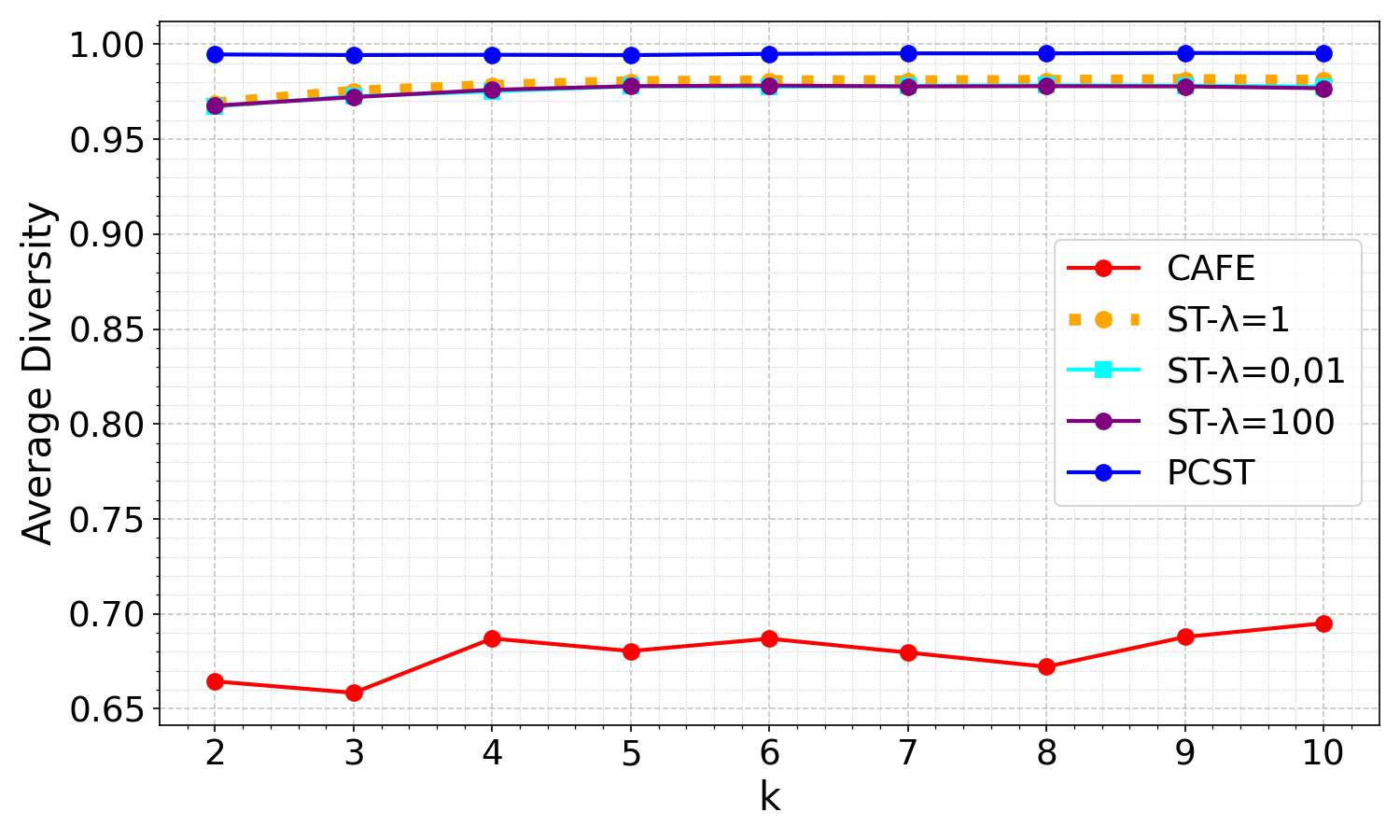}
        \caption{Item-group CAFE}
        \label{fig:div_cafe_item_group}
    \end{subfigure}
    \caption{Diversity}
        \label{fig:div}
\end{figure}

Figure \ref{fig:div} shows that original PGPR and CAFE paths have the lowest diversity due to their fixed 3-hop structure, leading to repetitive explanations. PCST  outperforms ST and baselines because it produces larger, hence more complex summaries resulting in higher diversity.

\subsubsection{Redundancy Metric}
\begin{figure}[ht]
    \centering
    
    \begin{subfigure}{0.24\textwidth} 
        \centering
        \includegraphics[width=\textwidth]{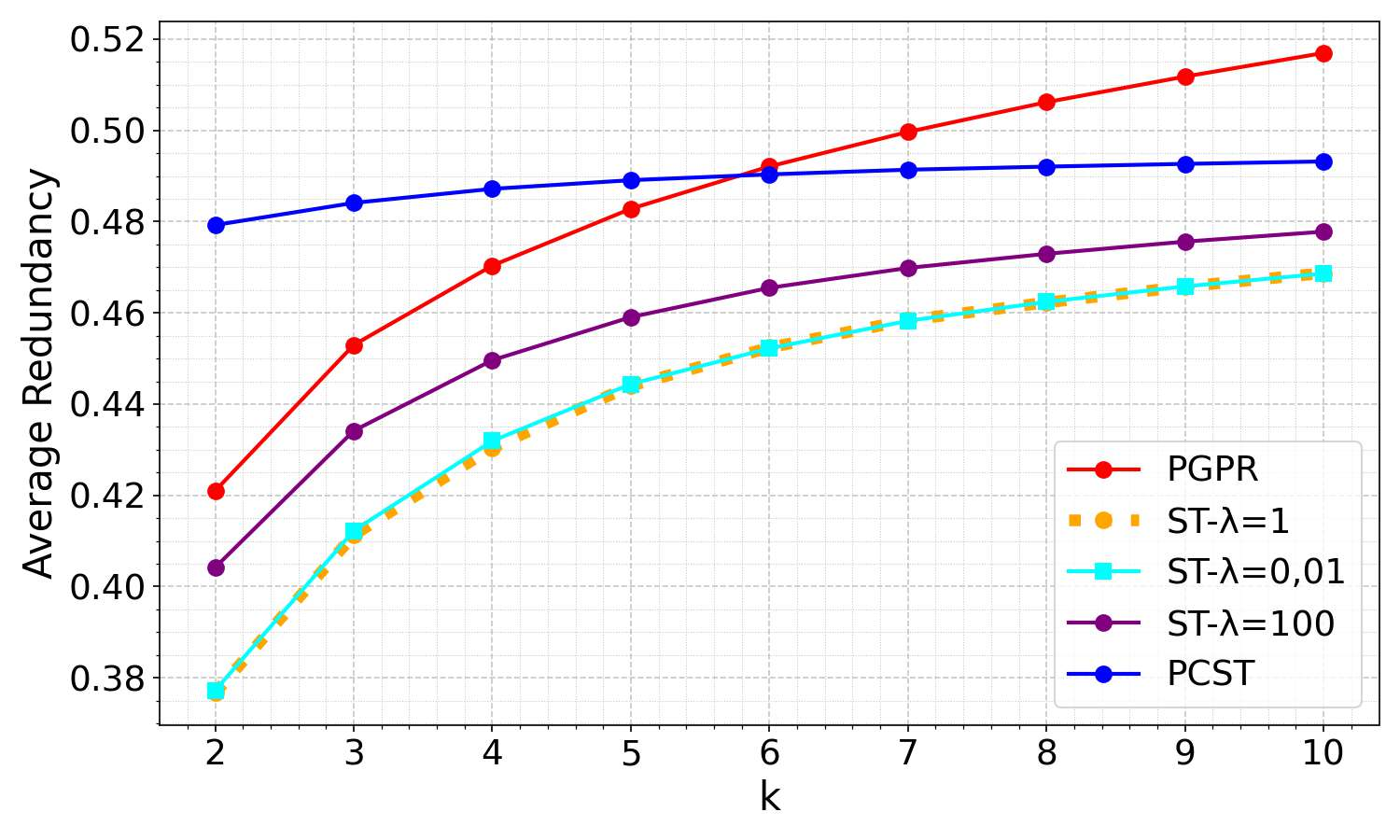}
        \caption{User-centric PGPR}
        \label{fig:red_pgpr_user}
    \end{subfigure}
    \hfill
    \begin{subfigure}{0.24\textwidth} 
        \centering     
        \includegraphics[width=\textwidth]{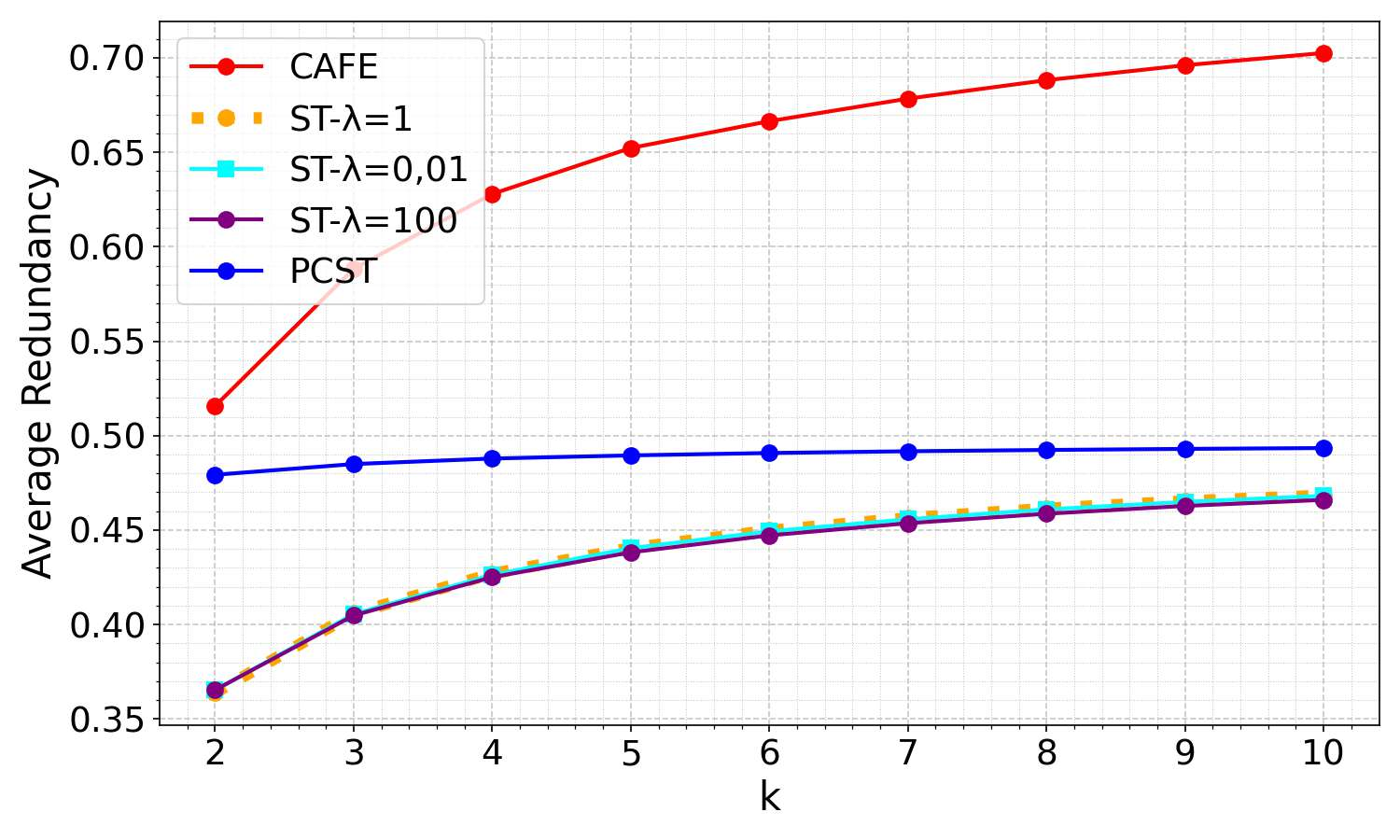}

        \caption{User-centric CAFE}
        \label{fig:red_cafe_user}
    \end{subfigure}


    \begin{subfigure}{0.24\textwidth} 
        \centering
        \includegraphics[width=\textwidth]{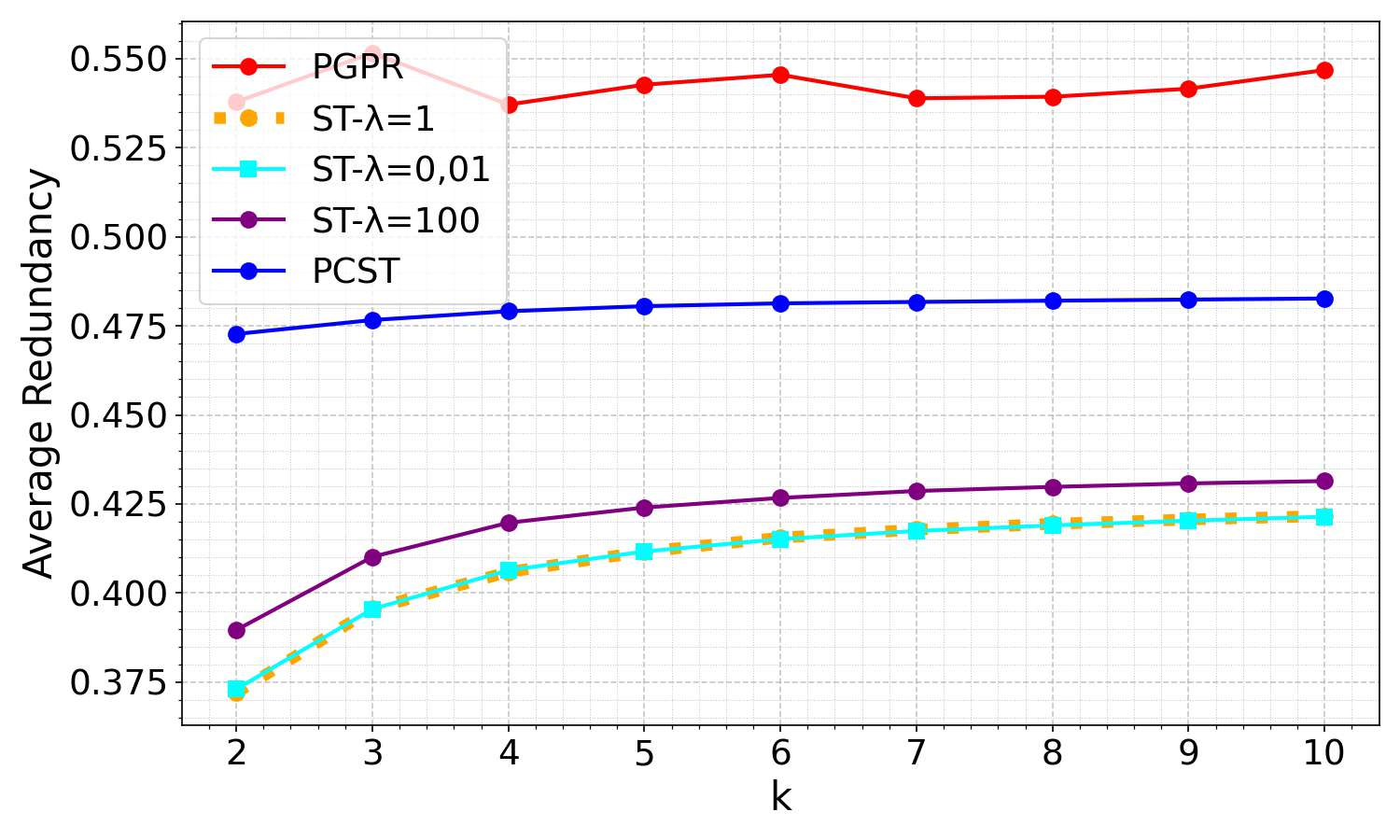}

        \caption{Item-centric PGPR}
        \label{fig:red_pgpr_item}
    \end{subfigure}
    \hfill
    \begin{subfigure}{0.24\textwidth} 
        \centering
        \includegraphics[width=\textwidth]{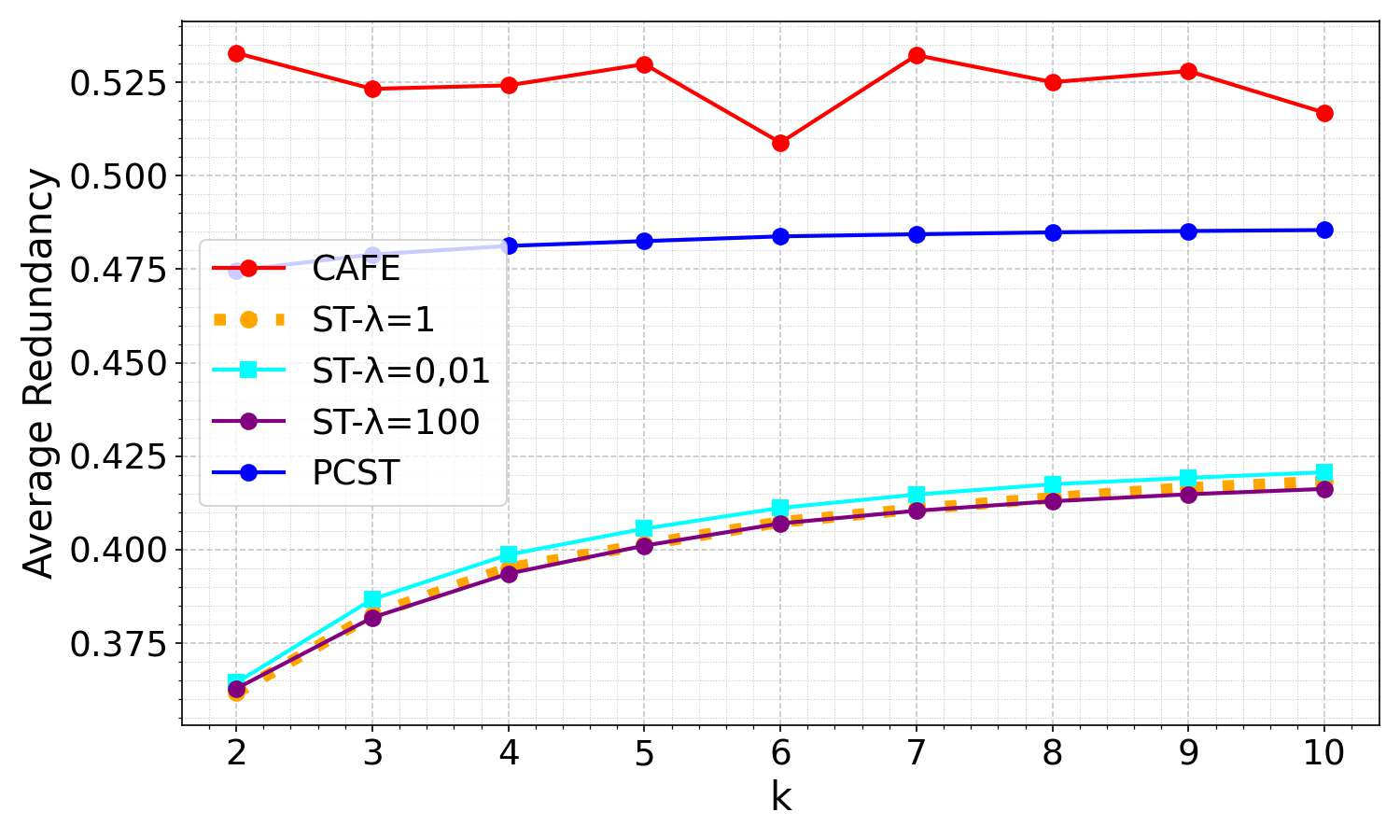}
        \caption{Item-centric CAFE}
        \label{fig:red_cafe_item}
    \end{subfigure}


    \begin{subfigure}{0.24\textwidth} 
        \centering
        \includegraphics[width=\textwidth]{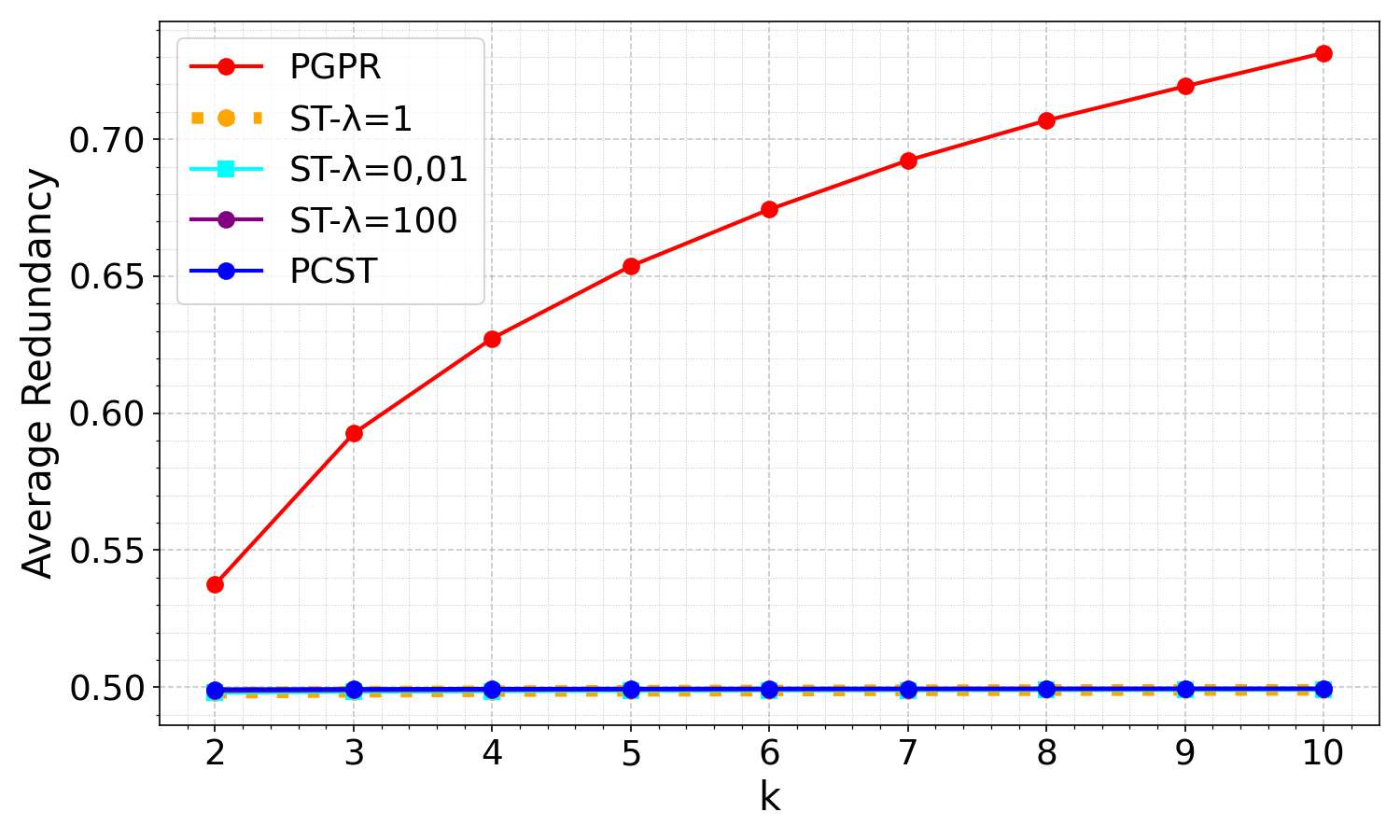}
        \caption{User-group PGPR}
        \label{fig:red_pgpr_user_group}
    \end{subfigure}
    \hfill
    \begin{subfigure}{0.24\textwidth} 
        \centering
        \includegraphics[width=\textwidth]{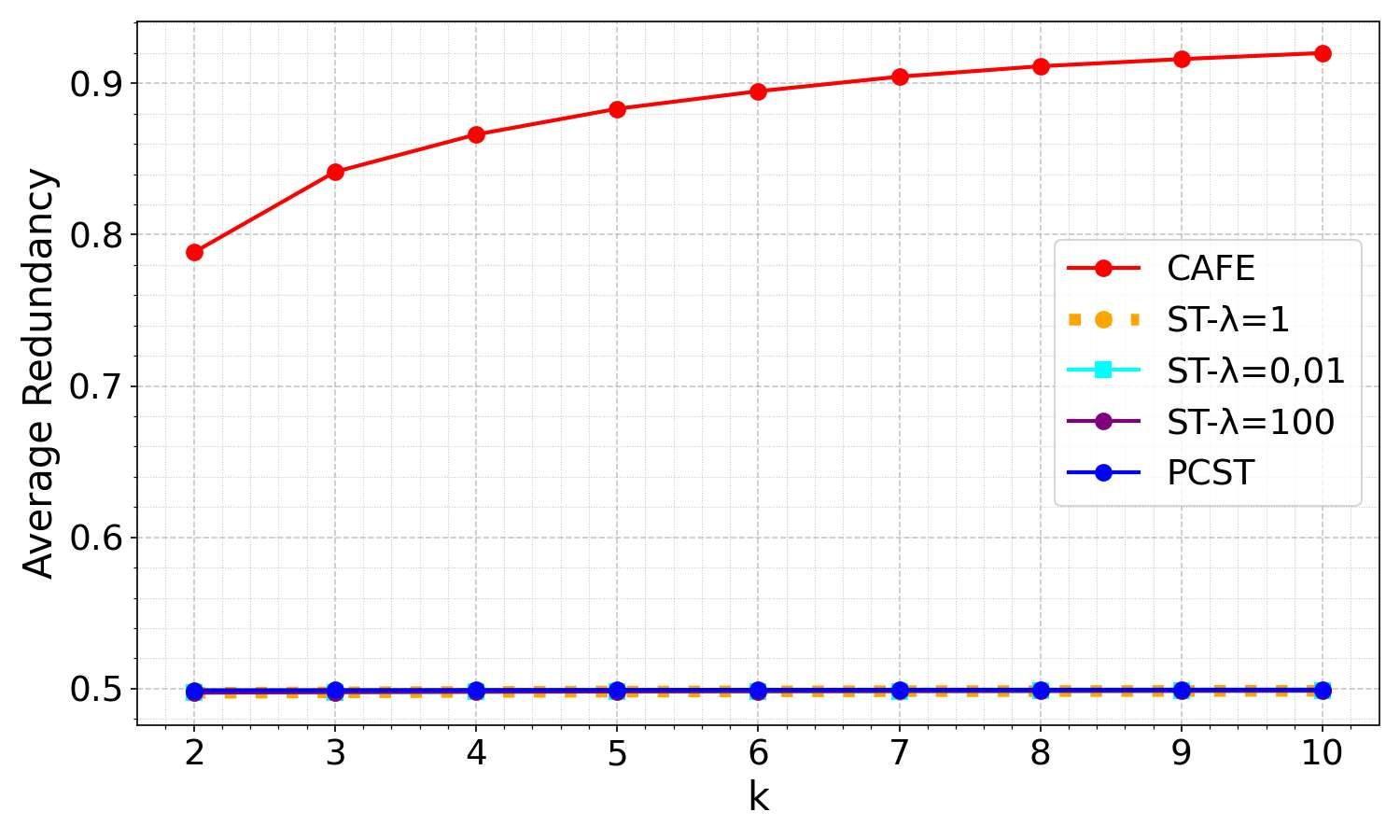}
        \caption{User-group CAFE}
        \label{fig:red_cafe_user_group}
    \end{subfigure}


    \begin{subfigure}{0.24\textwidth} 
        \centering
        \includegraphics[width=\textwidth]{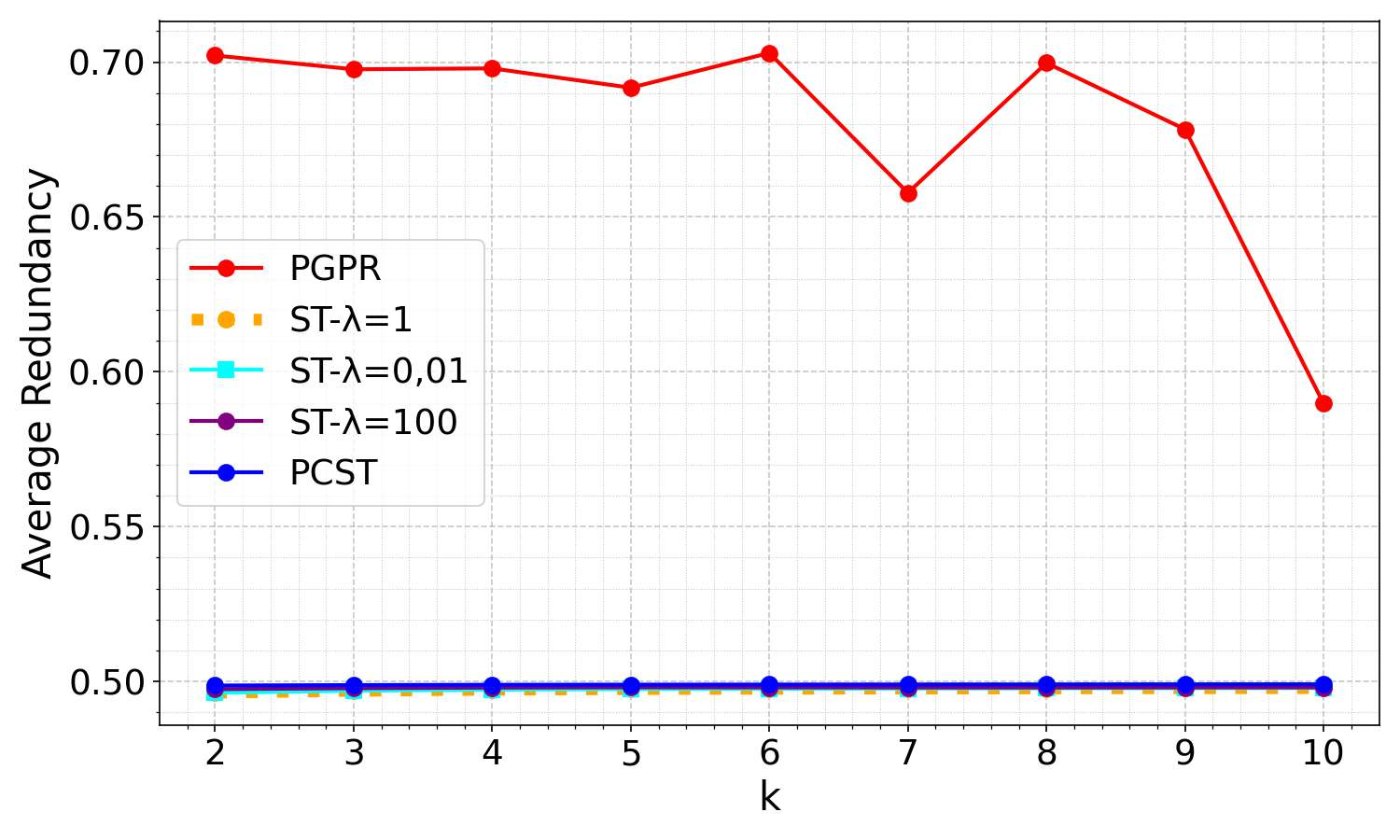}
        \caption{Item-group PGPR}
        \label{fig:red_pgpr_item_group}
    \end{subfigure}
    \hfill
    \begin{subfigure}{0.24\textwidth} 
        \centering
        \includegraphics[width=\textwidth]{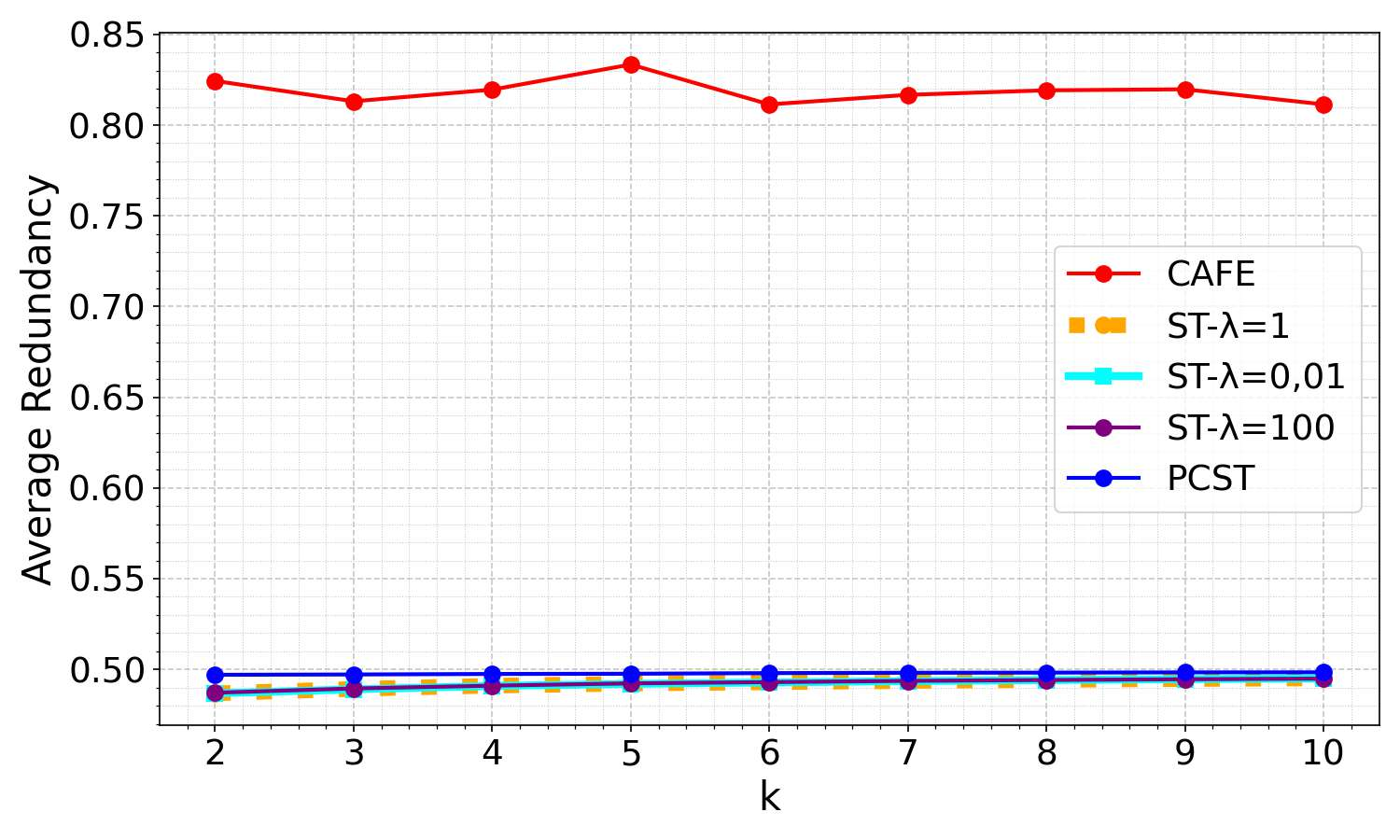}
        \caption{Item-group CAFE}
        \label{fig:red_cafe_item_group}
    \end{subfigure}
    \caption{Redundancy}
    \label{fig:red}
\end{figure}

The redundancy metric, $R(S)$, quantifies duplicate nodes in explanation paths, with higher scores indicating more duplicates and thus less informative explanations. For summary subgraphs, redundancy is defined as the proportion of duplicate nodes within the summary subgraph: $R(S) = \frac{\text{Number of duplicate nodes in } S}{|V_S|}$.
Figure \ref{fig:red} shows that PGPR and CAFE produce repetitive explanations, while PCST and ST yield more efficient summaries with minimal duplication. However, the larger node inclusiveness causes PCST to have higher redundancy than ST, as it often incorporates duplicate nodes. In contrast, ST's optimization for minimal edge weight naturally reduces redundancy.

\subsubsection{Consistency Metric}
\begin{figure}[ht]
    \centering
    
    \begin{subfigure}{0.24\textwidth} 
        \centering
        \includegraphics[width=\textwidth]{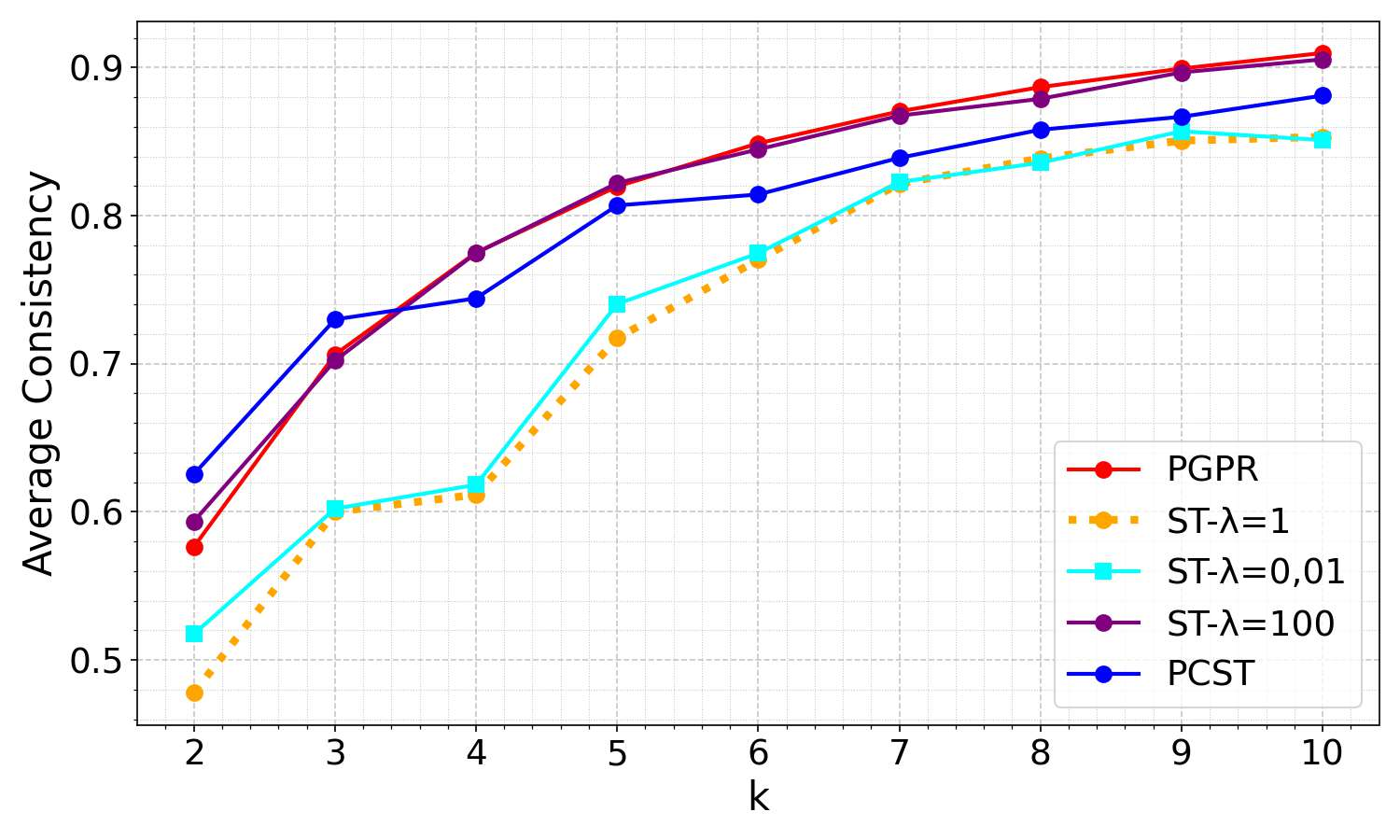}

        \caption{User-centric PGPR}
        \label{fig:con_pgpr_user}
    \end{subfigure}
    \hfill
    \begin{subfigure}{0.24\textwidth} 
        \centering     
        \includegraphics[width=\textwidth]{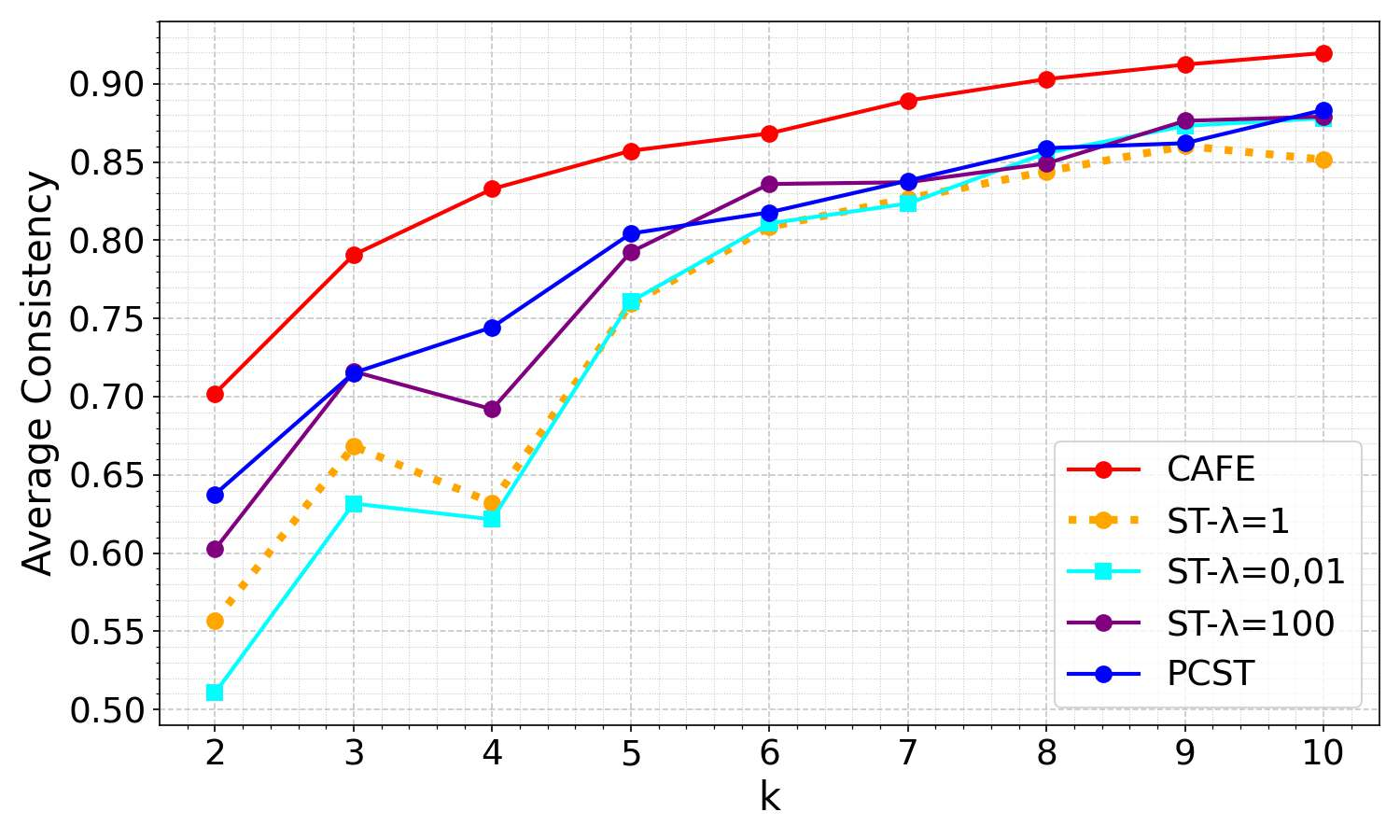}

        \caption{User-centric CAFE}
        \label{fig:con_cafe_user}
    \end{subfigure}


    \begin{subfigure}{0.24\textwidth} 
        \centering
        \includegraphics[width=\textwidth]{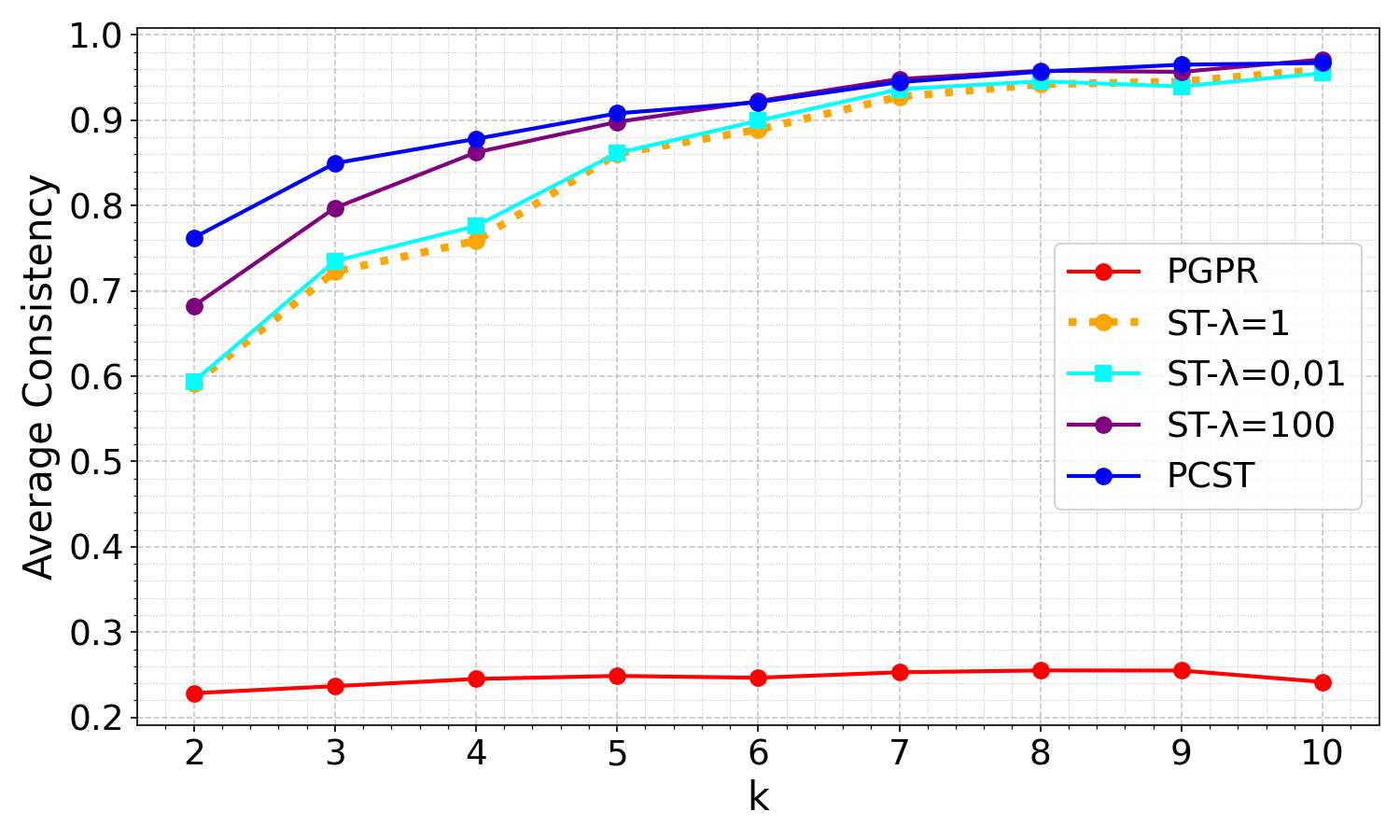}

        \caption{Item-centric PGPR}
        \label{fig:con_pgpr_item}
    \end{subfigure}
    \hfill
    \begin{subfigure}{0.24\textwidth} 
        \centering
        \includegraphics[width=\textwidth]{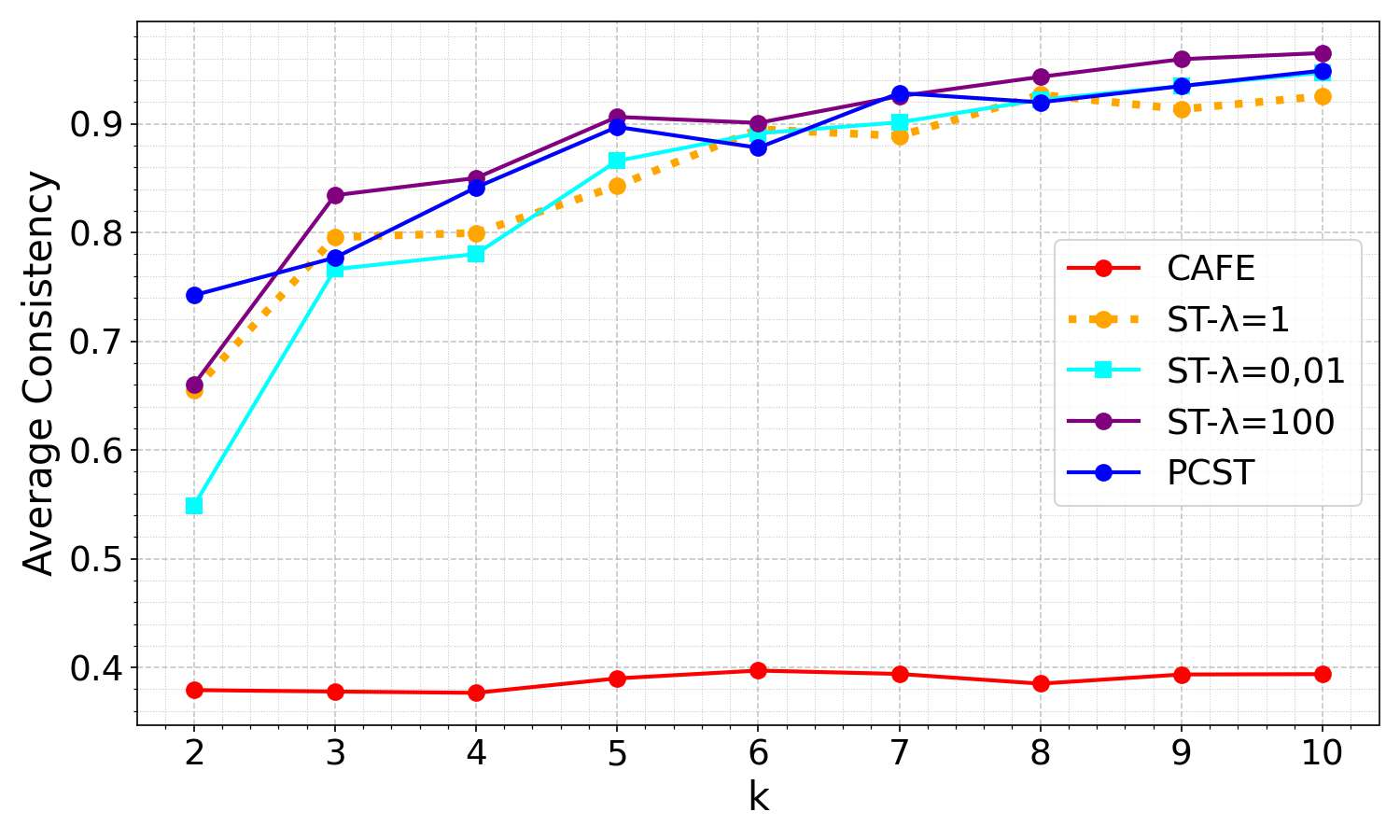}
        \caption{Item-centric CAFE}
        \label{fig:con_cafe_item}
    \end{subfigure}


    \begin{subfigure}{0.24\textwidth} 
        \centering
        \includegraphics[width=\textwidth]{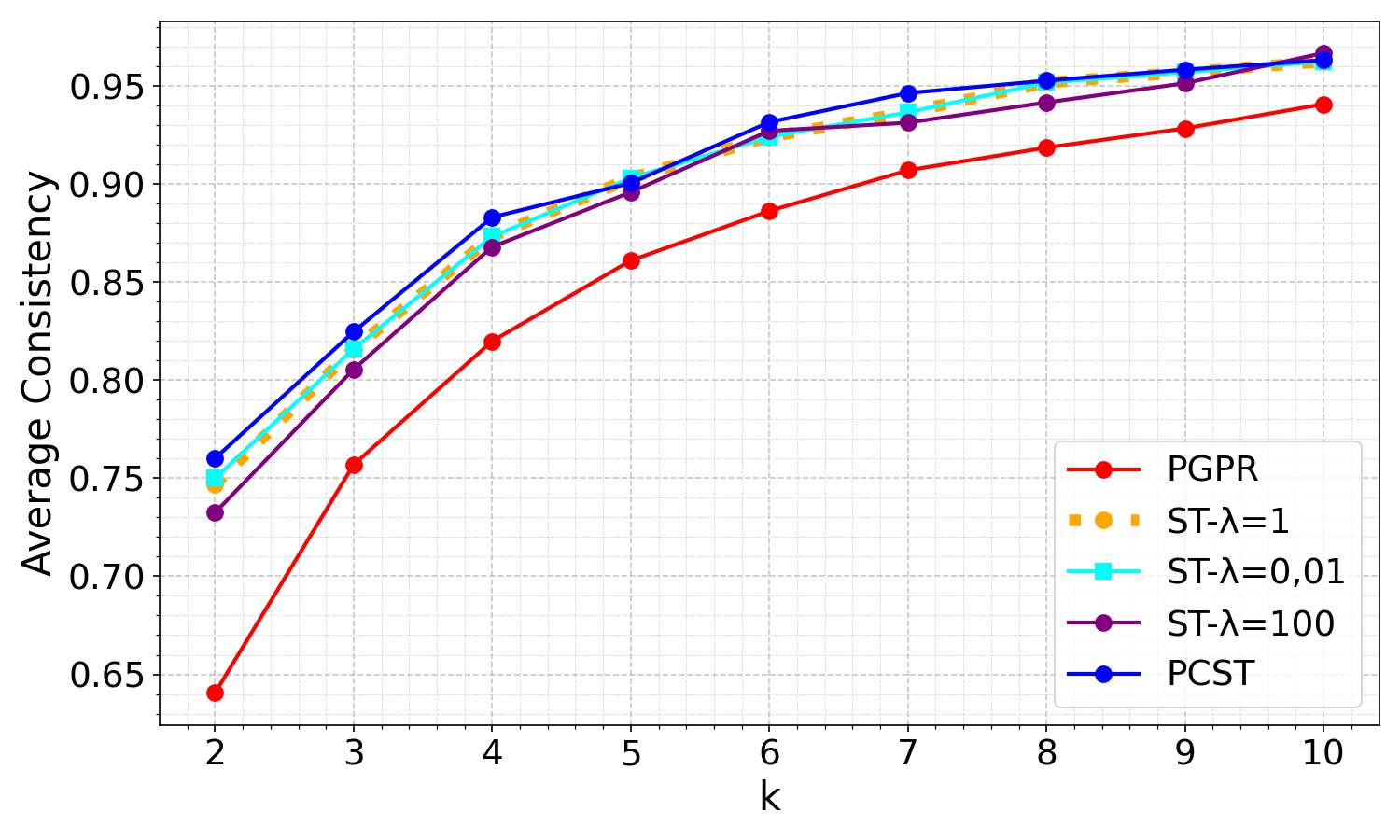}

        \caption{User-group PGPR}
        \label{fig:con_pgpr_user_group}
    \end{subfigure}
    \hfill
    \begin{subfigure}{0.24\textwidth} 
        \centering
        \includegraphics[width=\textwidth]{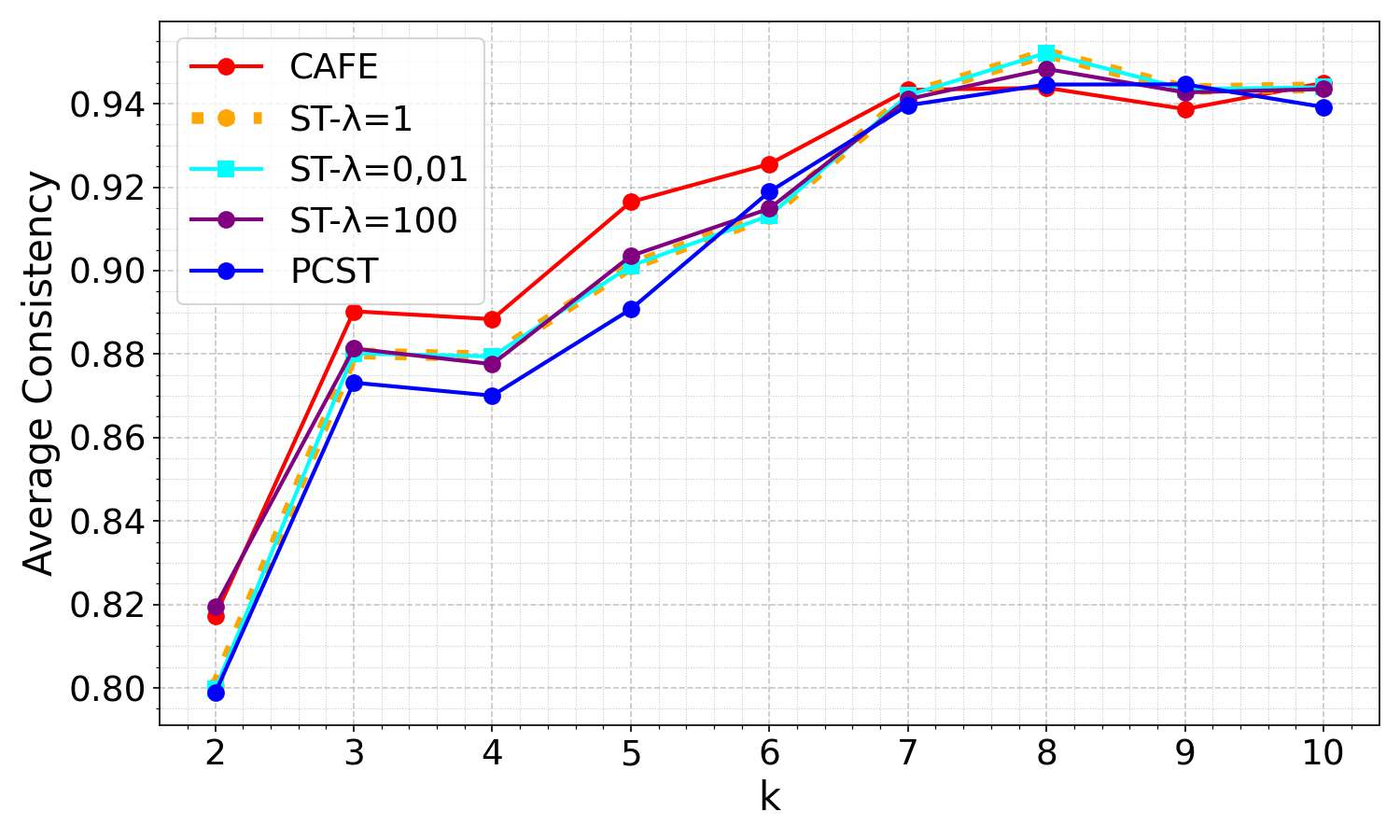}
        \caption{User-group CAFE}
        \label{fig:con_cafe_user_group}
    \end{subfigure}


    \begin{subfigure}{0.24\textwidth} 
        \centering
        \includegraphics[width=\textwidth]{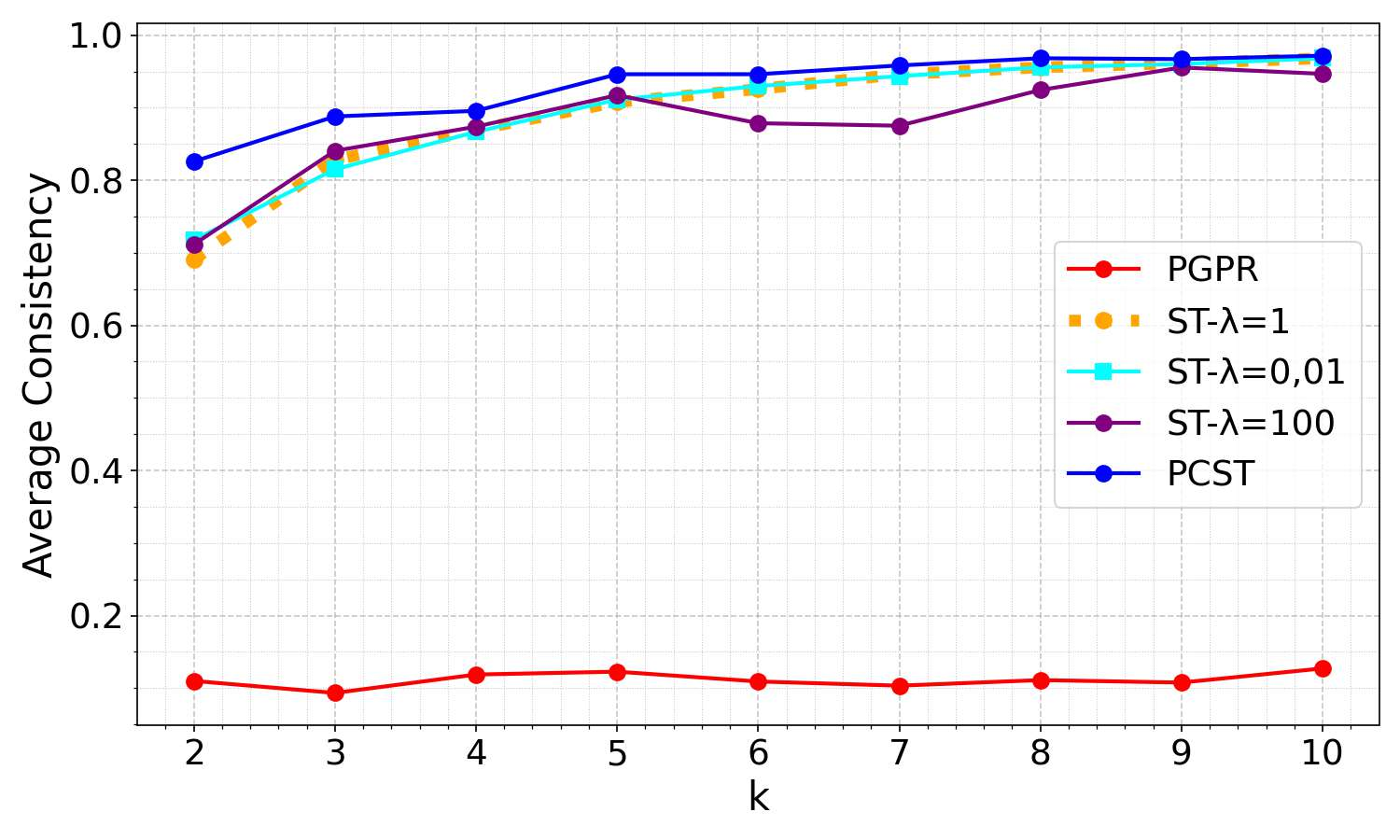}
        \caption{Item-group PGPR}
        \label{fig:con_pgpr_item_group}
    \end{subfigure}
    \hfill
    \begin{subfigure}{0.24\textwidth} 
        \centering
        \includegraphics[width=\textwidth]{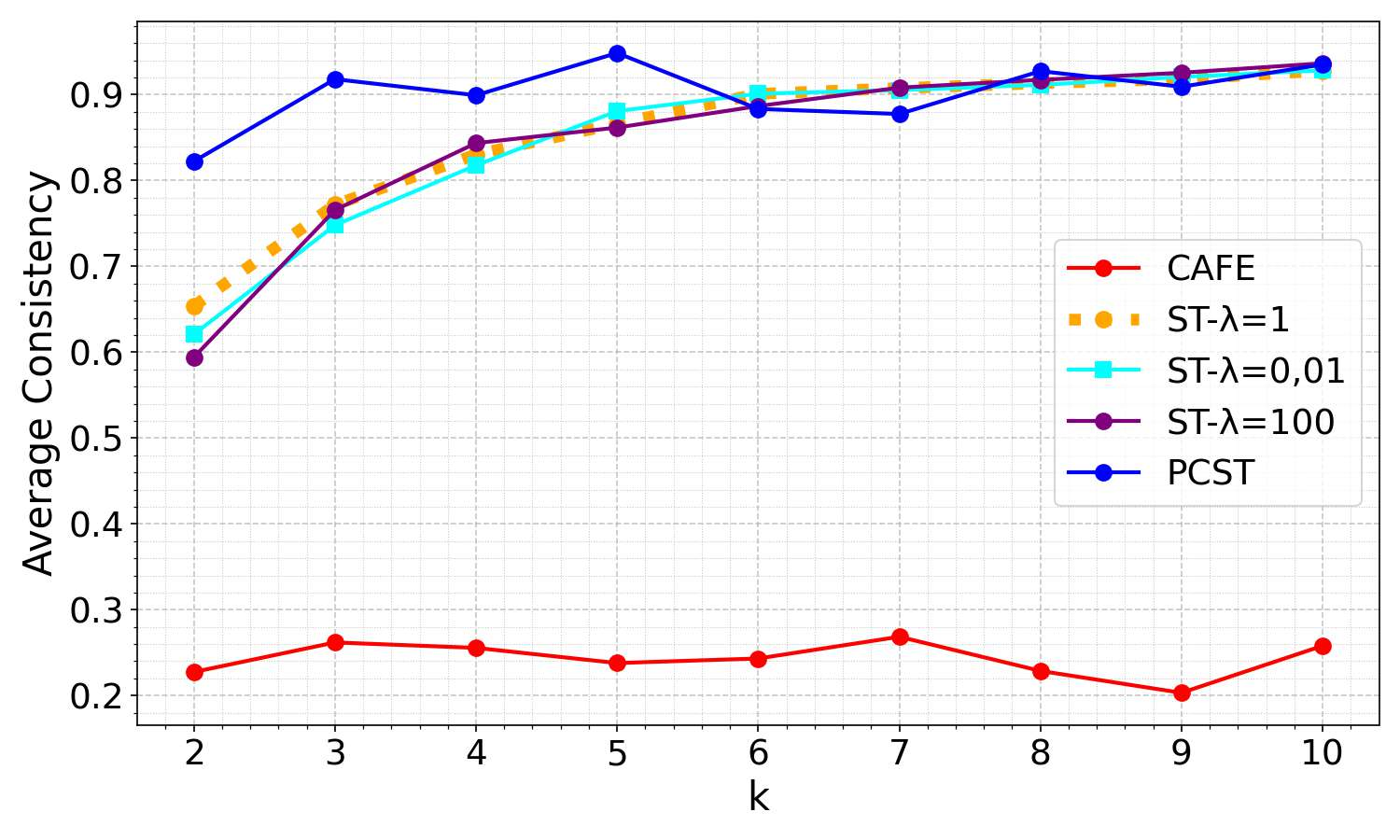}
        \caption{Item-group CAFE}
        \label{fig:con_cafe_item_group}
    \end{subfigure}
    \caption{Consistency}
    \label{fig:con}
\end{figure}

The consistency metric, $C(S)$, quantifies the stability of explanation paths as \( k \) (the number of summarized recommendations) varies. A higher consistency score indicates more stable explanations. Let $S_k$ be the subgraph that summarizes $k$ recommendations. Consistency measures how similar subgraphs \( S_k \) and \( S_{k+1} \) are as \( k \) changes, calculated via Jaccard similarity: $J(S_k, S_{k+1}) = \frac{|V_{S_k} \cap V_{S_{k+1}}|}{|V_{S_k} \cup V_{S_{k+1}}|}$. The overall consistency \( C(S) \) is the average Jaccard similarity across all \( k \) changes: $C(S) = \frac{1}{K-1} \sum_{k=1}^{K-1} J(S_k, S_{k+1})$, where \( K \) is the number of recommendations summarized by $S$.
Figure \ref{fig:con} shows that in user-centric scenarios, baselines outperform ST and PCST in consistency due to their stable, fixed 3-hop paths tied to user-item interactions. ST and PCST maintain high consistency across scenarios due to their algorithmic stability in forming knowledge graph connections: ST minimally extends the tree with the necessary edges to connect one additional terminal node with each \( k\) increment, while PCST adjusts only the node's prize, preserving structural coherence as \( k\) changes.

\subsubsection{Relevance Metric}
\begin{figure}[ht]
    \centering   
    \begin{subfigure}{0.24\textwidth}
        \centering
        \includegraphics[width=\textwidth]{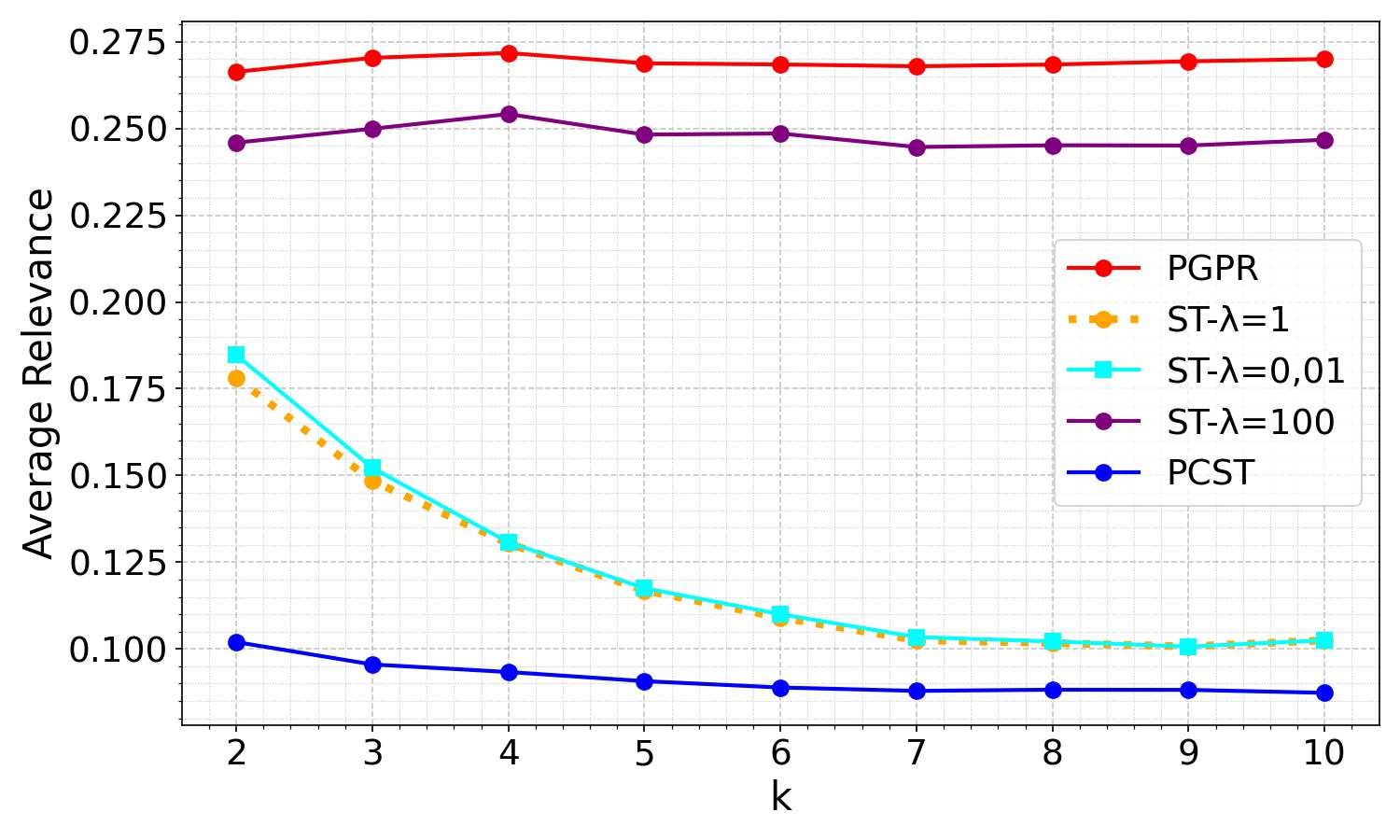}
        \caption{User-centric PGPR}
        \label{fig:rel_pgpr_user}
    \end{subfigure}
    \hfill
    \begin{subfigure}{0.24\textwidth} 
        \centering
        \includegraphics[width=\textwidth]{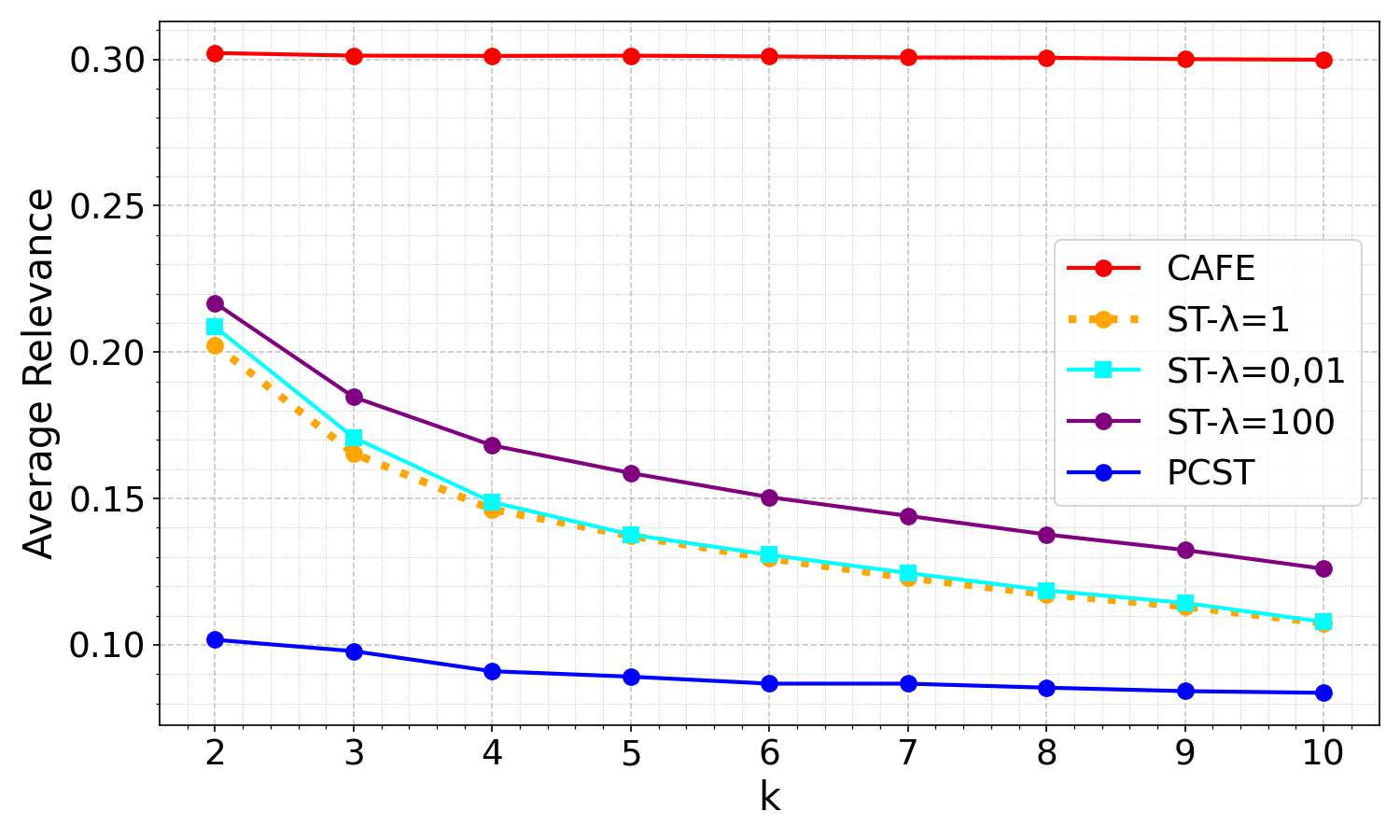}
        \caption{User-centric CAFE}
        \label{fig:rel_cafe_user}
    \end{subfigure}
    \begin{subfigure}{0.24\textwidth}
        \centering
        \includegraphics[width=\textwidth]{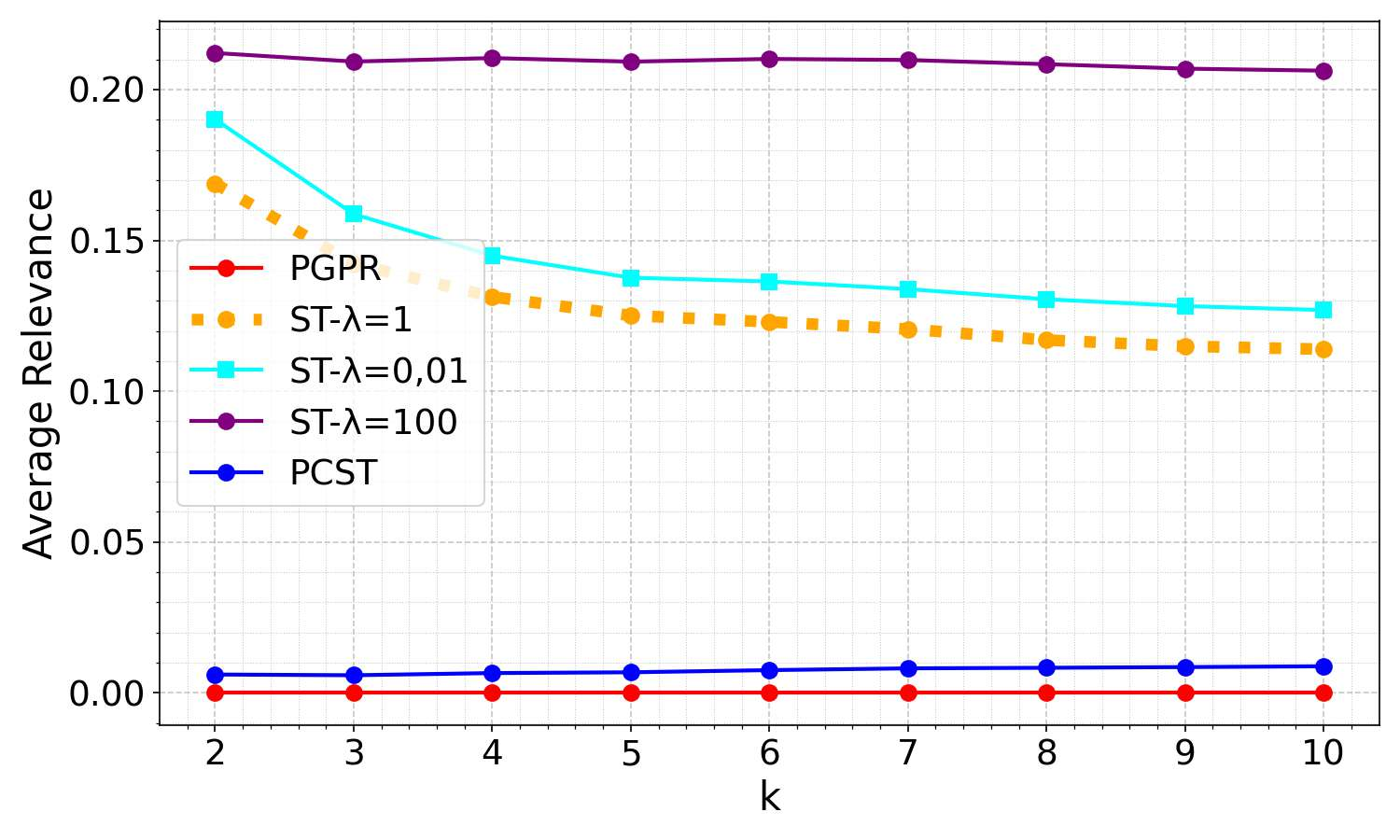}
        \caption{Item-centric PGPR}
        \label{fig:rel_pgpr_item}
    \end{subfigure}
    \hfill
    \begin{subfigure}{0.24\textwidth}
        \centering
        \includegraphics[width=\textwidth]{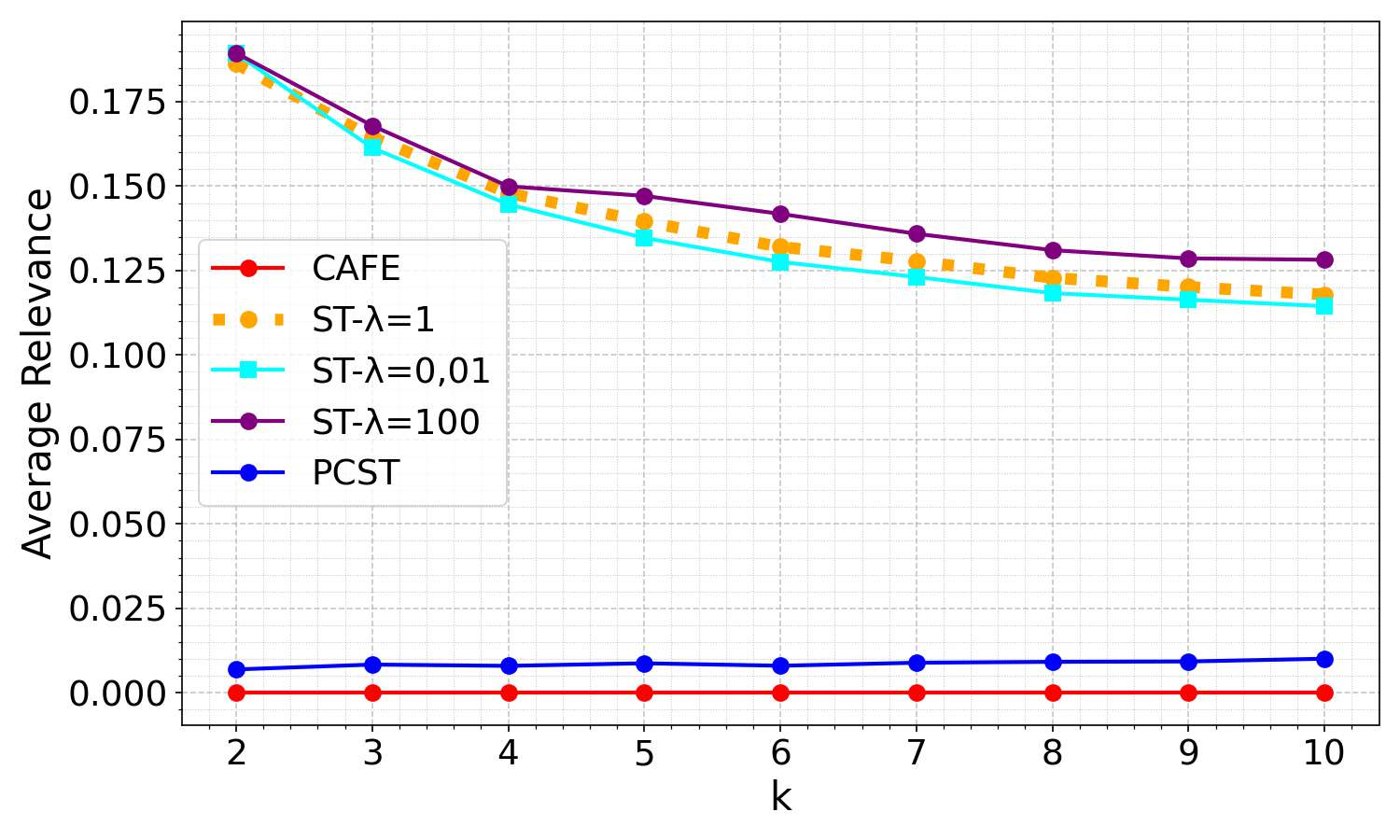}
        \caption{Item-centric CAFE}
        \label{fig:rel_cafe_item}
    \end{subfigure}
    \begin{subfigure}{0.24\textwidth}
        \centering
        \includegraphics[width=\textwidth]{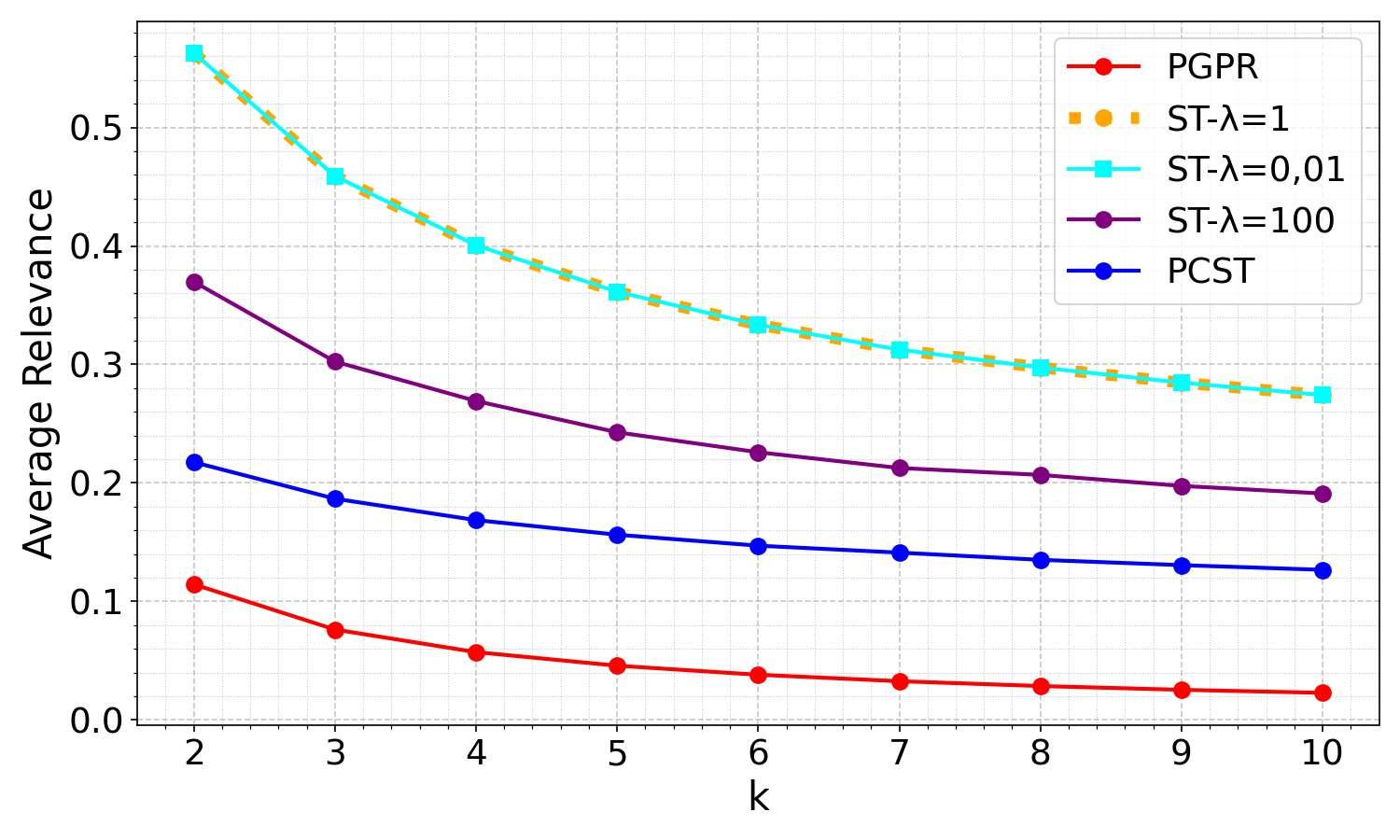}
        \caption{User-group PGPR}
        \label{fig:rel_pgpr_user_group}
    \end{subfigure}
    \hfill
    \begin{subfigure}{0.24\textwidth}
        \centering
        \includegraphics[width=\textwidth]{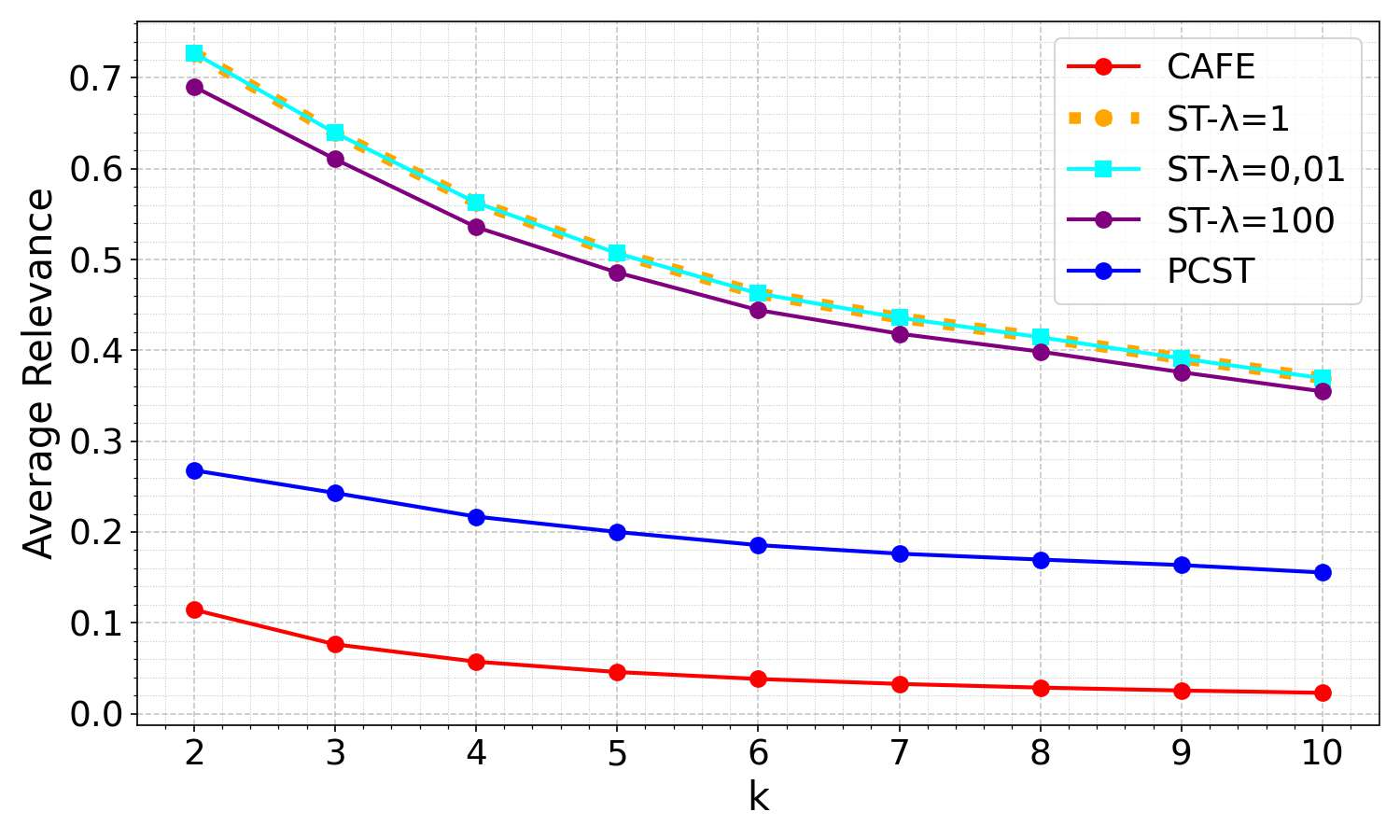}
        \caption{User-group CAFE}
        \label{fig:rel_cafe_user_group}
    \end{subfigure}
    \begin{subfigure}{0.24\textwidth} 
        \centering
        \includegraphics[width=\textwidth]{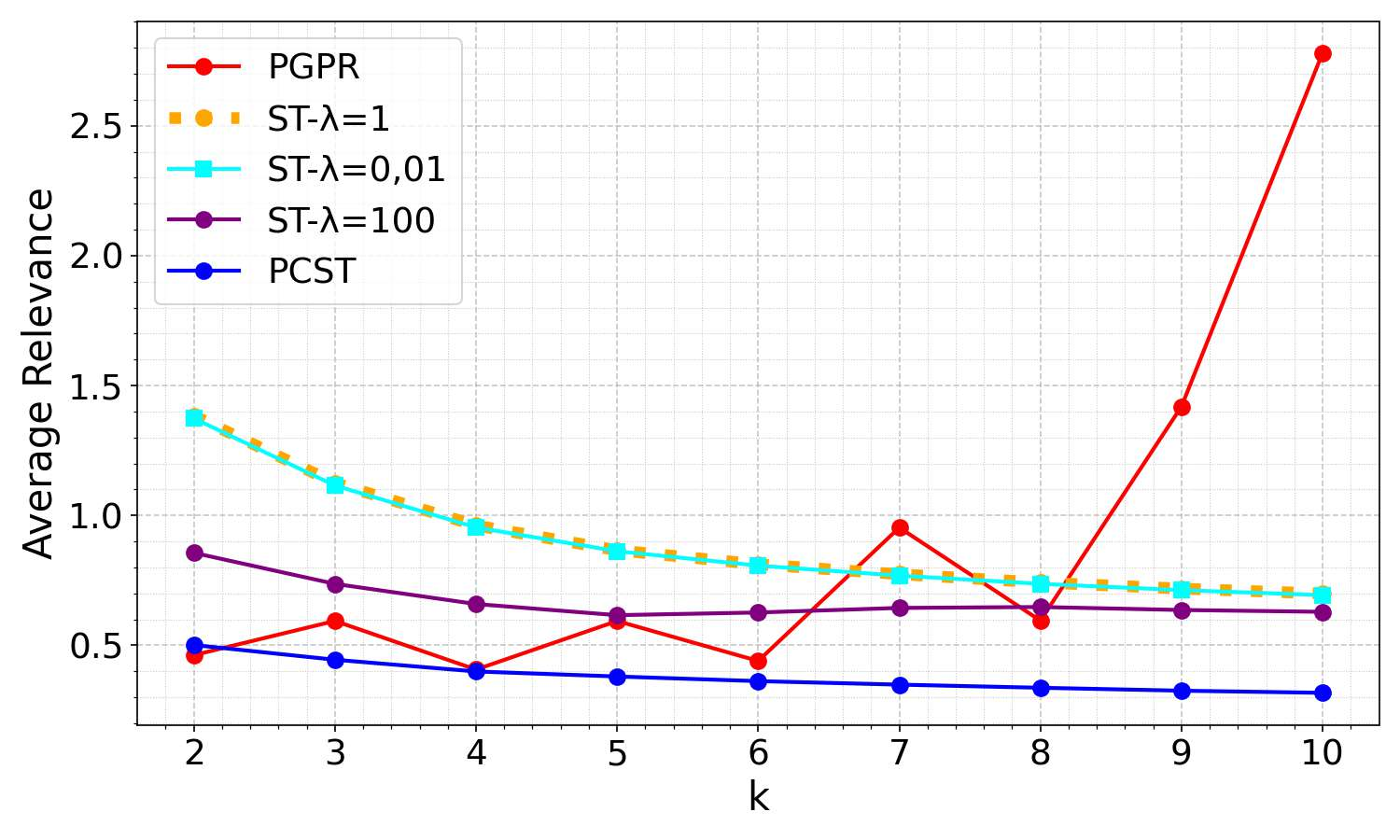}
        \caption{Item-group PGPR}
        \label{fig:rel_pgpr_item_group}
    \end{subfigure}
    \hfill
    \begin{subfigure}{0.24\textwidth}
        \centering
        \includegraphics[width=\textwidth]{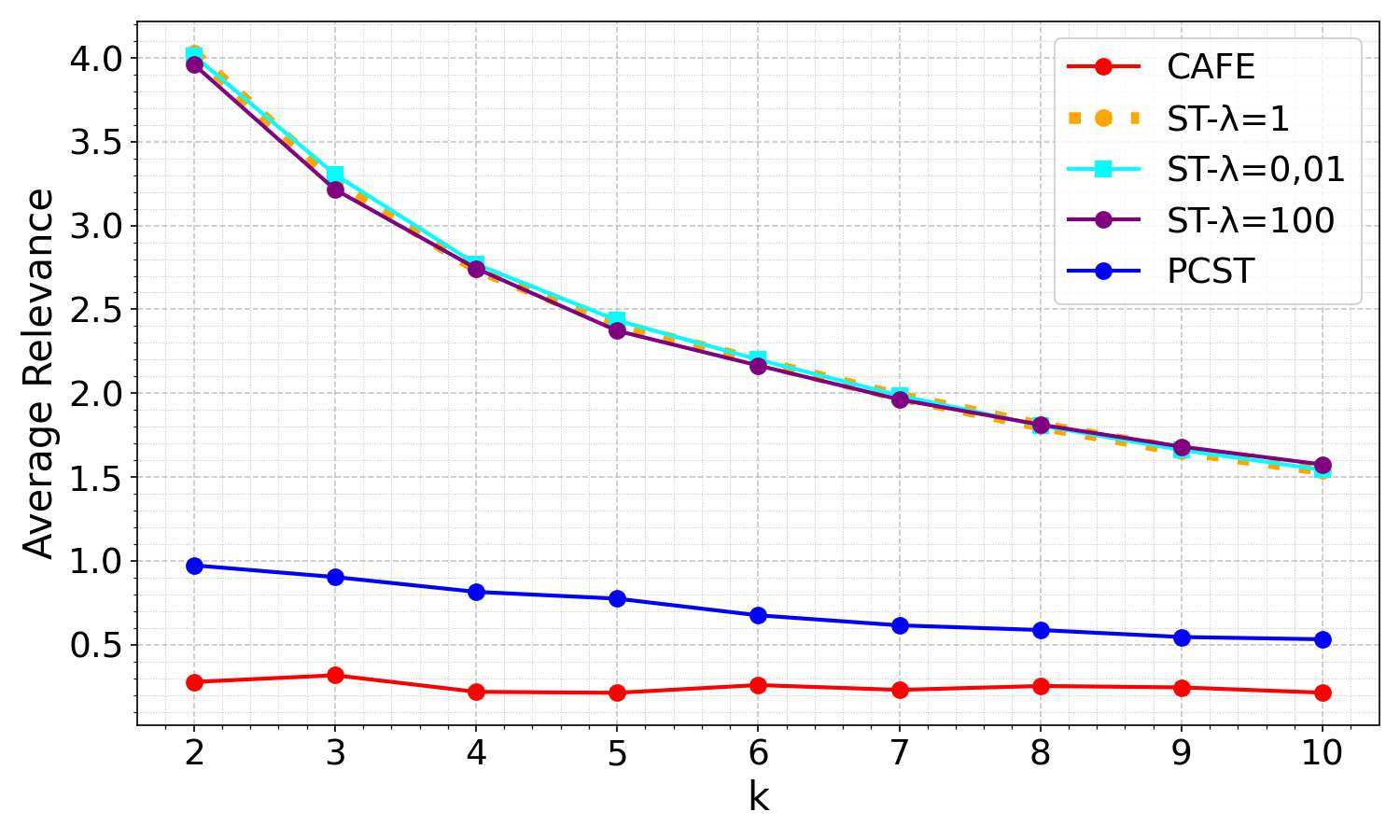}
        \caption{Item-group CAFE}
        \label{fig:rel_cafe_item_group}
    \end{subfigure}
    \caption{Relevance}
    \label{fig:rel}
\end{figure}

The relevance metric, $R(S)$, quantifies how well explanation paths align with user preferences, based on edge weights from historical user-item interactions. Relevance for an explanation is the total weight of its paths. In summary explanations, relevance is the total edge weight in the summary subgraph \( S \), reflecting connection strength based on \( G_M \) interactions: $R(S) = \sum_{e \in E_S} w_M(e).$
Figure \ref{fig:rel} shows that PGPR and CAFE provide the most relevant explanations in user-centric scenarios by prioritizing user-item interaction history. In other scenarios, ST and PCST consistently achieve higher relevance. ST’s relevance improves as \(\lambda\) increases, by including more user-item interactions which have larger weights. PCST, despite ignoring edge weights, often surpasses baselines by generating larger summaries that aggregate more total \( w_M \) weights.

\subsubsection{Privacy Metric}
\begin{figure}[ht]
    \centering   
    \begin{subfigure}{0.24\textwidth} 
        \centering
        \includegraphics[width=\textwidth]{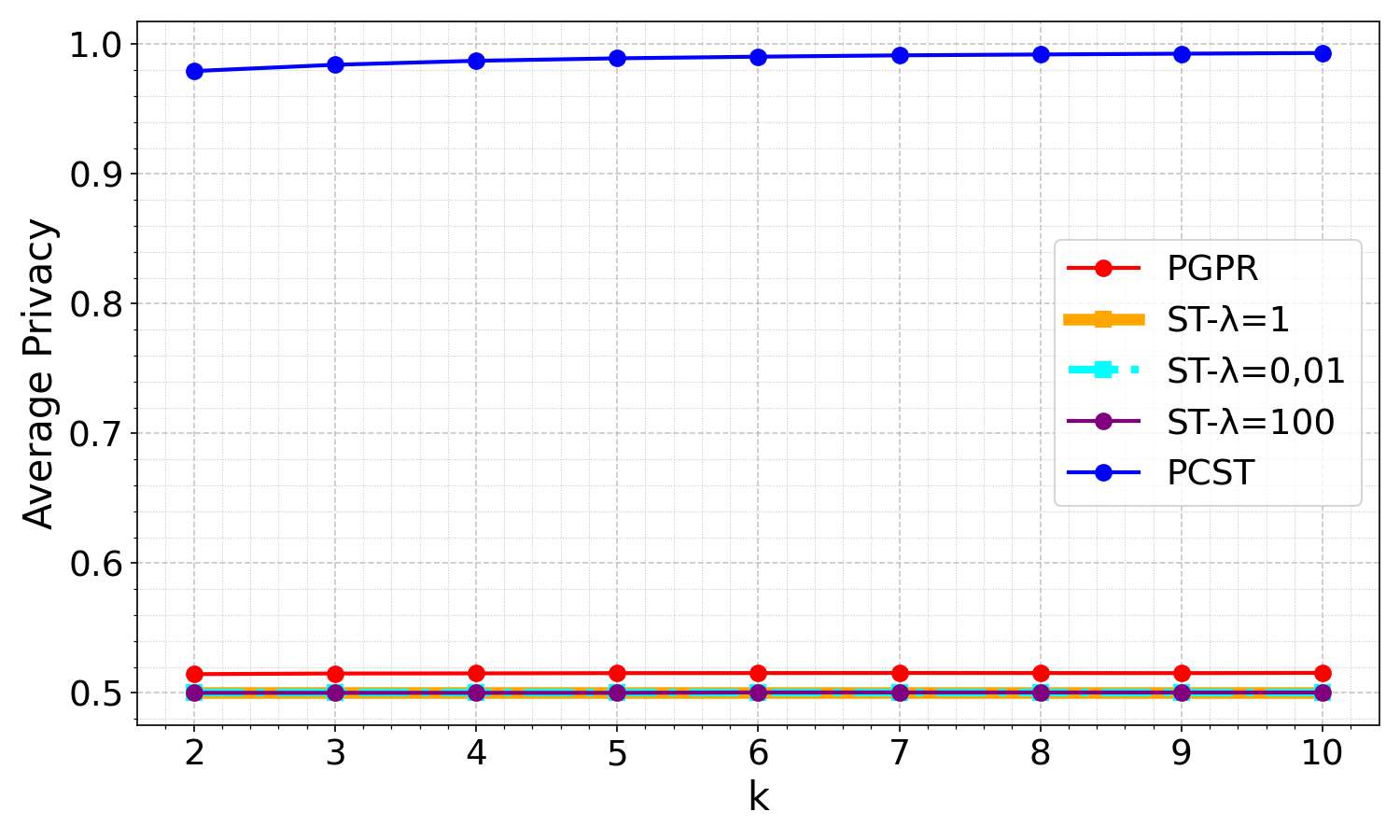}
        \caption{User-centric PGPR}
        \label{fig:priv_pgpr_user}
    \end{subfigure}
    \hfill
    \begin{subfigure}{0.24\textwidth}
        \centering
        \includegraphics[width=\textwidth]{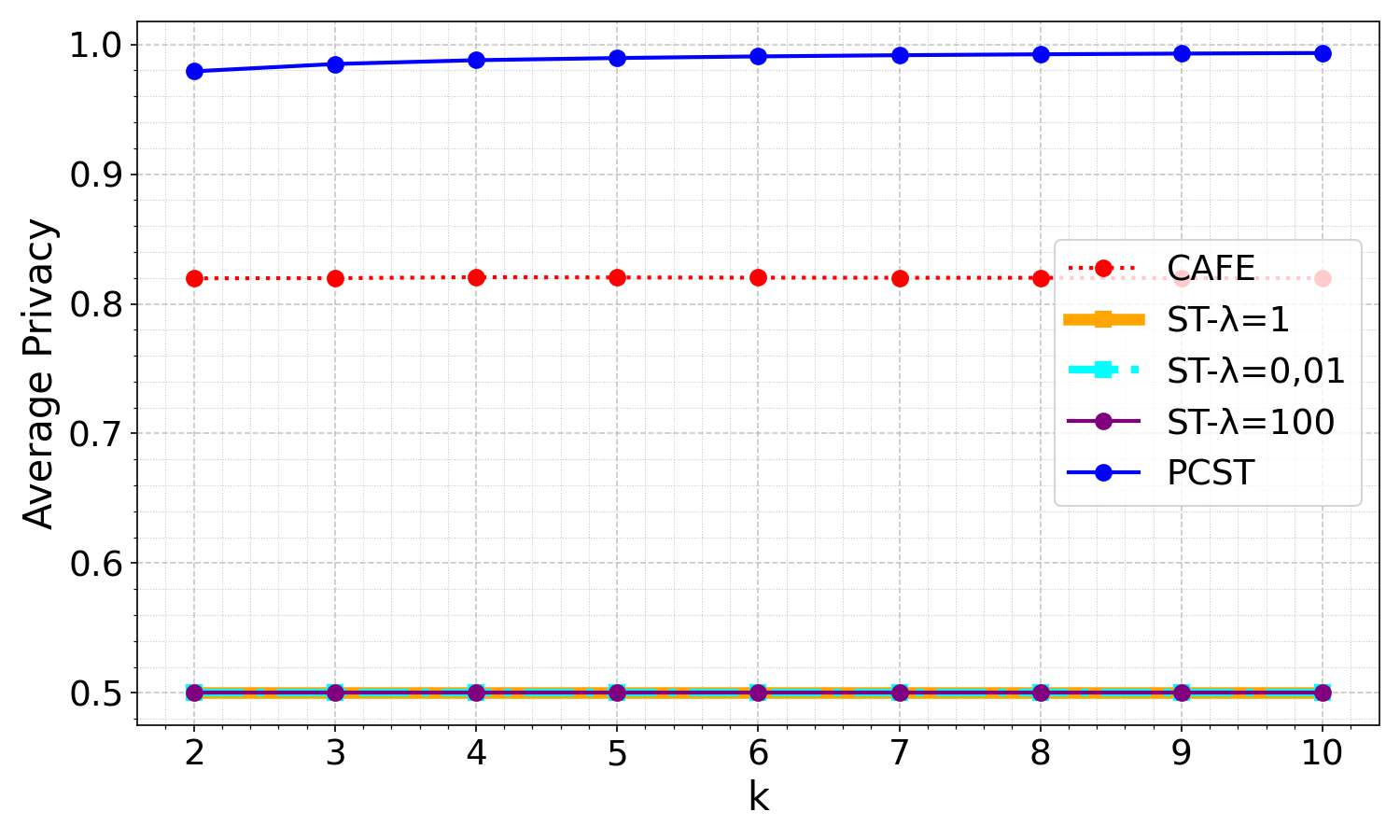}
        \caption{User-centric CAFE}
        \label{fig:priv_cafe_user}
    \end{subfigure}
    \begin{subfigure}{0.24\textwidth}
        \centering
        \includegraphics[width=\textwidth]{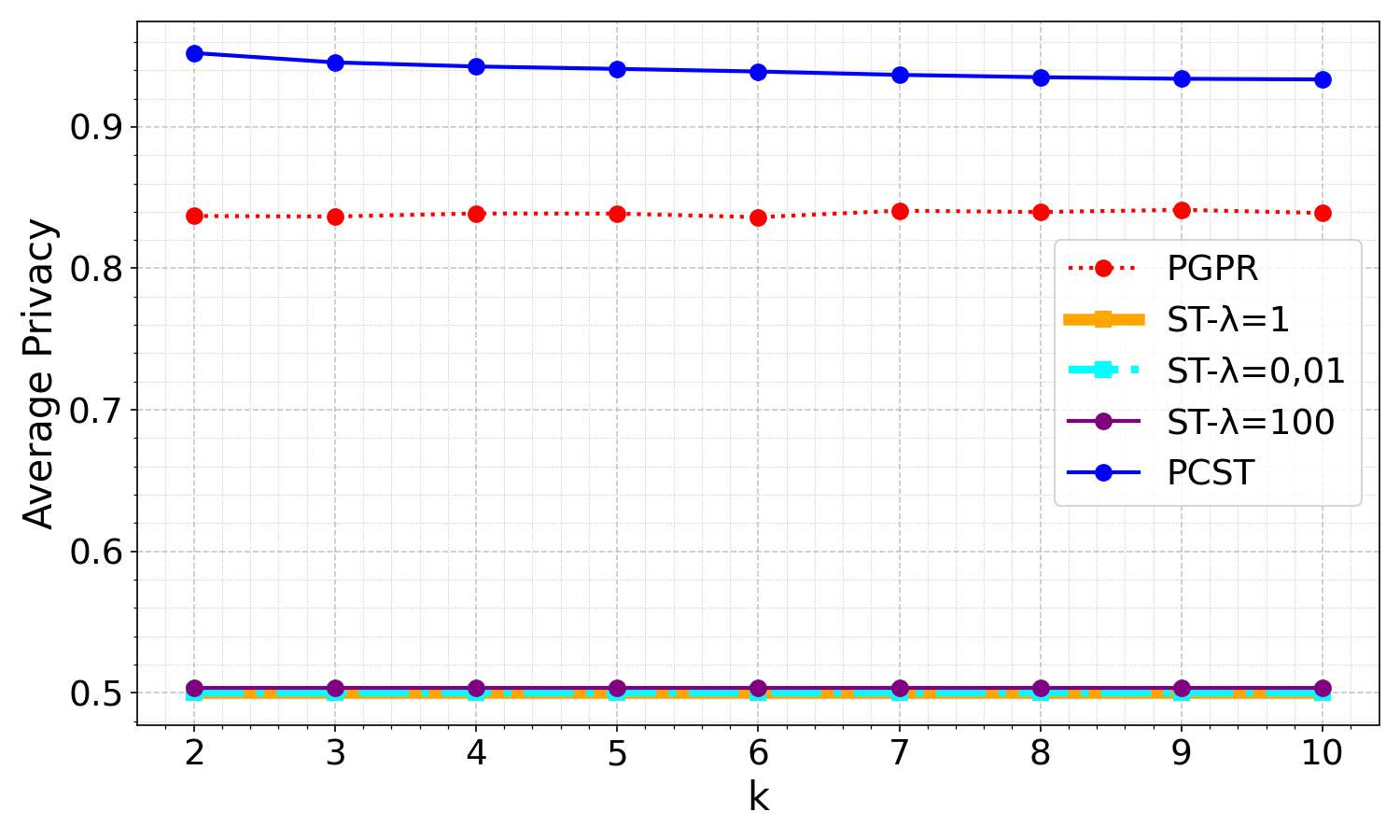}
        \caption{Item-centric PGPR}
        \label{fig:priv_pgpr_item}
    \end{subfigure}
    \hfill
    \begin{subfigure}{0.24\textwidth}
        \centering
        \includegraphics[width=\textwidth]{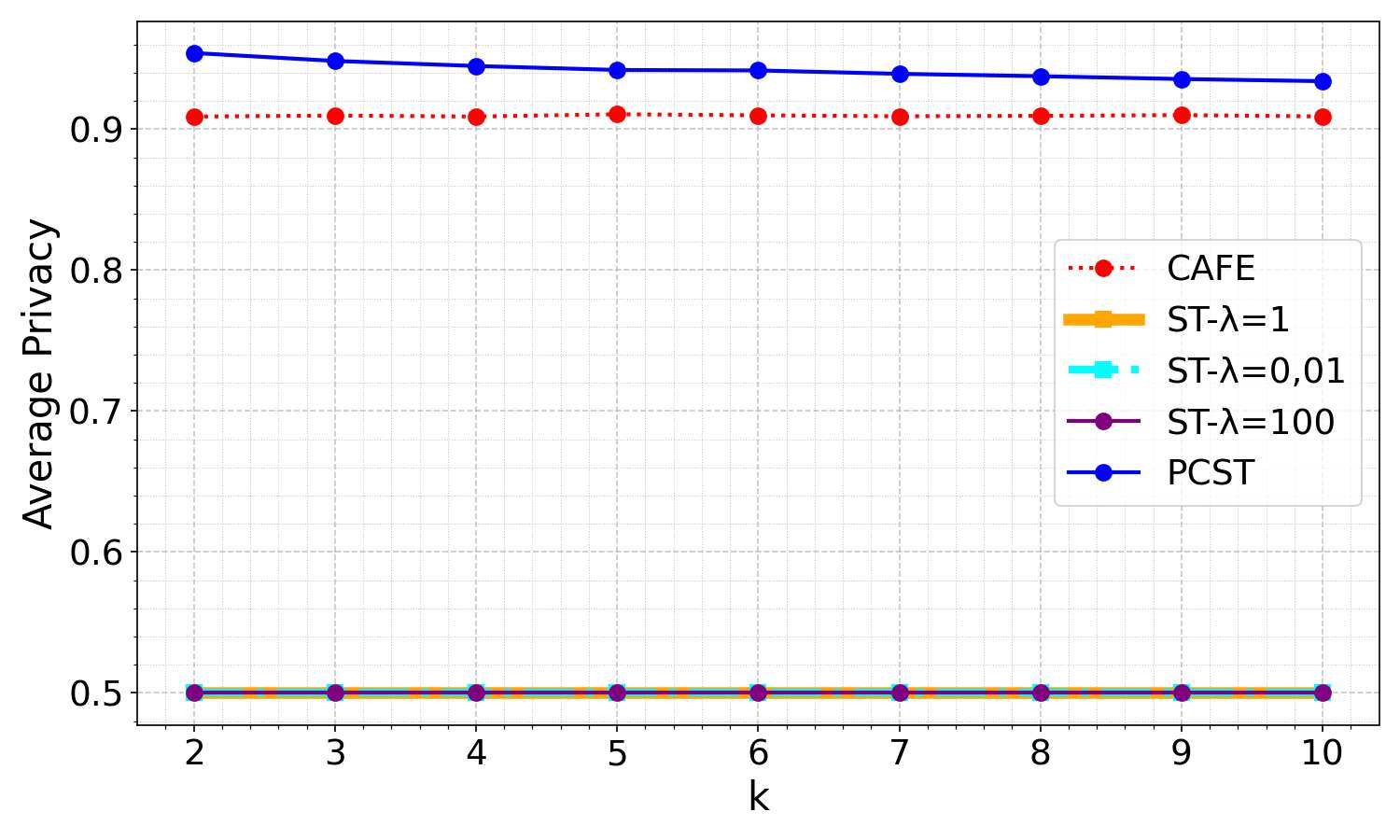}
        \caption{Item-centric CAFE}
        \label{fig:priv_cafe_item}
    \end{subfigure}
    \begin{subfigure}{0.24\textwidth} 
        \centering
        \includegraphics[width=\textwidth]{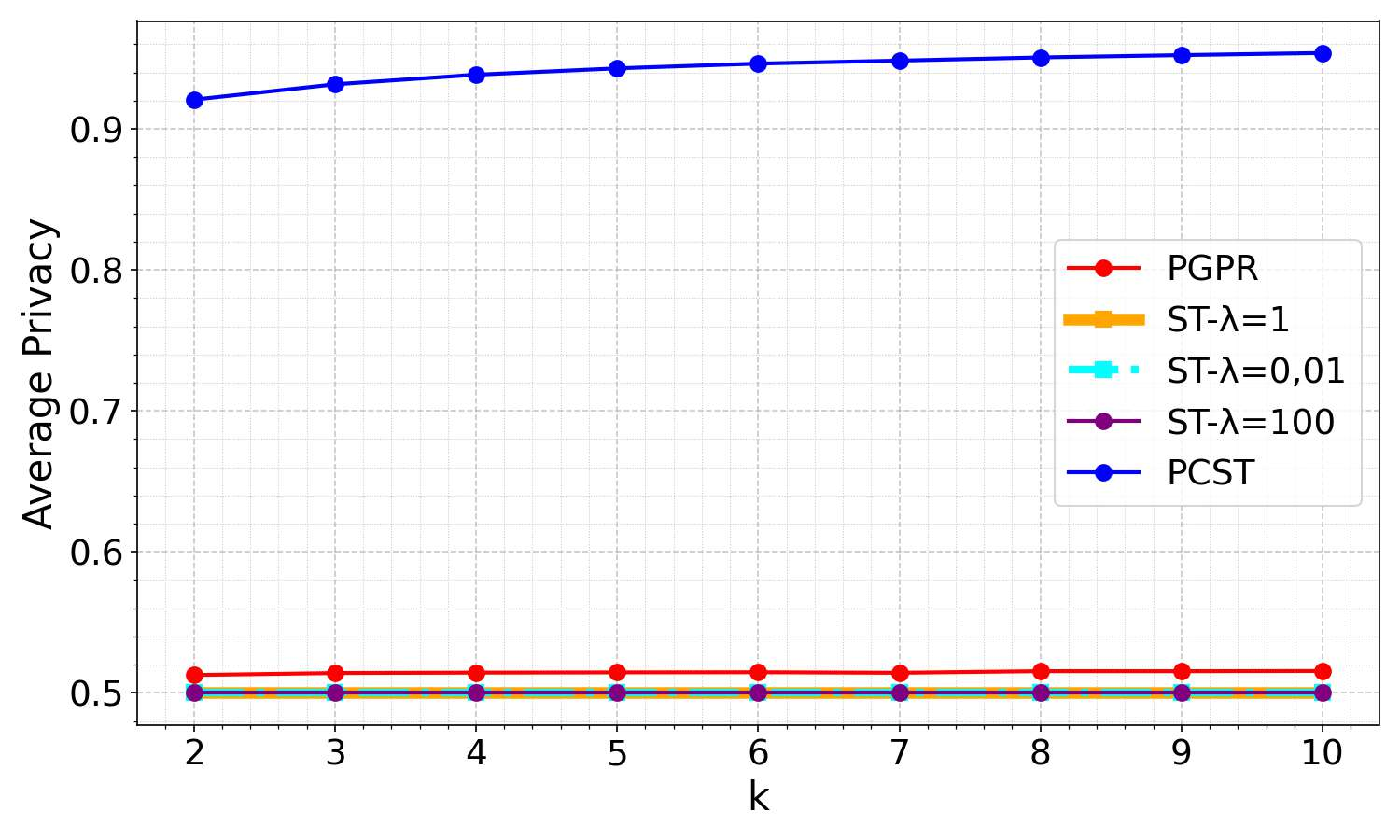}
        \caption{User-group PGPR}
        \label{fig:priv_pgpr_user_group}
    \end{subfigure}
    \hfill
    \begin{subfigure}{0.24\textwidth} 
        \centering
        \includegraphics[width=\textwidth]{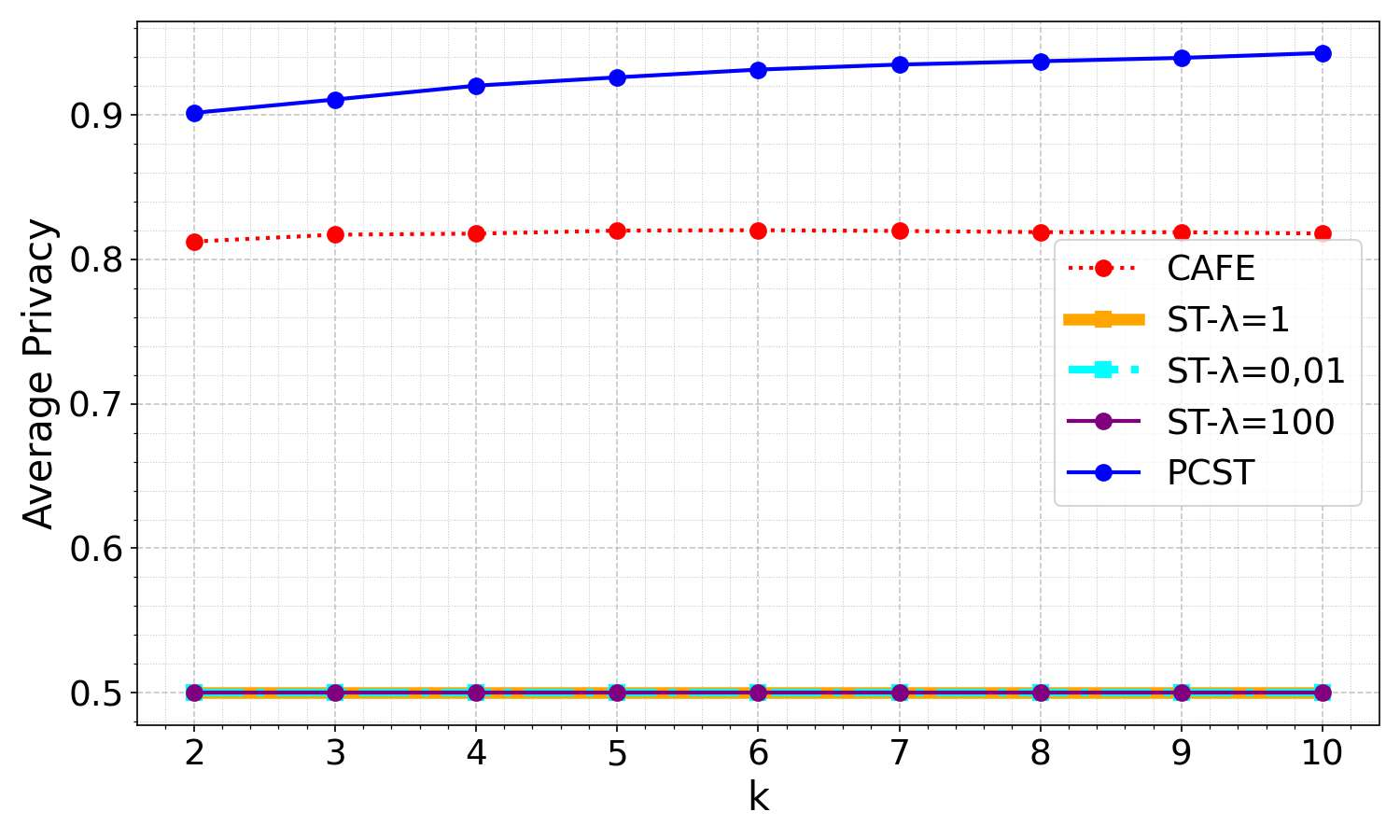}
        \caption{User-group CAFE}
        \label{fig:priv_cafe_user_group}
    \end{subfigure}
    \begin{subfigure}{0.24\textwidth} 
        \centering
        \includegraphics[width=\textwidth]{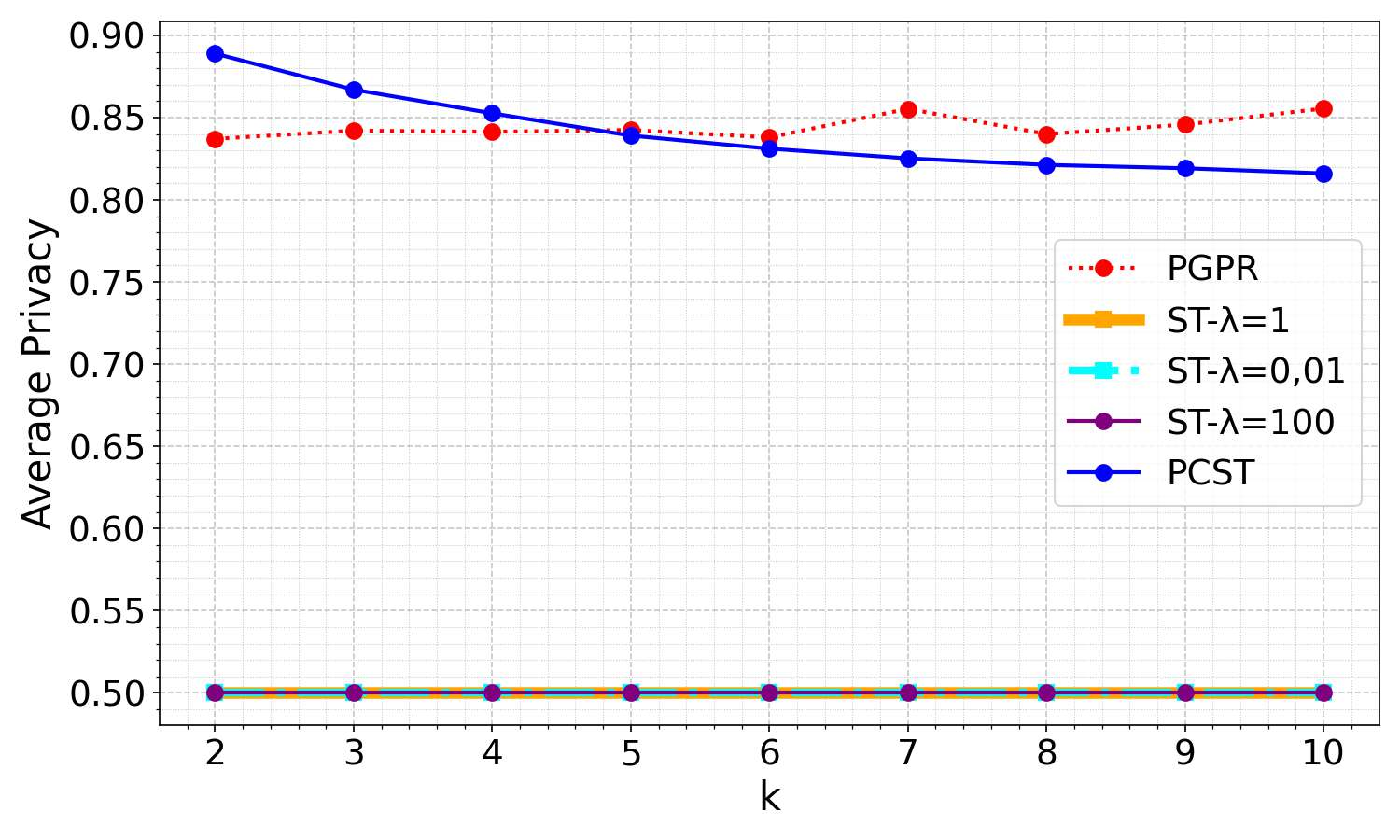}
        \caption{Item-group PGPR}
        \label{fig:priv_pgpr_item_group}
    \end{subfigure}
    \hfill
    \begin{subfigure}{0.24\textwidth}
        \centering
        \includegraphics[width=\textwidth]{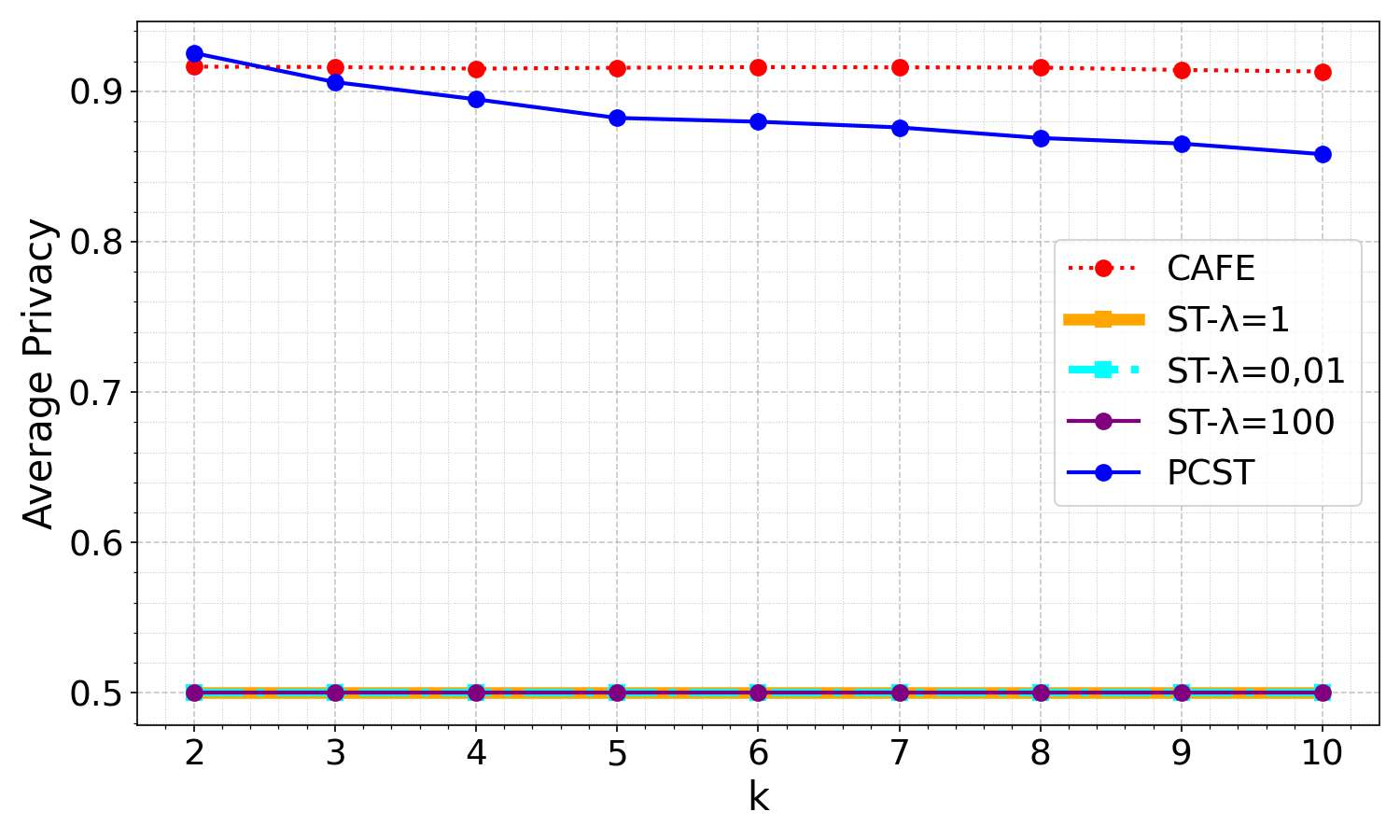}
        \caption{Item-group CAFE}
        \label{fig:priv_cafe_item_group}
    \end{subfigure}
    \caption{Privacy}
    \label{fig:priv}
\end{figure}

The privacy metric, $P(S)$, quantifies the exposure of user-specific information in explanation paths, with higher scores indicating better privacy protection (fewer user nodes relative to total nodes). In summaries, privacy measures the proportion of user nodes in the subgraph: $P(S) = 1 - \frac{\text{Number of user nodes in } S}{|V_S|}$.
As shown in Figure \ref{fig:priv}, PCST achieves the highest privacy scores across scenarios by favoring terminal nodes based on prizes, resulting in larger summaries that rely more on item nodes and external entities, limiting user node exposure. ST summaries, however, have lower privacy compared to the baselines because of the Steiner tree algorithms and the graph structure. In an ST summary, the goal is to connect terminal nodes (recommended items) through the paths with the highest edge weight within the graph. This leads to including user nodes in the summary since the only weighted edges are the user-item edges.

\subsubsection{Performance Metrics}
\begin{figure}[!t]
    \centering    
    \begin{subfigure}{0.24\textwidth} 
        \centering
        \includegraphics[width=\textwidth]{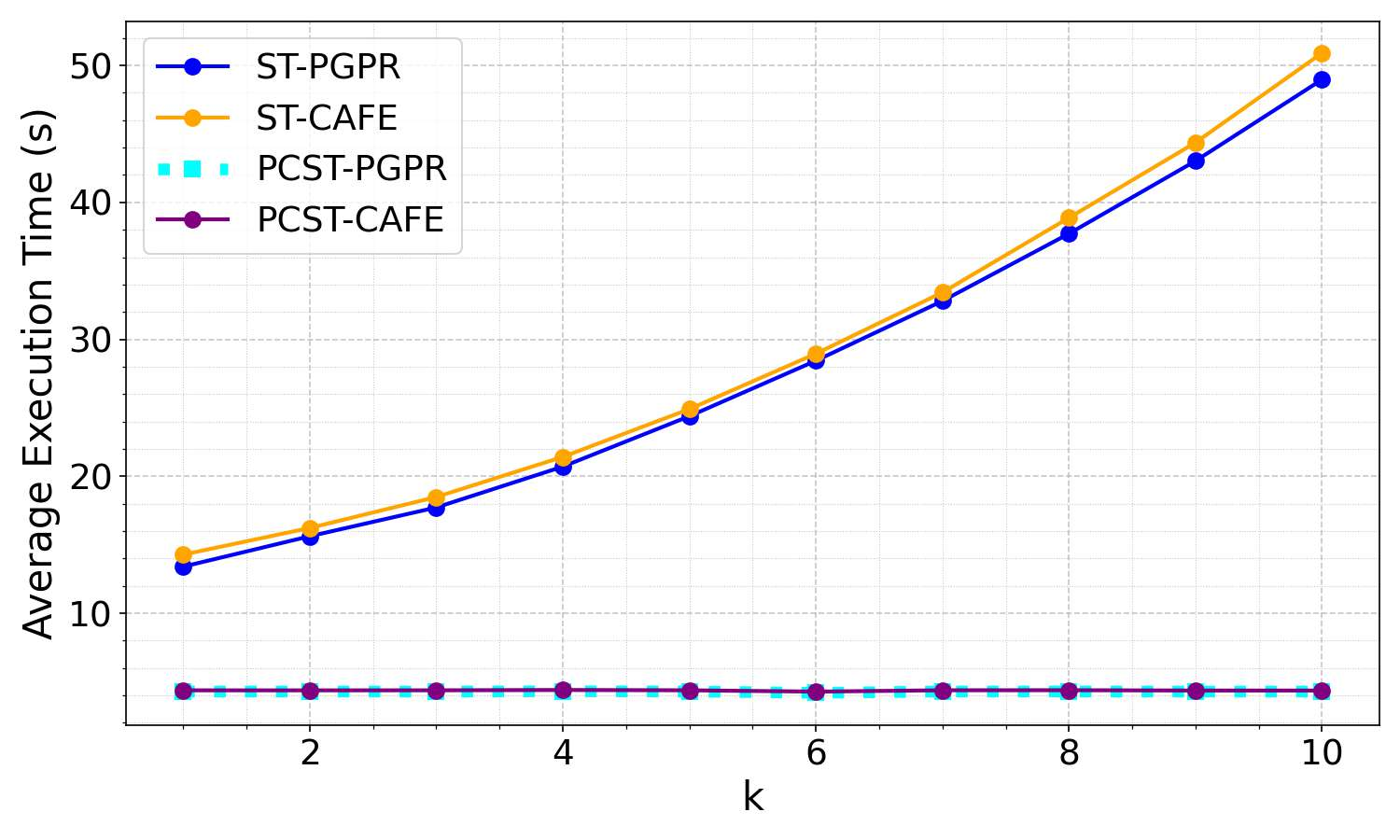}
        \caption{User-centric time}
        \label{fig:per_time_user}
    \end{subfigure}
    \hfill
    \begin{subfigure}{0.24\textwidth} 
        \centering
        \includegraphics[width=\textwidth]{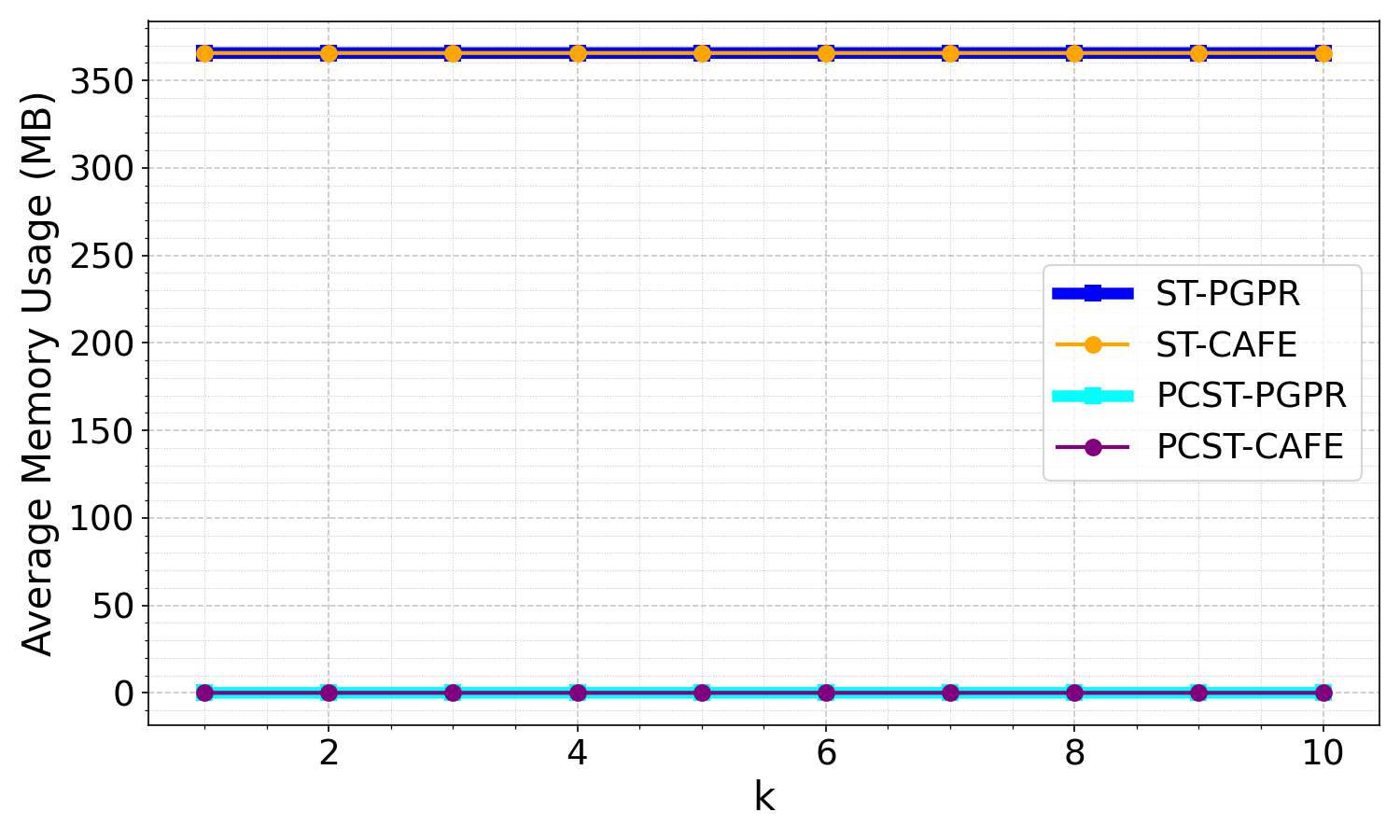}
        \caption{User-centric memory}
        \label{fig:per_mem_user}
    \end{subfigure}

    \begin{subfigure}{0.24\textwidth} 
        \centering
        \includegraphics[width=\textwidth]{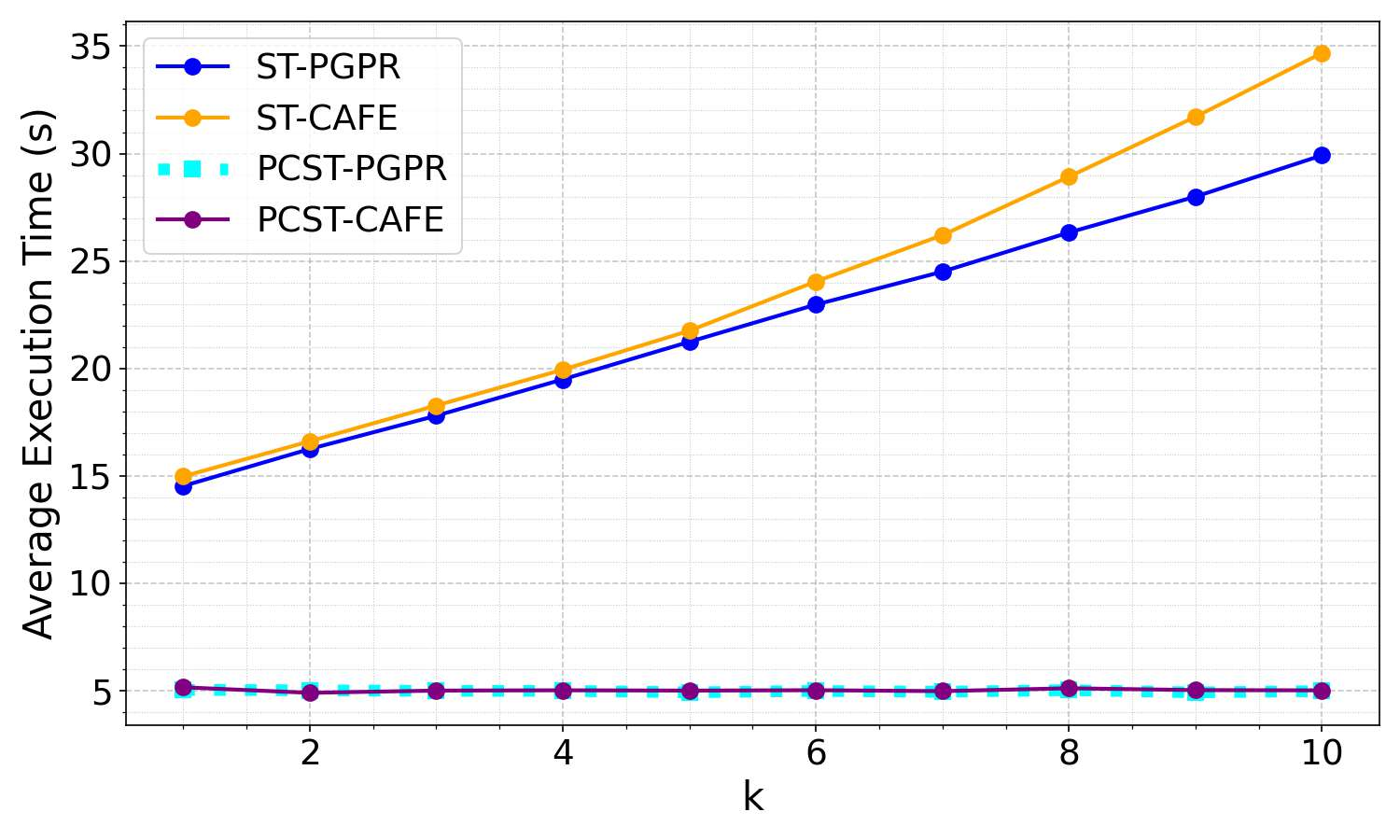}
        \caption{Item-centric time}
        \label{fig:per_time_item}
    \end{subfigure}
    \hfill
    \begin{subfigure}{0.24\textwidth} 
        \centering
        \includegraphics[width=\textwidth]{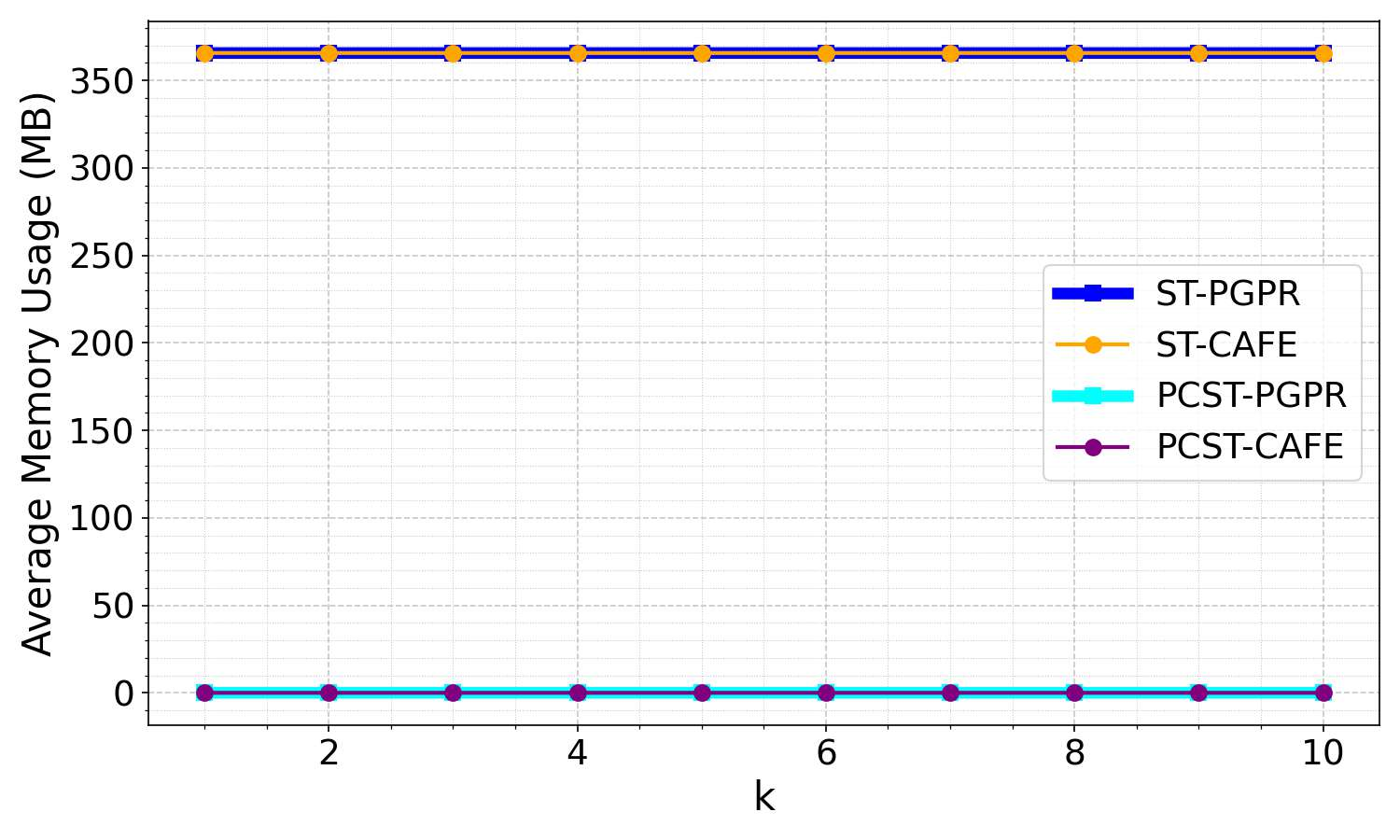}
        \caption{Item-centric memory}
        \label{fig:per_mem_item}
    \end{subfigure}

    \begin{subfigure}{0.24\textwidth} 
        \centering
        \includegraphics[width=\textwidth]{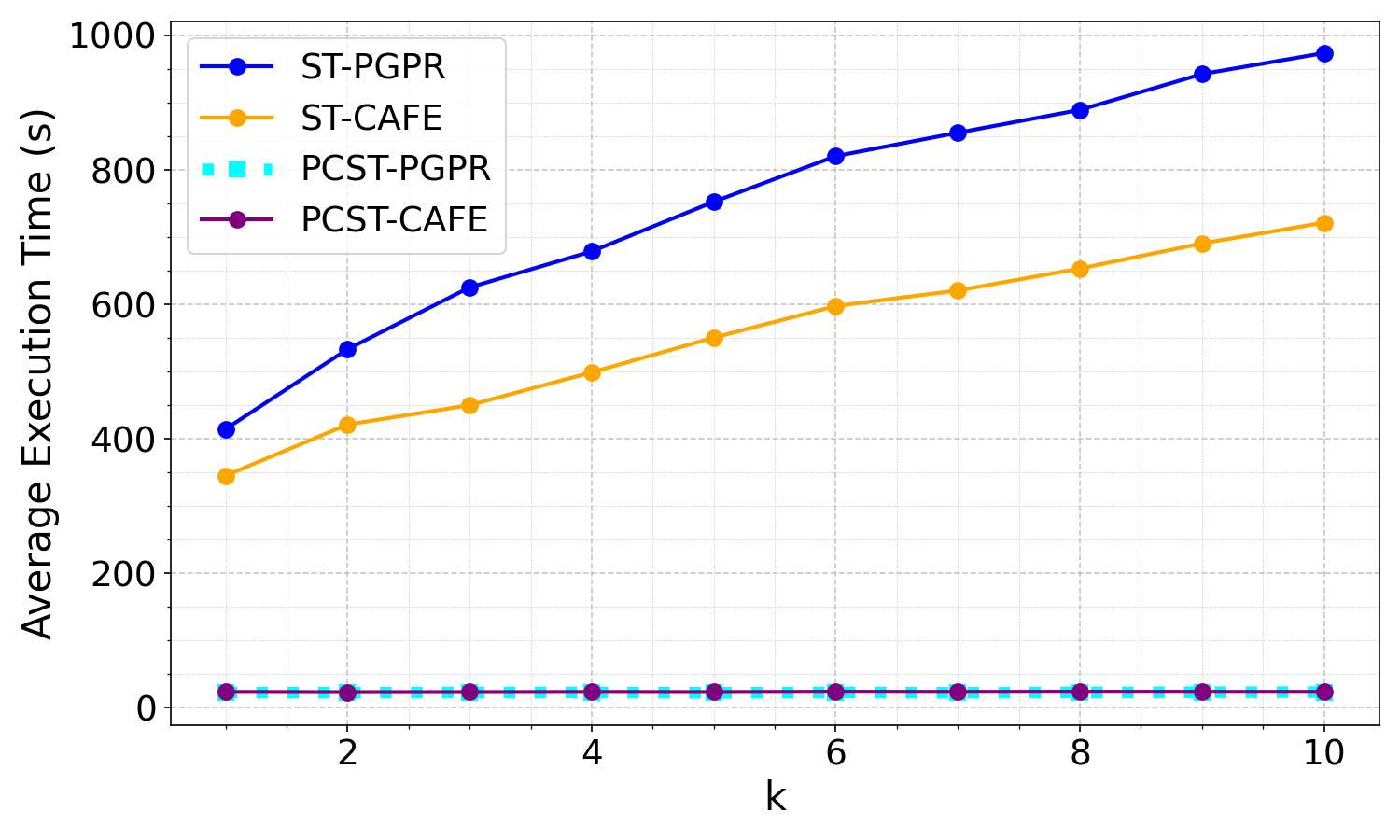}
        \caption{User-group time}
        \label{fig:per_time_user_group}
    \end{subfigure}
    \hfill
    \begin{subfigure}{0.24\textwidth} 
        \centering
        \includegraphics[width=\textwidth]{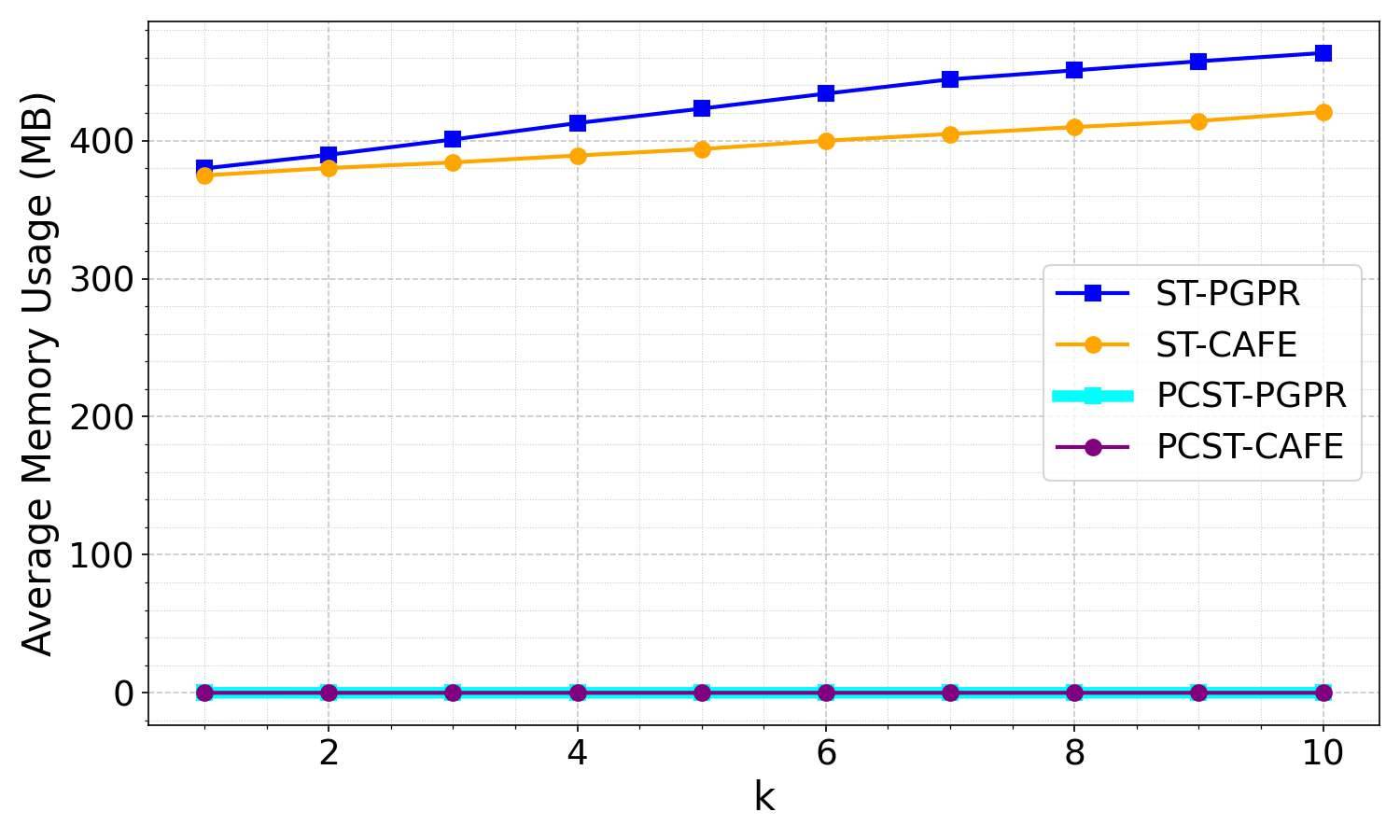}
        \caption{User-group memory}
        \label{fig:per_mem_user_group}
    \end{subfigure}


    \begin{subfigure}{0.24\textwidth} 
        \centering

        \includegraphics[width=\textwidth]{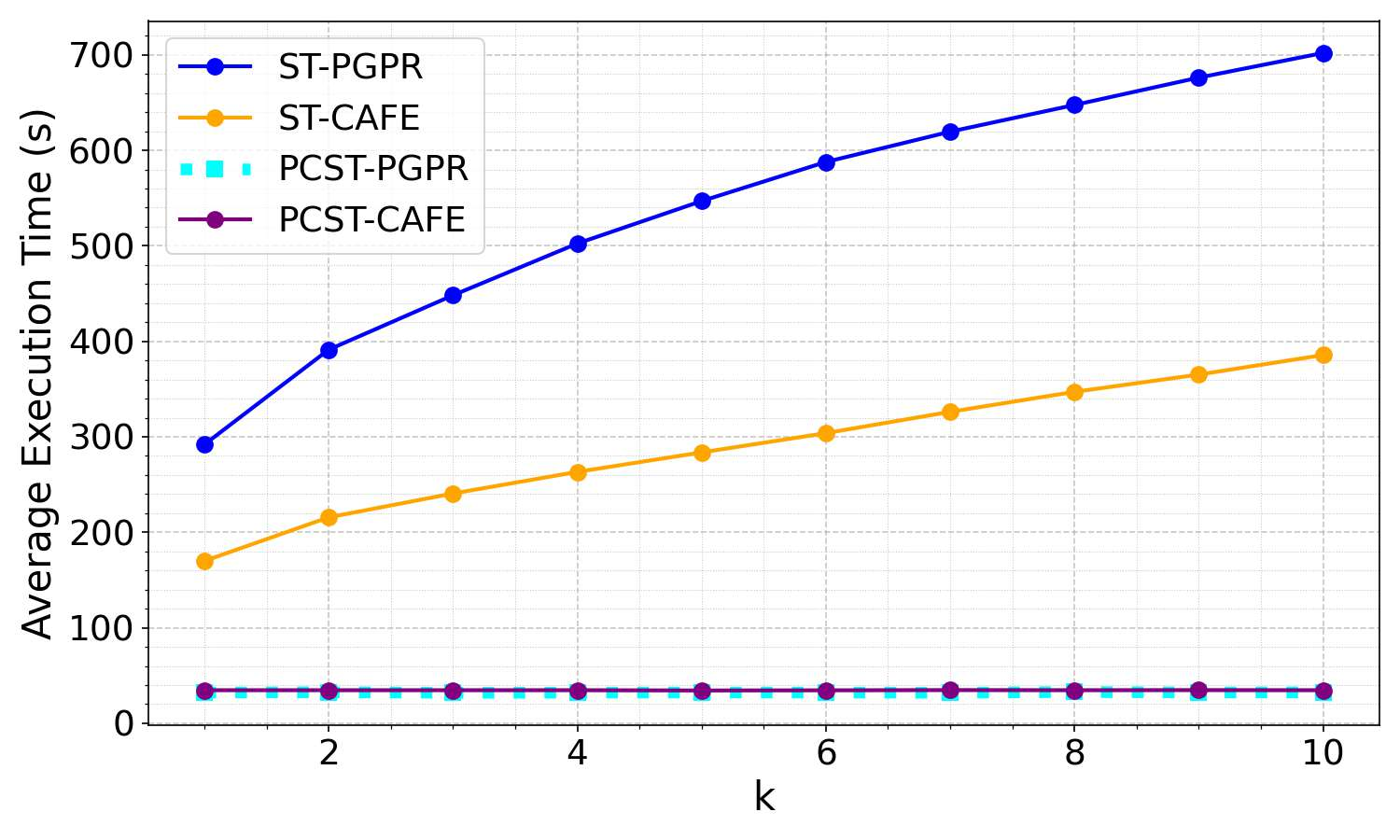}
        \caption{Item-group time}
        \label{fig:per_time_item_group}
    \end{subfigure}
    \hfill
    \begin{subfigure}{0.24\textwidth} 
        \centering
        \includegraphics[width=\textwidth]{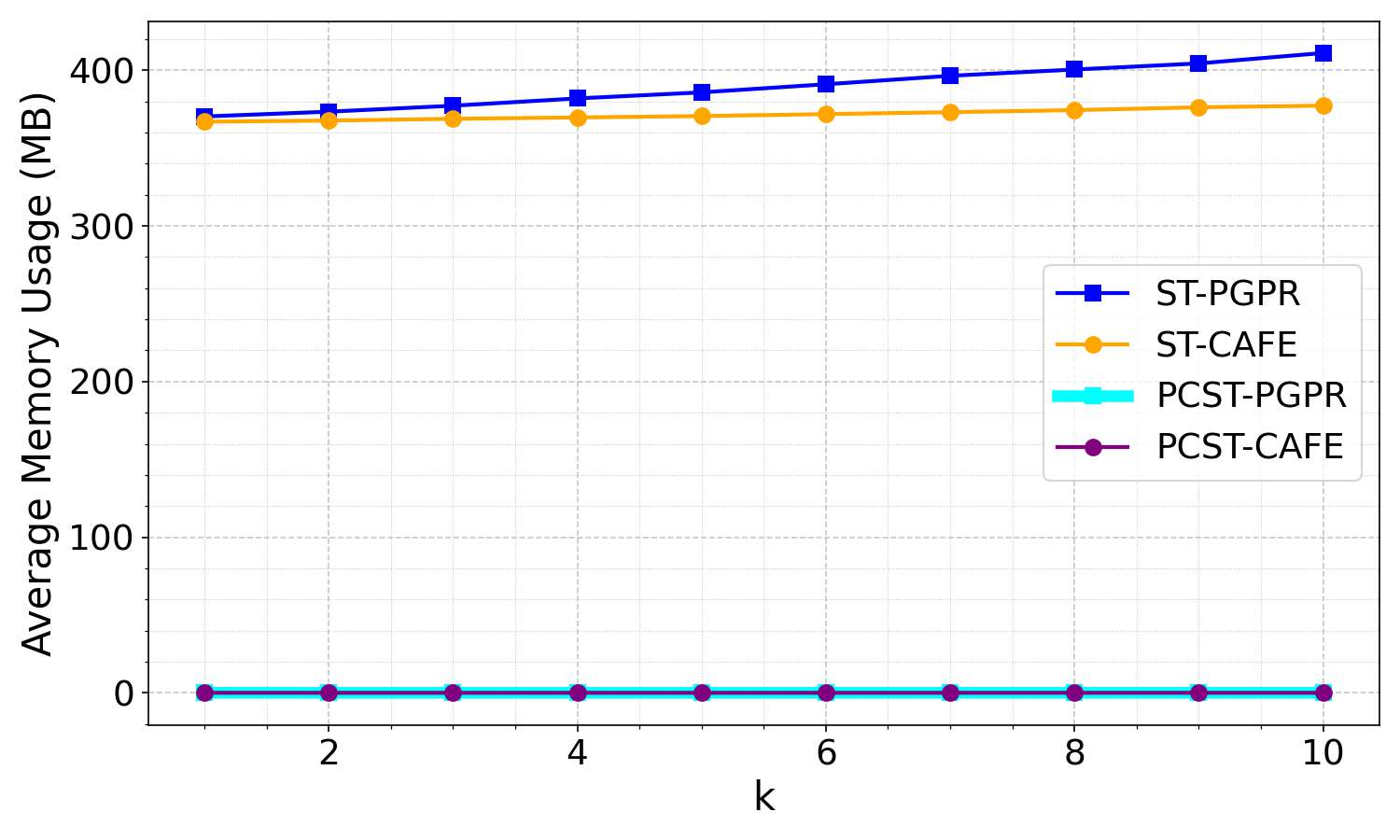}
        \caption{Item-group memory}
        \label{fig:per_mem_item_group}
    \end{subfigure}
    \caption{Performance}
    \label{fig:perf}
\end{figure}

\begin{figure}[ht]
    \centering
    \begin{subfigure}{0.24\textwidth}
        \centering
        \includegraphics[width=\textwidth]{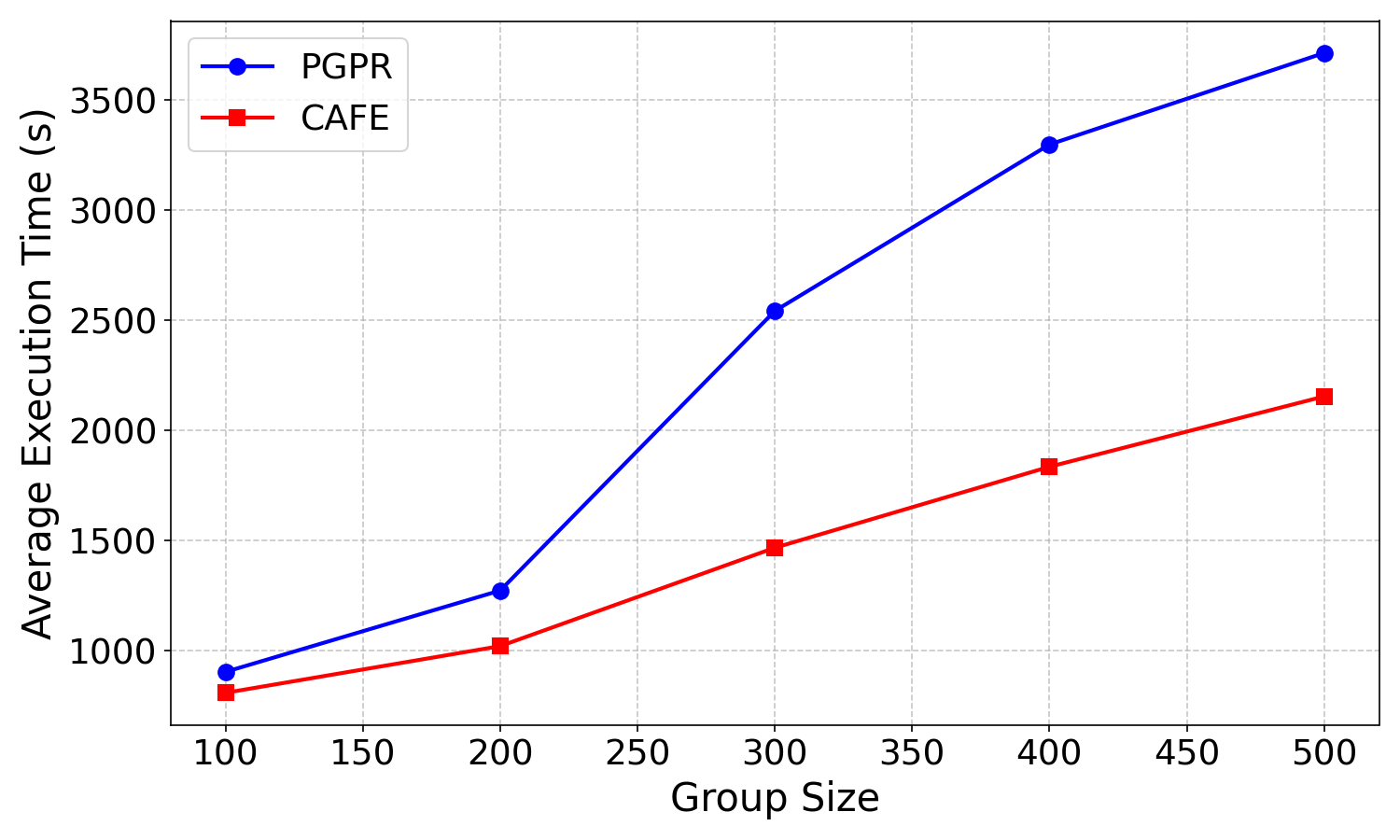}
        \caption{User-group ST}
    \end{subfigure}
    \hfill
    \begin{subfigure}{0.24\textwidth}
        \centering
        \includegraphics[width=\textwidth]{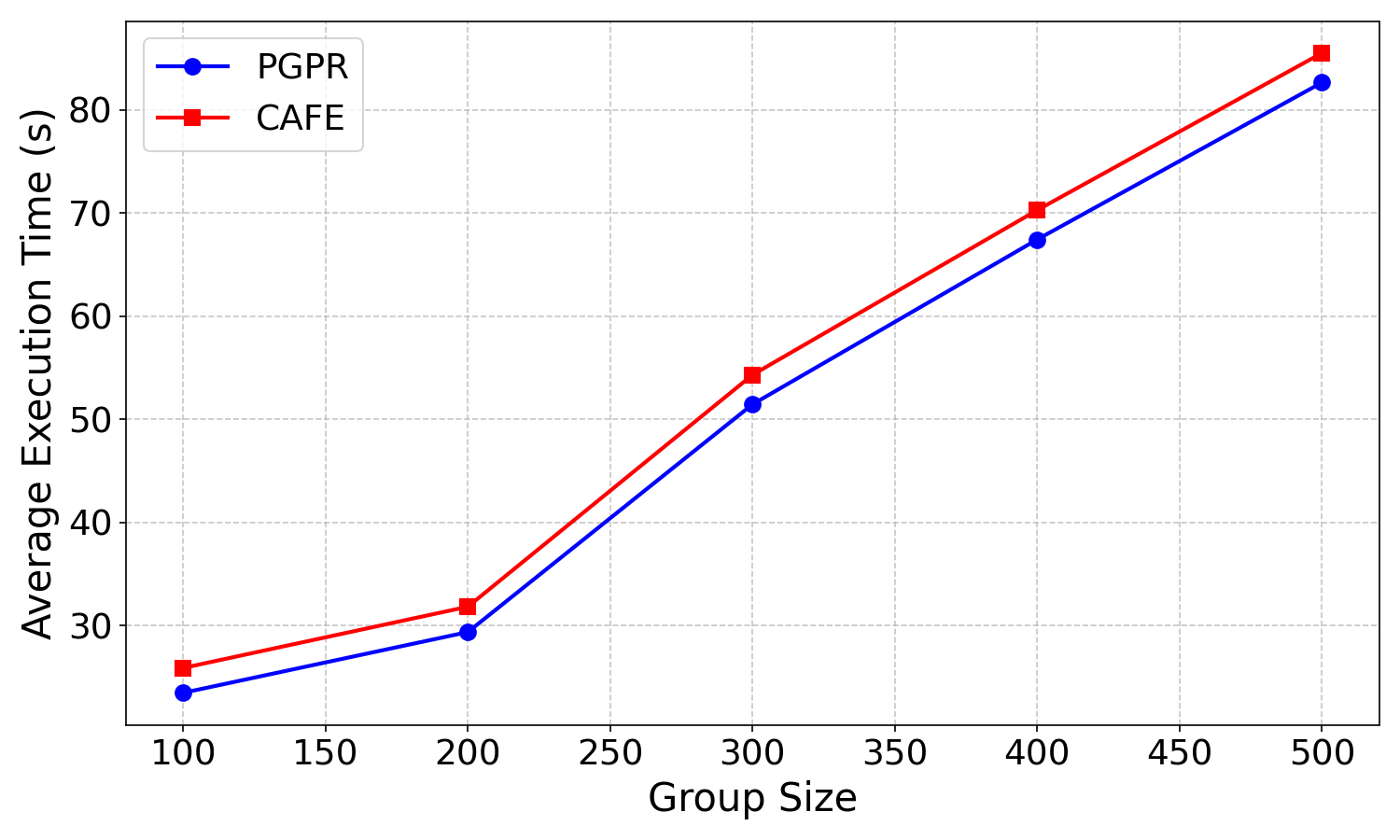}
        \caption{User-group PCST}
    \end{subfigure}
    \hfill
    \begin{subfigure}{0.24\textwidth}
        \centering
        \includegraphics[width=\textwidth]{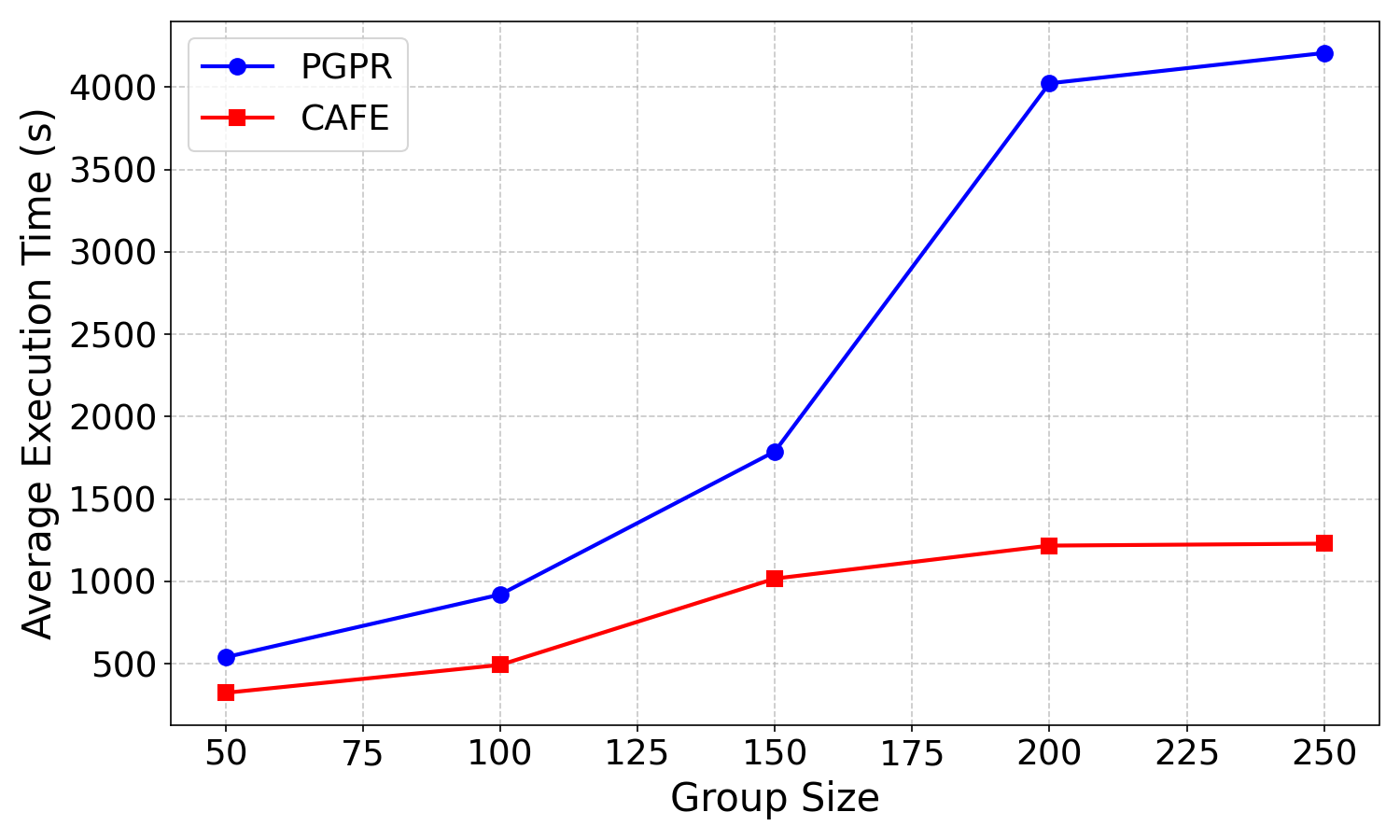}
        \caption{Item-group ST}
    \end{subfigure}
    \hfill
    \begin{subfigure}{0.24\textwidth}
        \centering
        \includegraphics[width=\textwidth]{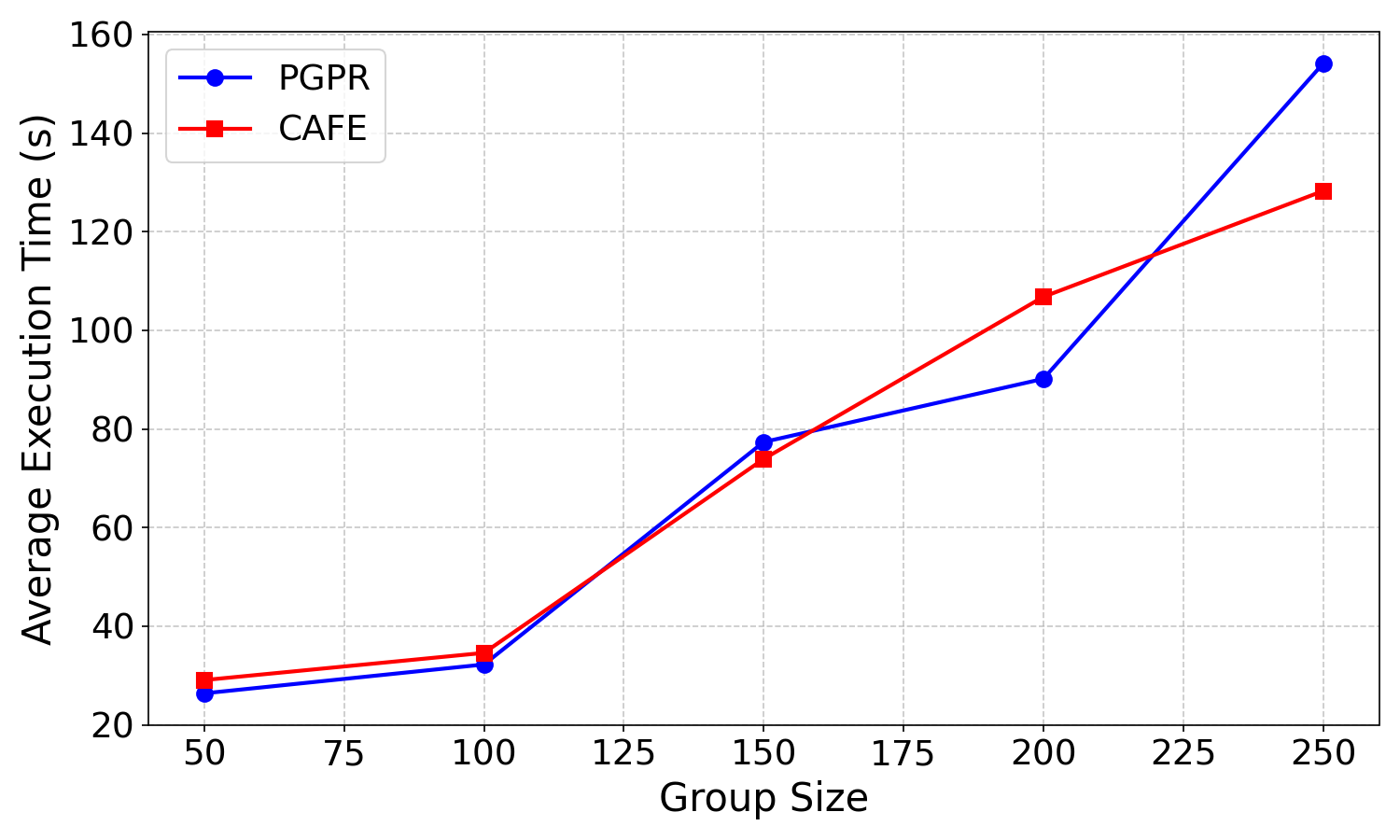}
        \caption{Item-group PCST}
    \end{subfigure}

    \caption{Performance for different group sizes}
    \label{fig:group_size}
\end{figure}

The performance metrics assess the computational efficiency of the summarization process, focusing on execution time and memory usage across different \( k \) values (number of recommended items or users). Figure \ref{fig:perf} shows that PCST performs significantly better, especially in large-scale scenarios, with the performance gap widening as \( k \) increases. Additionally, Figure \ref{fig:group_size} illustrates that ST's complexity, which depends on the number of terminal nodes \( |T| \), limits scalability as group size grows, causing execution times to increase rapidly. In contrast, PCST's independence from \( |T| \) allows it to scale efficiently, with gradual increases in execution time, remaining efficient even for larger groups.

\begin{table}[ht]
    \centering
    \caption{Synthetic Graph Statistics}
    \Large
    \resizebox{0.44\textwidth}{!}{%
    \begin{tabular}{|l|c|c|c|c|c|}
        \hline
        \textbf{Property} & \textbf{Graph 1} & \textbf{Graph 2} & \textbf{Graph 3} & \textbf{Graph 4} & \textbf{Graph 5} \\ \hline
        Number of users & 3,043 & 4,565 & 6,087 & 7,609 & 9,131 \\ \hline
        Number of items & 1,956 & 2,935 & 3,913 & 4,891 & 5,870 \\ \hline
        Number of external entities & 5,452 & 8,178 & 10,905 & 13,631 & 16,357 \\ \hline
        \textbf{Total number of nodes} & 10,000 & 15,000 & 20,000 & 25,000 & 30,000 \\ \hline
        \textbf{Total edges} & 559,734 & 839,601 & 1,119,468 & 1,399,335 & 1,679,202 \\ \hline
    \end{tabular}
    }
    \label{tab:graph_statistics_synthetic}
\end{table}

\begin{figure}[ht]
    \centering

    \begin{subfigure}{0.24\textwidth}
        \centering
        \includegraphics[width=\textwidth]{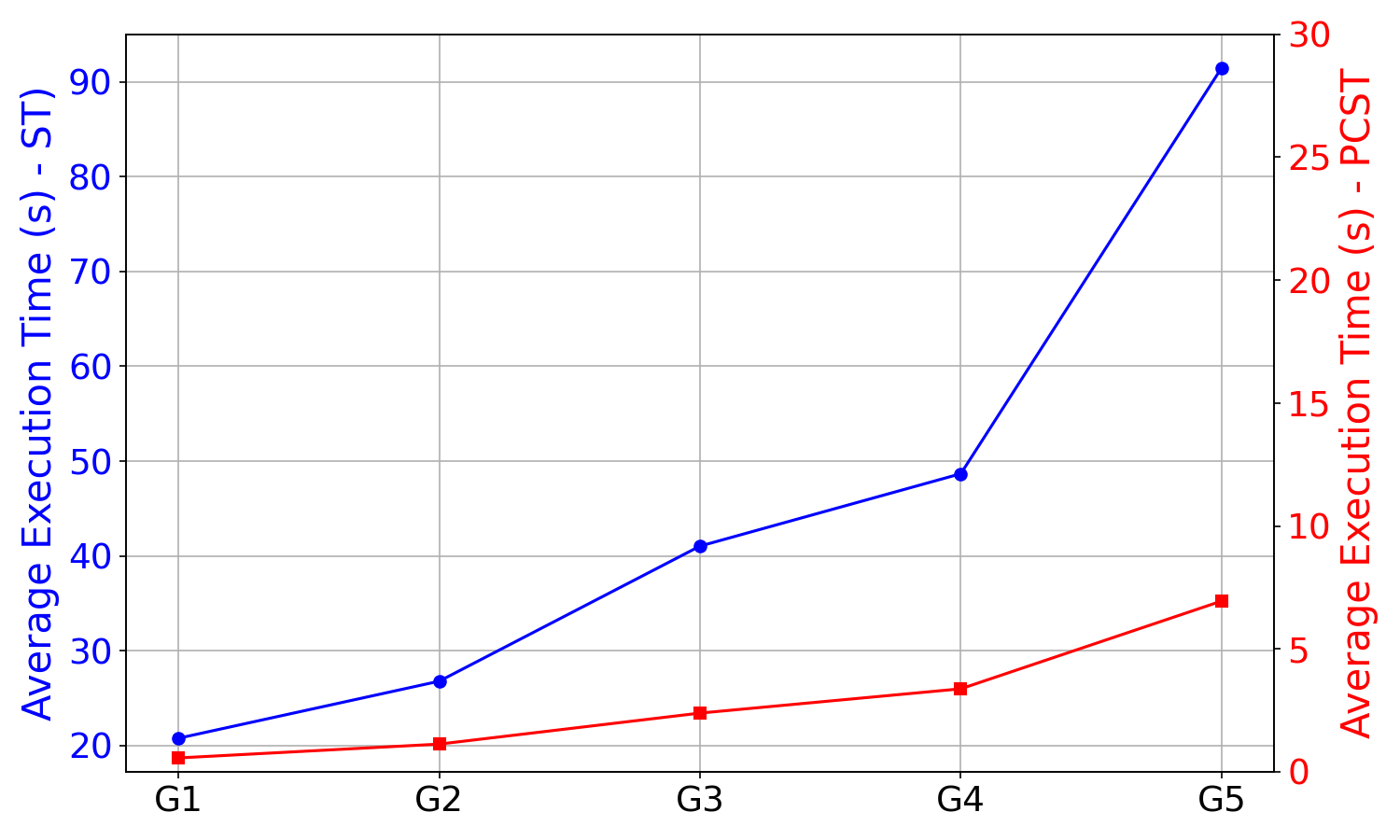}
        \caption{User-centric time}
    \end{subfigure}
    \hfill
    \begin{subfigure}{0.24\textwidth}
        \centering
        \includegraphics[width=\textwidth]{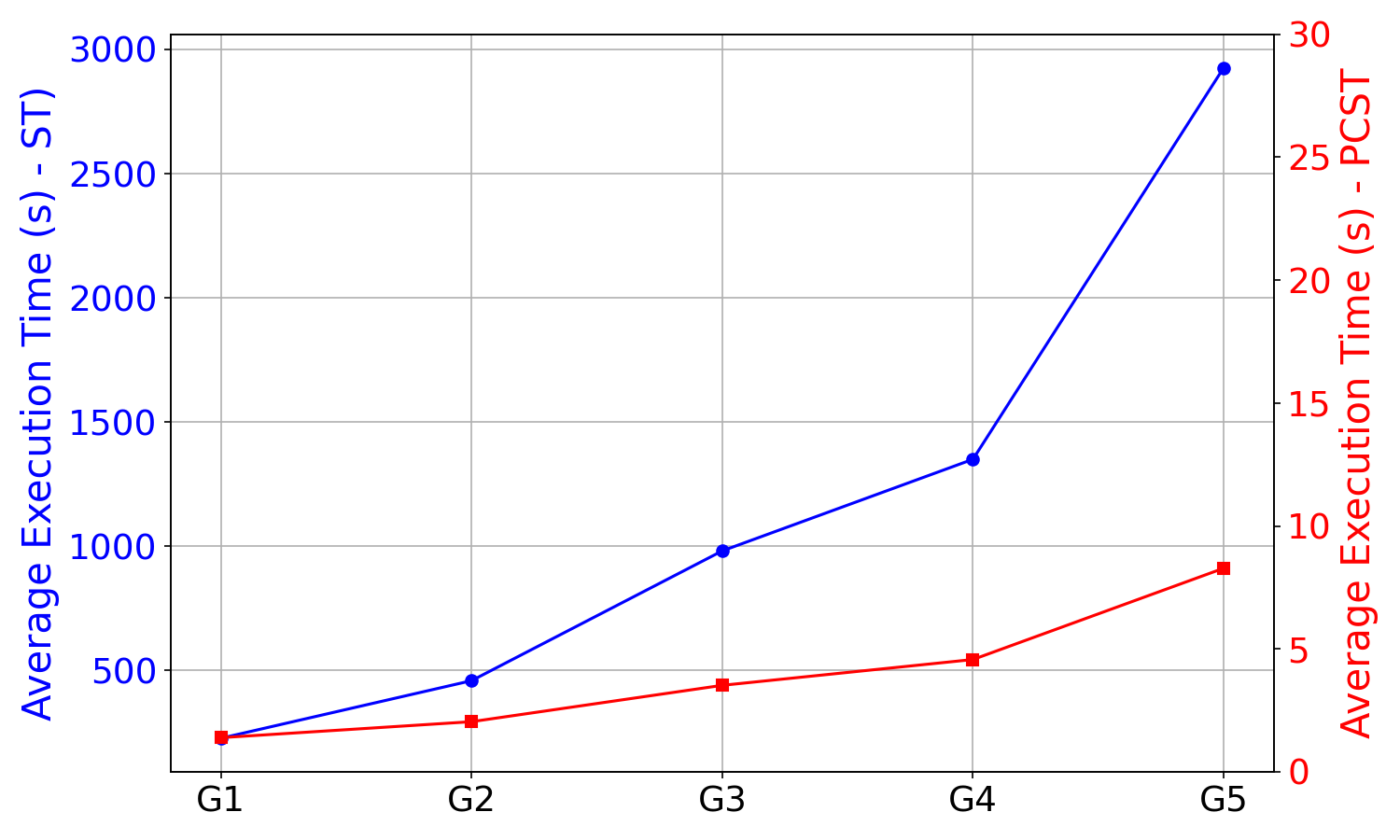}
        \caption{User-group time}
    \end{subfigure}

    \begin{subfigure}{0.24\textwidth}
        \centering
        \includegraphics[width=\textwidth]{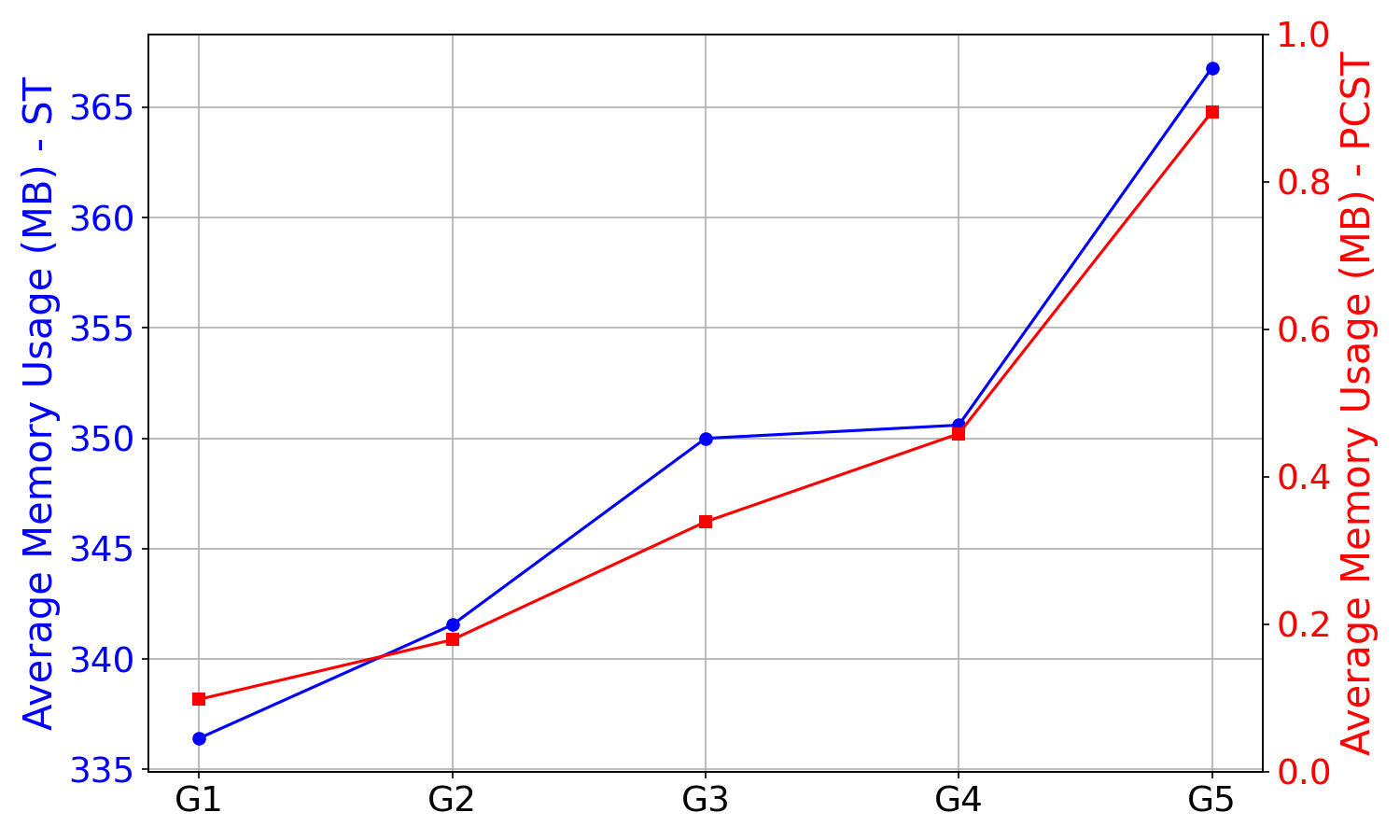}
        \caption{User-centric memory}
    \end{subfigure}
    \hfill
    \begin{subfigure}{0.24\textwidth}
        \centering
        \includegraphics[width=\textwidth]{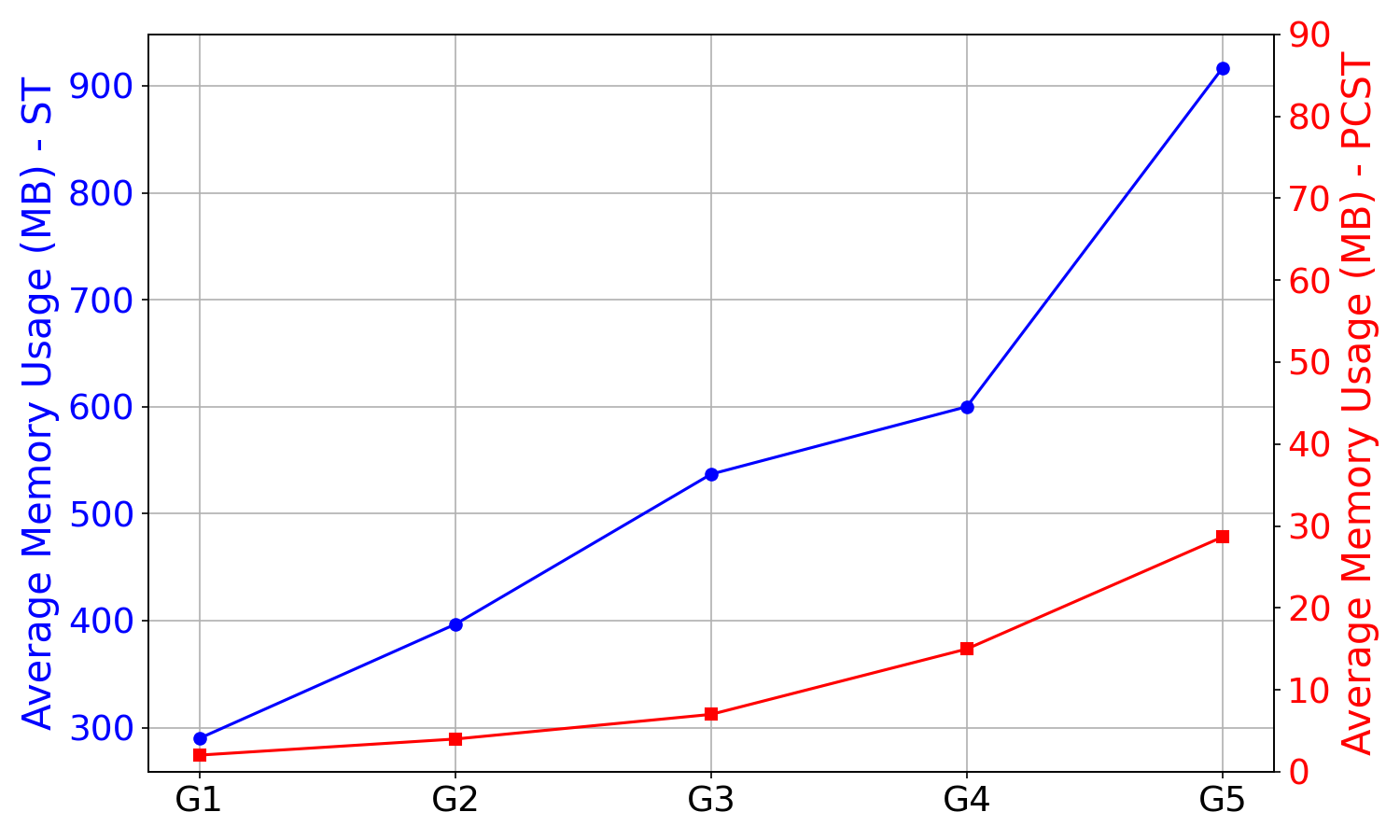}
        \caption{User-group memory}
    \end{subfigure}
    
    \caption{Performance for different graph sizes}
    \label{fig:synthetic}
\end{figure}

To further assess scalability, we generated synthetic random graphs with varying numbers of nodes and degrees for users, items, and external nodes set to be similar to the ML1M data. Graph characteristics are summarized in Table \ref{tab:graph_statistics_synthetic}. We report performance for \( k = 10 \) recommended items and two user groups of 100 users each. We test our algorithms on synthetic paths connecting users to items via random paths of length 3 as in the baselines. 
Figure \ref{fig:synthetic} shows that PCST scales more efficiently than ST, as expected. While both algorithms experience longer execution times as the graph size increases, PCST (both user-centric and user-group) demonstrates a slower rate of increase, particularly for larger graphs (G4 and G5). In contrast, ST execution time increases much more rapidly, especially in the user-group scenario. Results show that performance metrics are consistent with those observed in the ML1M graph.

\textit{\textbf{Comparing ST and PCST.}} In summary, ST performs better in comprehensibility and relevance, while PCST is superior in diversity, privacy, and scalability. Both methods reduce redundancy by limiting node duplication. Although PCST has lower comprehensibility than ST, it is useful for group explanations due to its scalability for large groups, as shown in Figures \ref{fig:perf} and \ref{fig:group_size}.

\begin{figure}[ht]
    \centering
    \begin{subfigure}{0.24\textwidth}
        \centering
        \includegraphics[width=\textwidth]{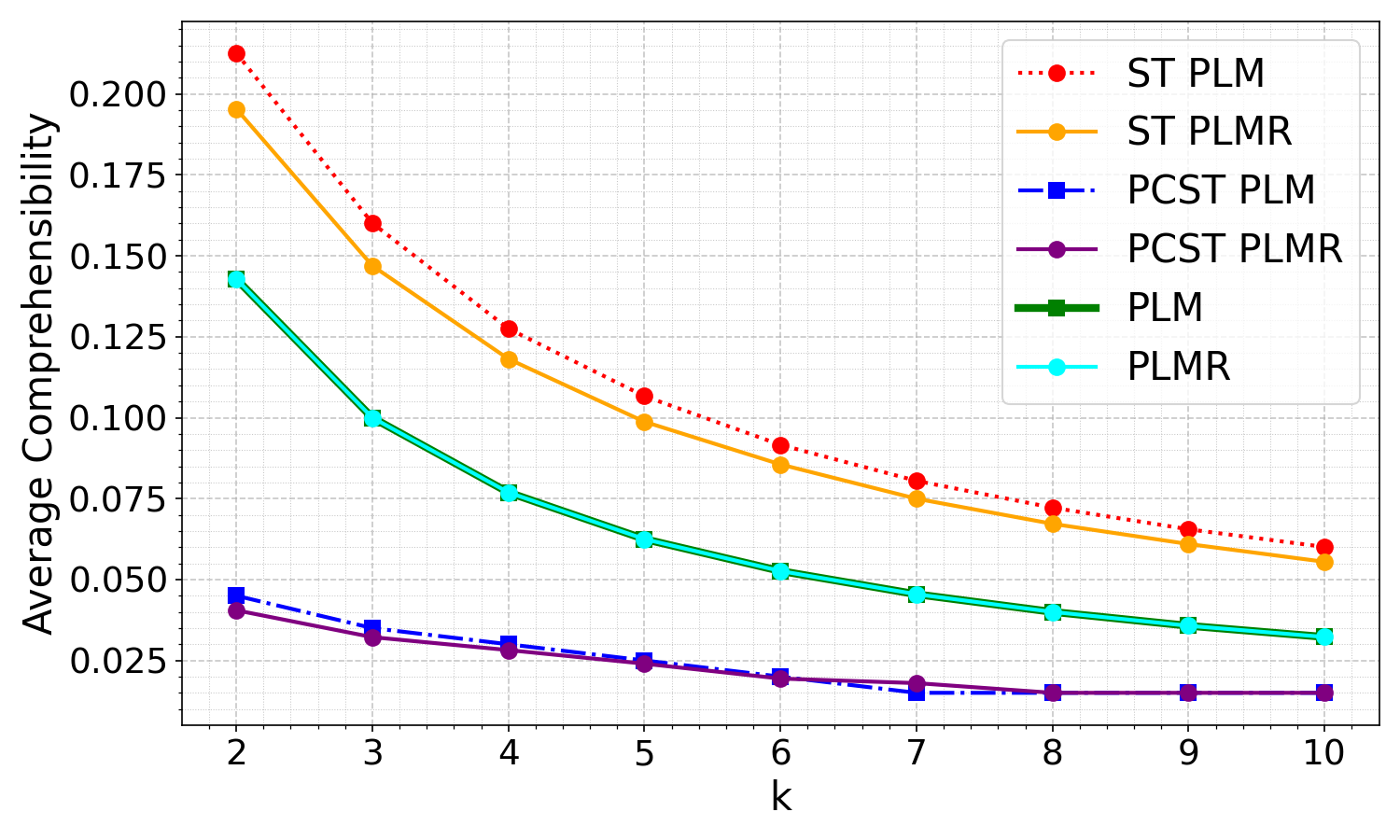}
        \caption{User-centric}
    \end{subfigure}
    \hfill
    \begin{subfigure}{0.24\textwidth}
        \centering
        \includegraphics[width=\textwidth]{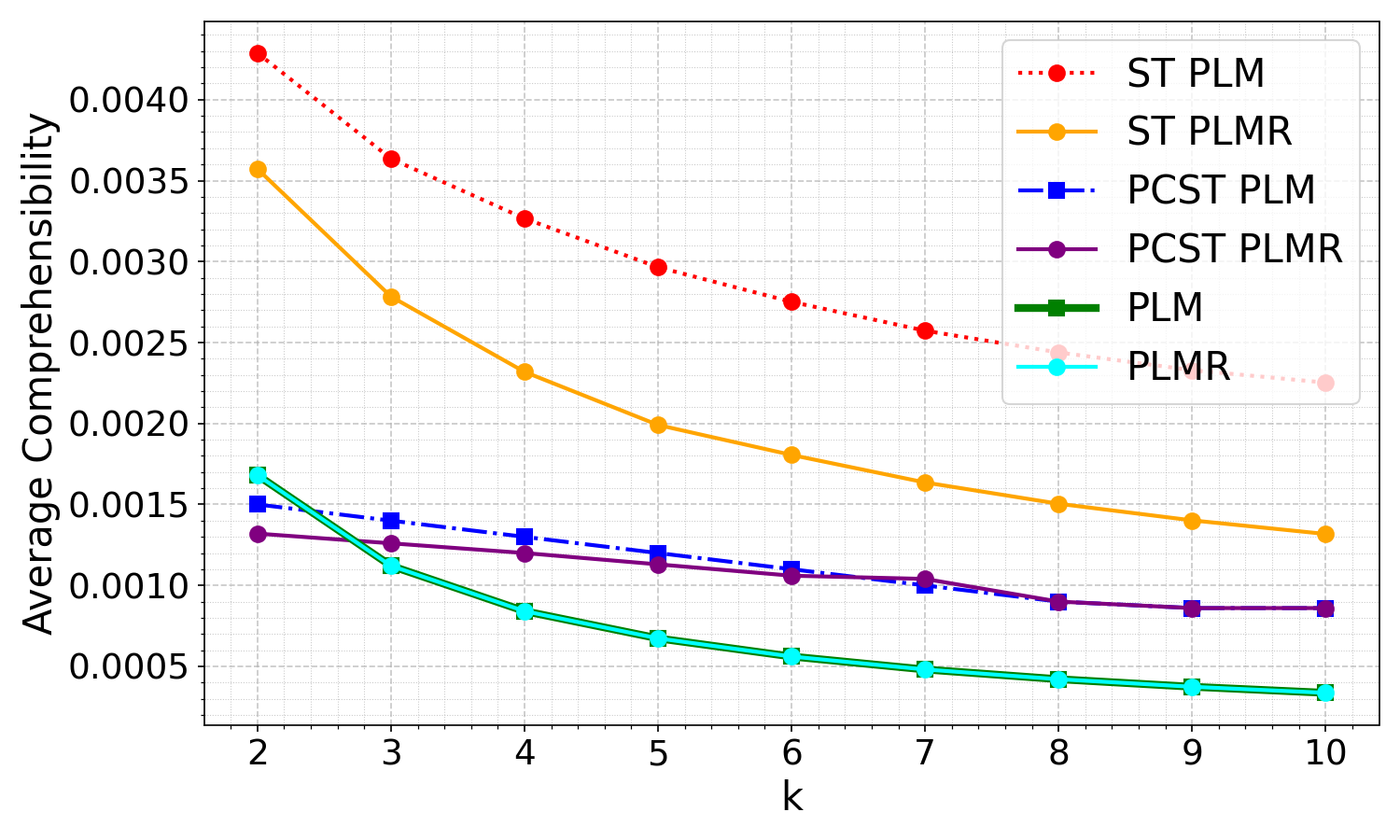}
        \caption{User-group}
    \end{subfigure}
    \caption{Comprehensibility for PLM and PLMR baselines}
    \label{fig:perf_cmpr_more}
\end{figure}

\begin{figure}[ht]
    \centering

    \begin{subfigure}{0.24\textwidth}
        \centering
        \includegraphics[width=\textwidth]{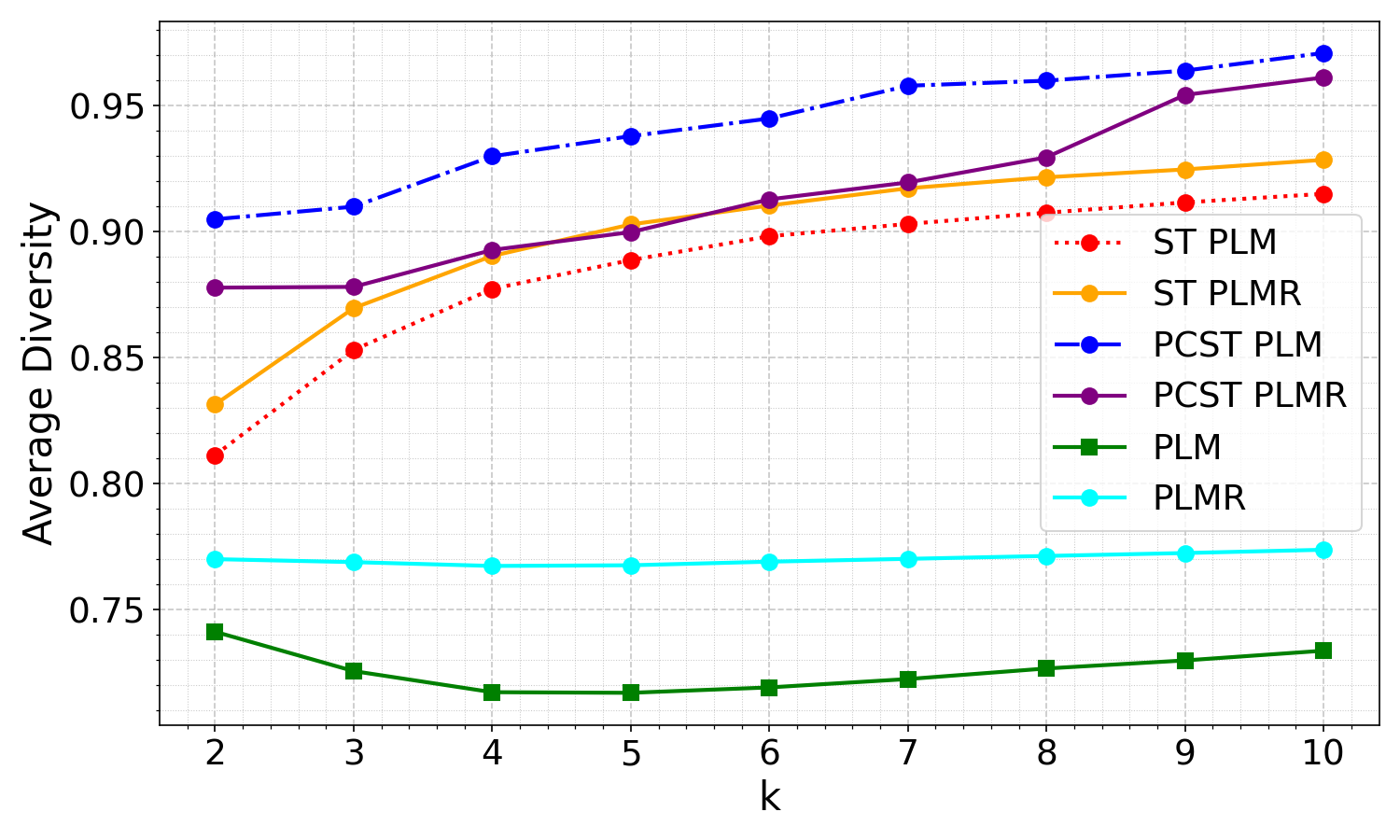}
        \caption{User-centric}
    \end{subfigure}
    \hfill
    \begin{subfigure}{0.24\textwidth}
        \centering
        \includegraphics[width=\textwidth]{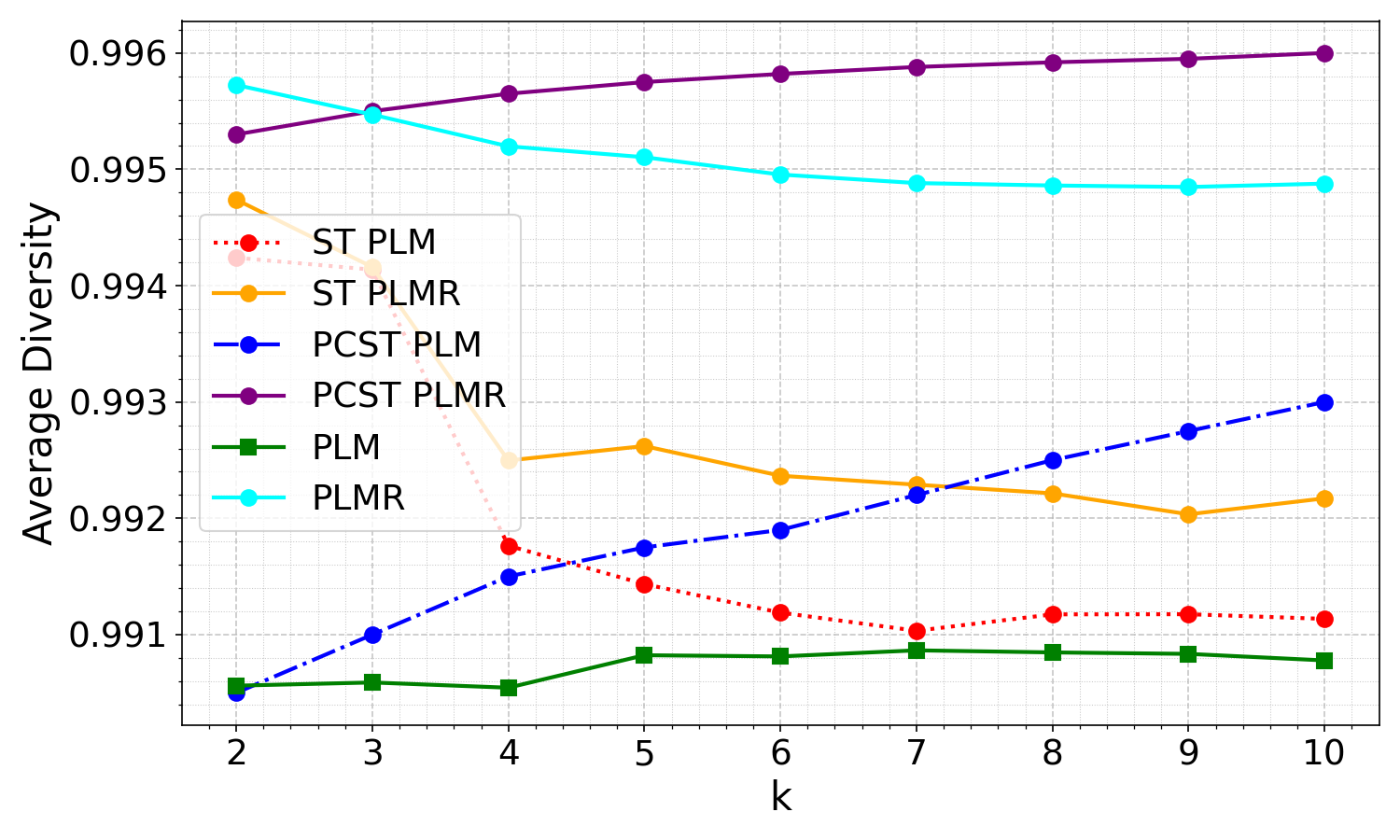}
        \caption{User-group}
    \end{subfigure}
    \caption{Diversity for PLM and PLMR baselines}
    \label{fig:perf_div_morebaselines}
\end{figure}

\textit{\textbf{Additional Baselines.}}
\label{baselines}
We extend our evaluation with two additional baselines: PLM \cite{plm} and PLMR \cite{balloccu2023faithful}, state-of-the-art language model-based methods for path-based recommendation explanations. We focus on comprehensibility and diversity, the top metrics from our user study (Section \ref{section:eval}). Experiments on the ML1M dataset for \( k = 1 \text{ to } 10 \) allow direct comparison with previous baselines. Results in Figure \ref{fig:perf_cmpr_more} are consistent with the findings for the PGPR and CAFE baselines in Figure \ref{fig:cmp}, and show that ST improves the comprehensibility of PLM and PLMR, while PCST shows slightly better comprehensibility at higher \( k \)-values in user-group scenarios. Figure \ref{fig:perf_div_morebaselines} shows that, though PLM and PLMR exhibit higher diversity based on their inherent path generation methodology compared to the PGPR and CAFE (Figure \ref{fig:div}), PCST further enhances diversity, with ST offering moderate diversity, particularly in the user-group scenario. These results align with findings from the PGPR and CAFE baselines.


\begin{figure}[!t]
    \centering

    \begin{subfigure}{0.24\textwidth}
        \centering
        \includegraphics[width=\textwidth]{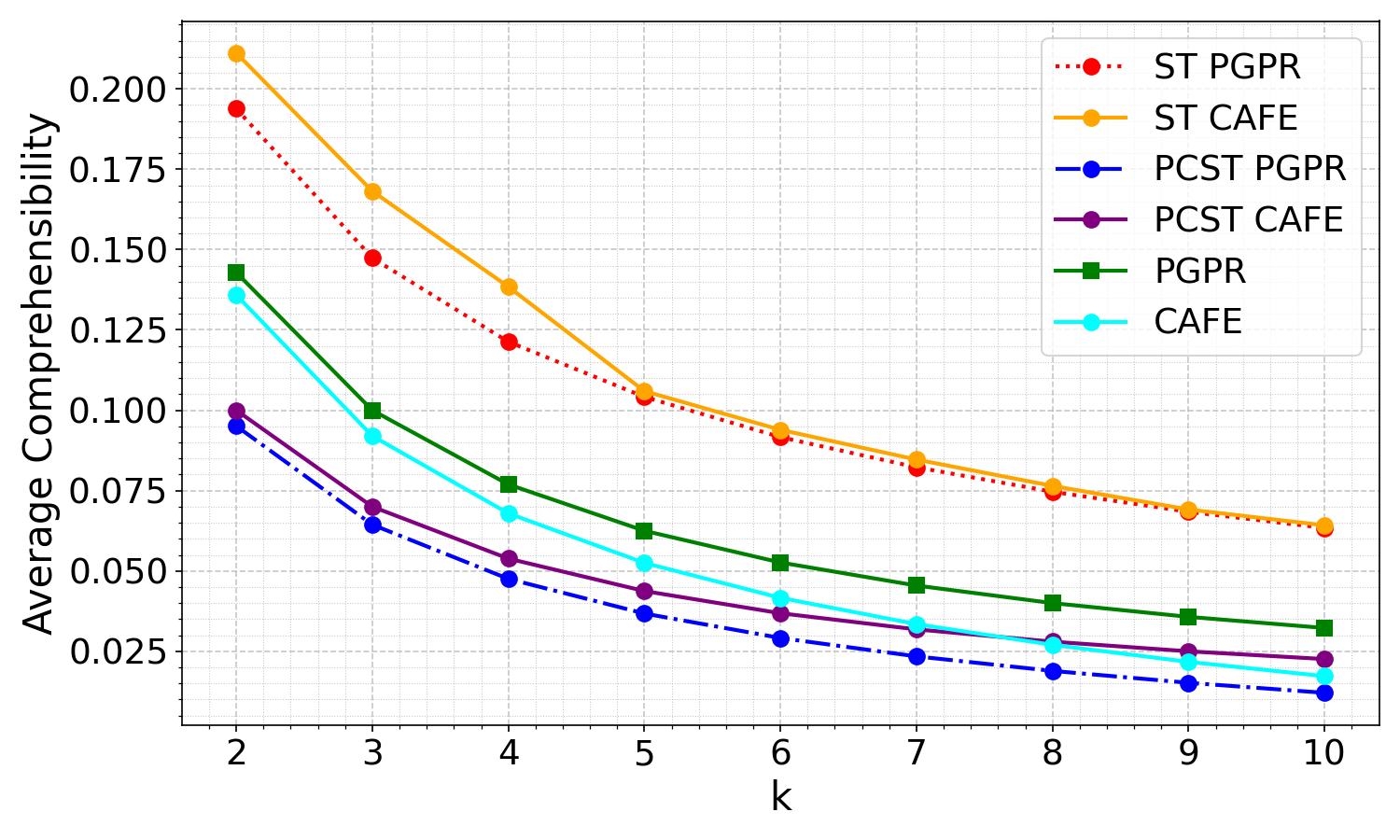}
        \caption{User-centric}
    \end{subfigure}
    \hfill
    \begin{subfigure}{0.24\textwidth}
        \centering
        \includegraphics[width=\textwidth]{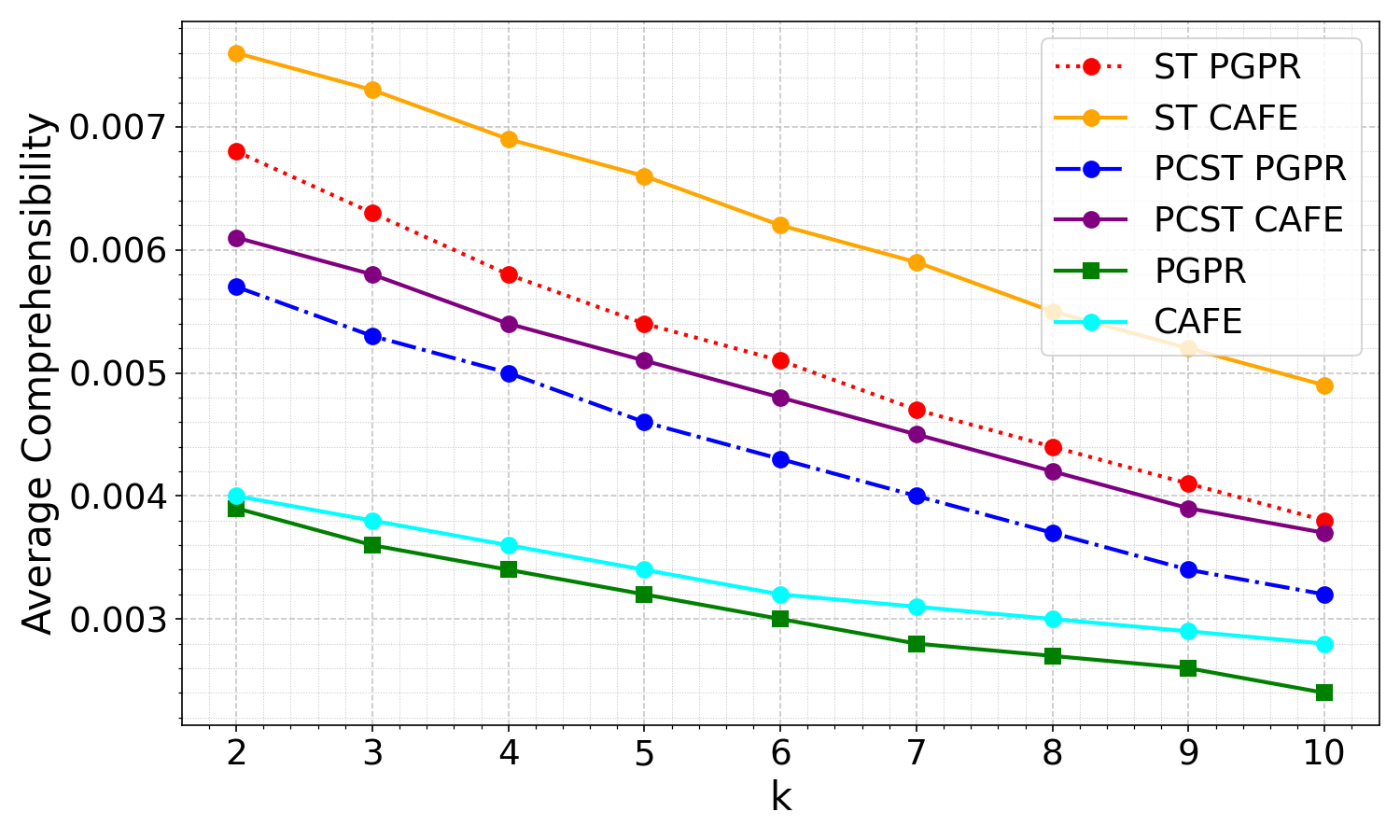}
        \caption{User-group}
    \end{subfigure}
    \caption{Comprehensibility for LFM1M dataset}
    \label{fig:perf_4}
\end{figure}

\begin{figure}[!t]
    \centering

    \begin{subfigure}{0.24\textwidth}
        \centering
        \includegraphics[width=\textwidth]{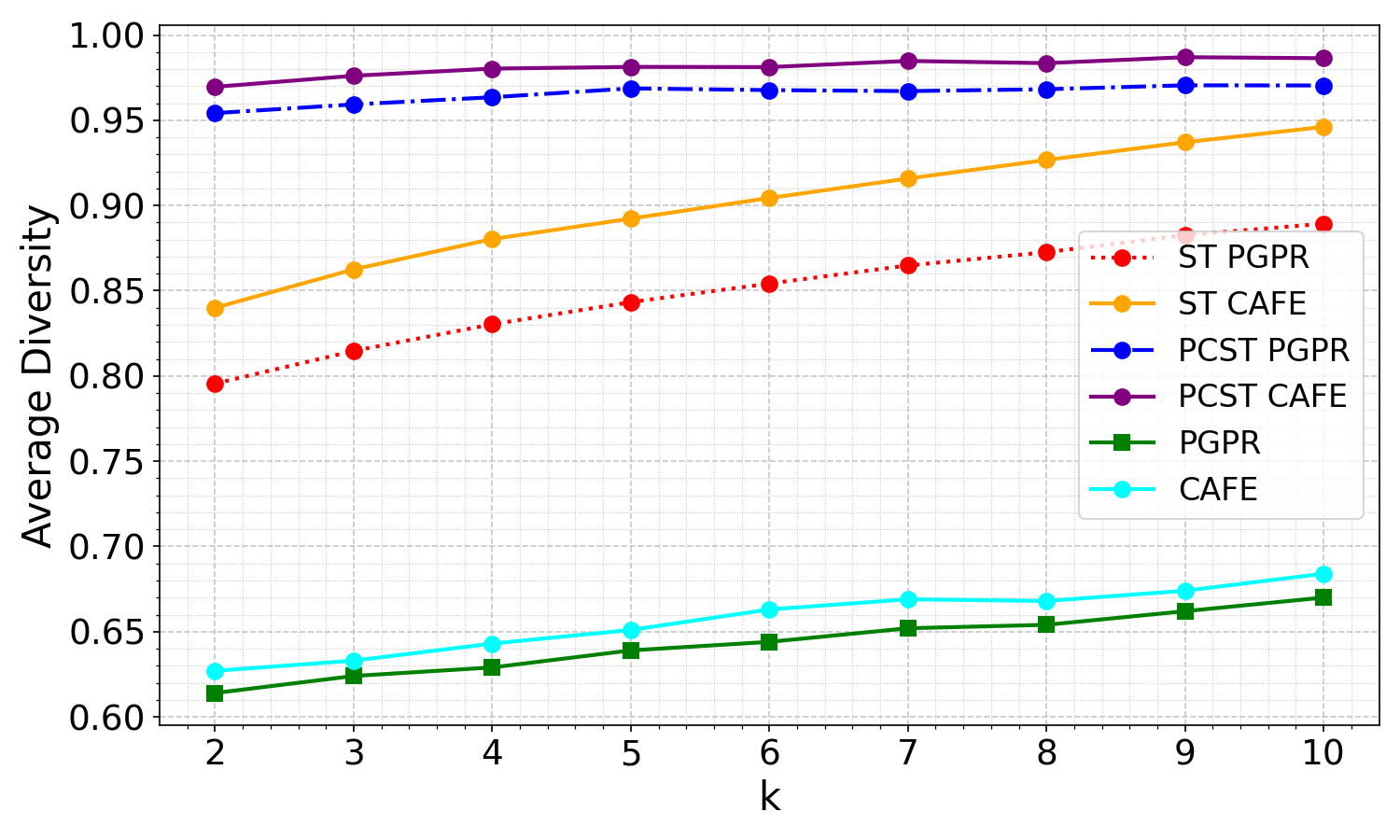}
        \caption{User-centric}
    \end{subfigure}
    \hfill
    \begin{subfigure}{0.24\textwidth}
        \centering
        \includegraphics[width=\textwidth]{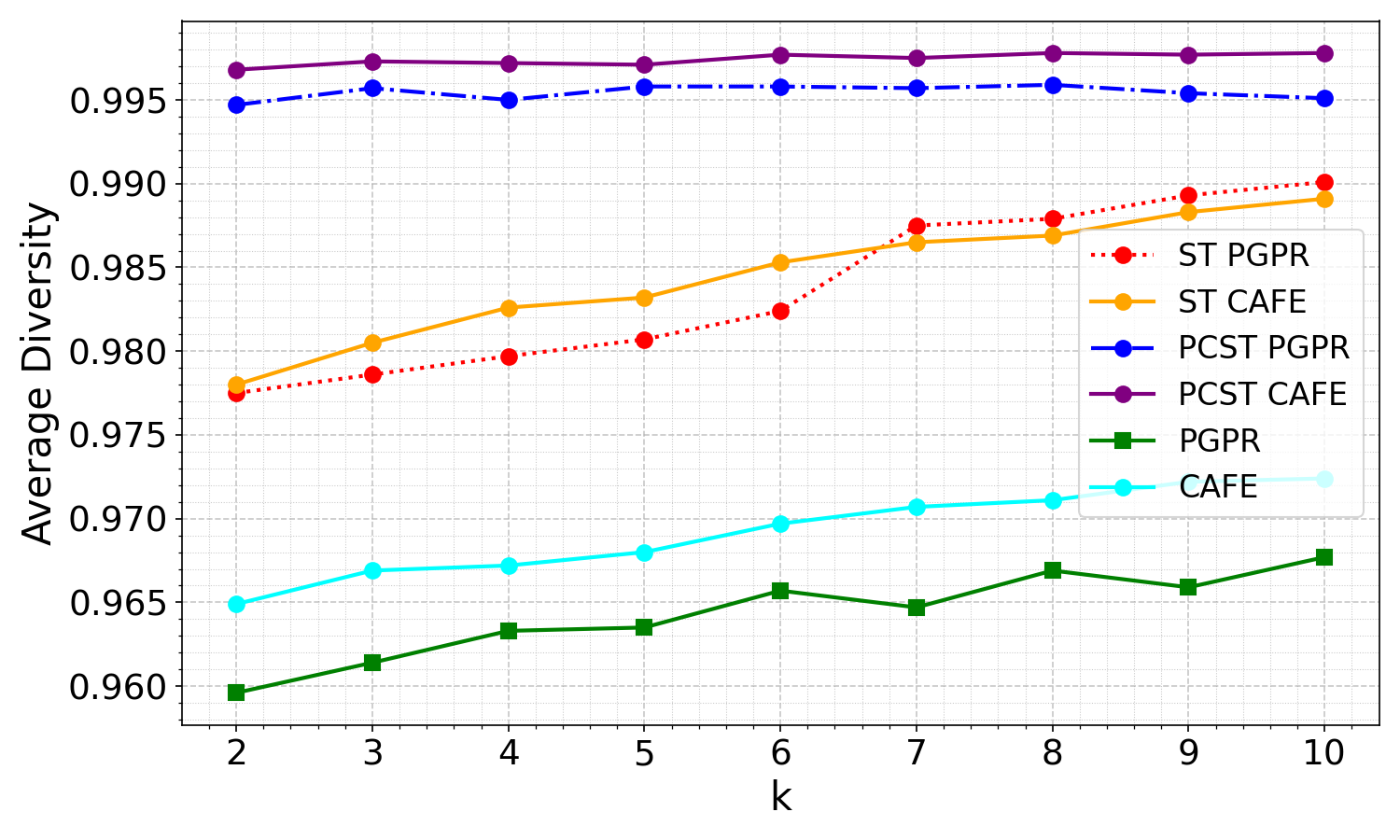}
        \caption{User-group}
    \end{subfigure}
    \caption{Diversity for LFM1M dataset}
    \label{fig:perf_lmfm}
\end{figure}

\textit{\textbf{Additional Dataset.}}
\label{lfm1m}
To further validate our approach, we ran experiments on the LastFM-1M (LFM1M) dataset, a subset of LastFM-1B \cite{lfm1b}, containing 1,091,274 user-song interactions across 4,817 users, 12,492 tracks, and 17,491 external entities. A new knowledge graph was created similar to ML1M, with recommendation paths generated using PGPR and CAFE for \( k = 1 \text{ to } 10 \). We evaluated comprehensibility and diversity, the top-rated metrics from our user study. Results in Figures \ref{fig:perf_4} and \ref{fig:perf_lmfm} align with findings from ML1M (Figures \ref{fig:cmp} and \ref{fig:div}).


\begin{figure}[ht]
    \centering
    \begin{subfigure}{0.24\textwidth}
        \centering
        \includegraphics[width=\textwidth]{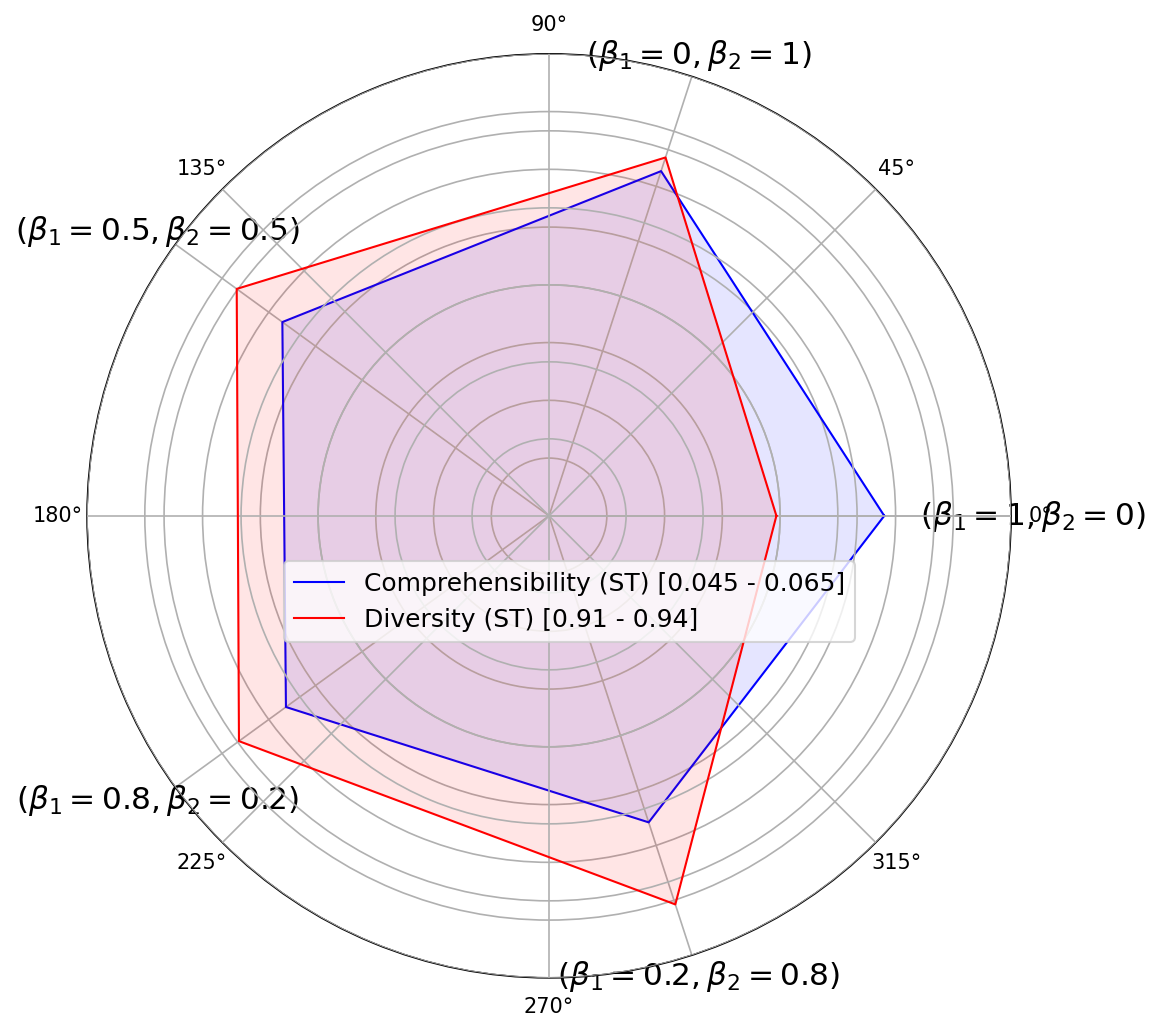}
        \caption{User-centric}
    \end{subfigure}
    \hfill
    \begin{subfigure}{0.24\textwidth}
        \centering
        \includegraphics[width=\textwidth]{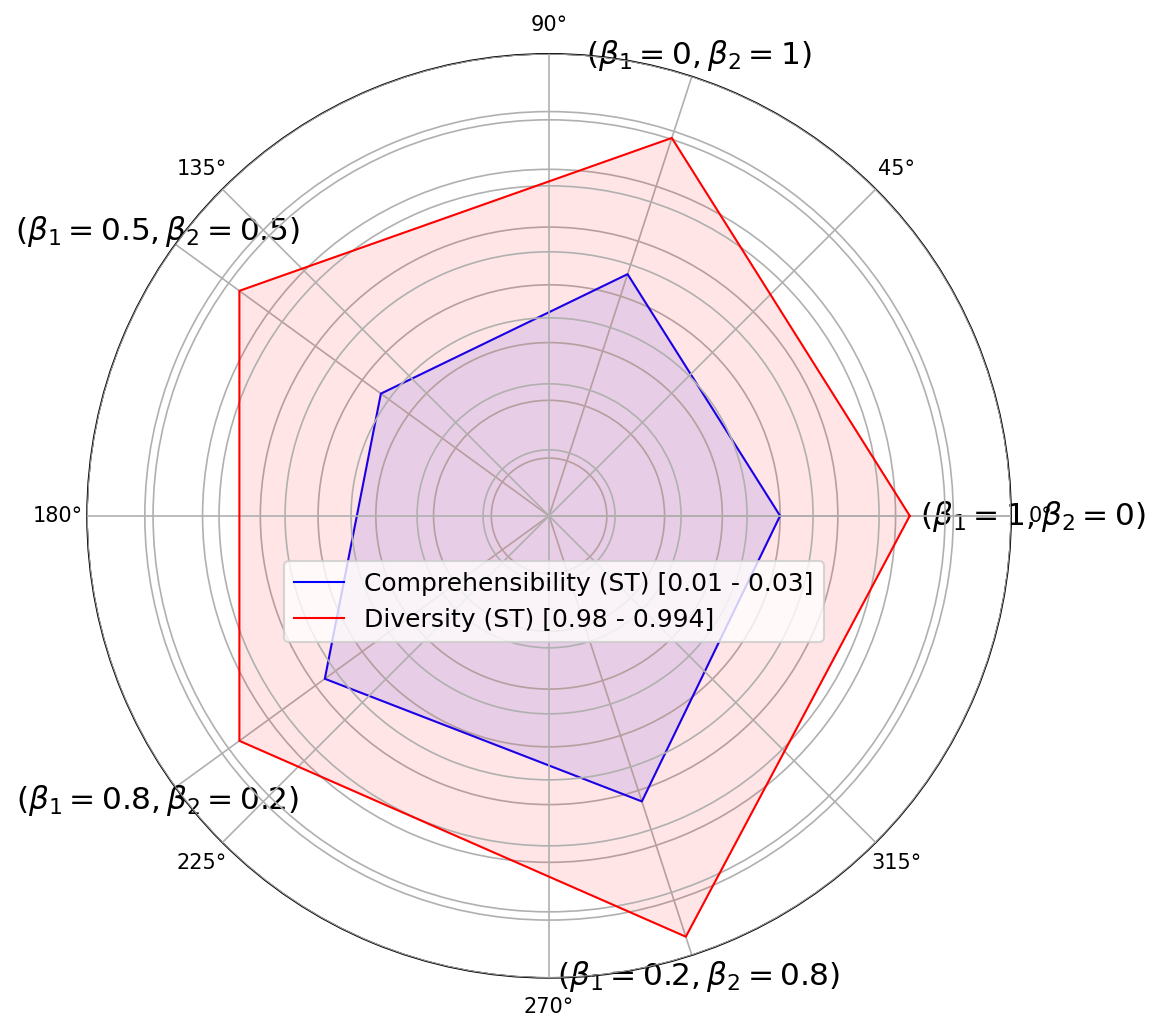}
        \caption{User-group}
    \end{subfigure}
    \caption{Comprehensibility and diversity for various $\beta_1$ and $\beta_2$ combinations}
    \label{fig:radar_chart}
\end{figure}

\textit{\textbf{Recency Effect on Summaries.}}
\label{sec:recency}
In this experiment, we investigate the impact of recency on summary explanations by adjusting the values of parameters \( \beta_1 \) and \( \beta_2 \), which control the balance between rating and recency in the recommendation process. We evaluate the ST algorithm on both user-centric and user-group scenarios with \( k = 10 \) and measure the comprehensibility and diversity of the generated summaries using as input the PGPR paths. Figure \ref{fig:radar_chart} illustrates the results across five combinations of \( \beta_1 \) and \( \beta_2 \). The blue region represents comprehensibility, while the red region represents diversity, with each axis corresponding to a different combination of \( \beta_1 \) and \( \beta_2 \). The highest diversity is achieved when the recency weight is large, whereas the highest comprehensibility occurs when the rating weight is dominant. This can be explained by the fact that emphasizing ratings tends to favor popular items, leading to fewer unique items in the paths and producing smaller, more concise summaries. In contrast, when recency is prioritized, newer and less common items are included in the paths, resulting in greater diversity.

\begin{figure}[!t]
    \centering
    \begin{subfigure}{0.24\textwidth}
        \centering
        \includegraphics[width=\textwidth]{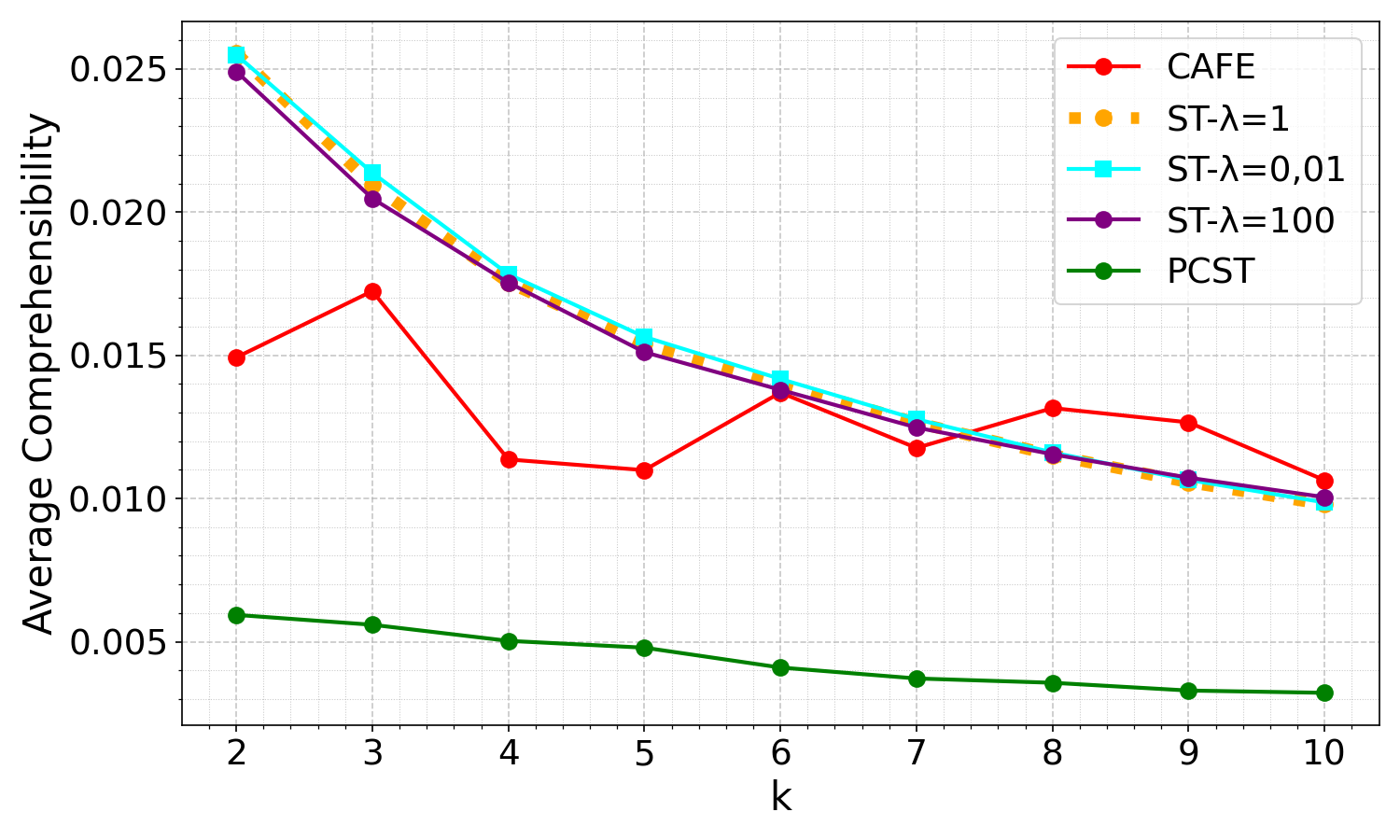}
        \caption{Popular items} 
        \label{fig:cmpr_cafe_item_group_pop}
    \end{subfigure}
    \hfill
        \centering
    \begin{subfigure}{0.24\textwidth}
        \centering
        \includegraphics[width=\textwidth]{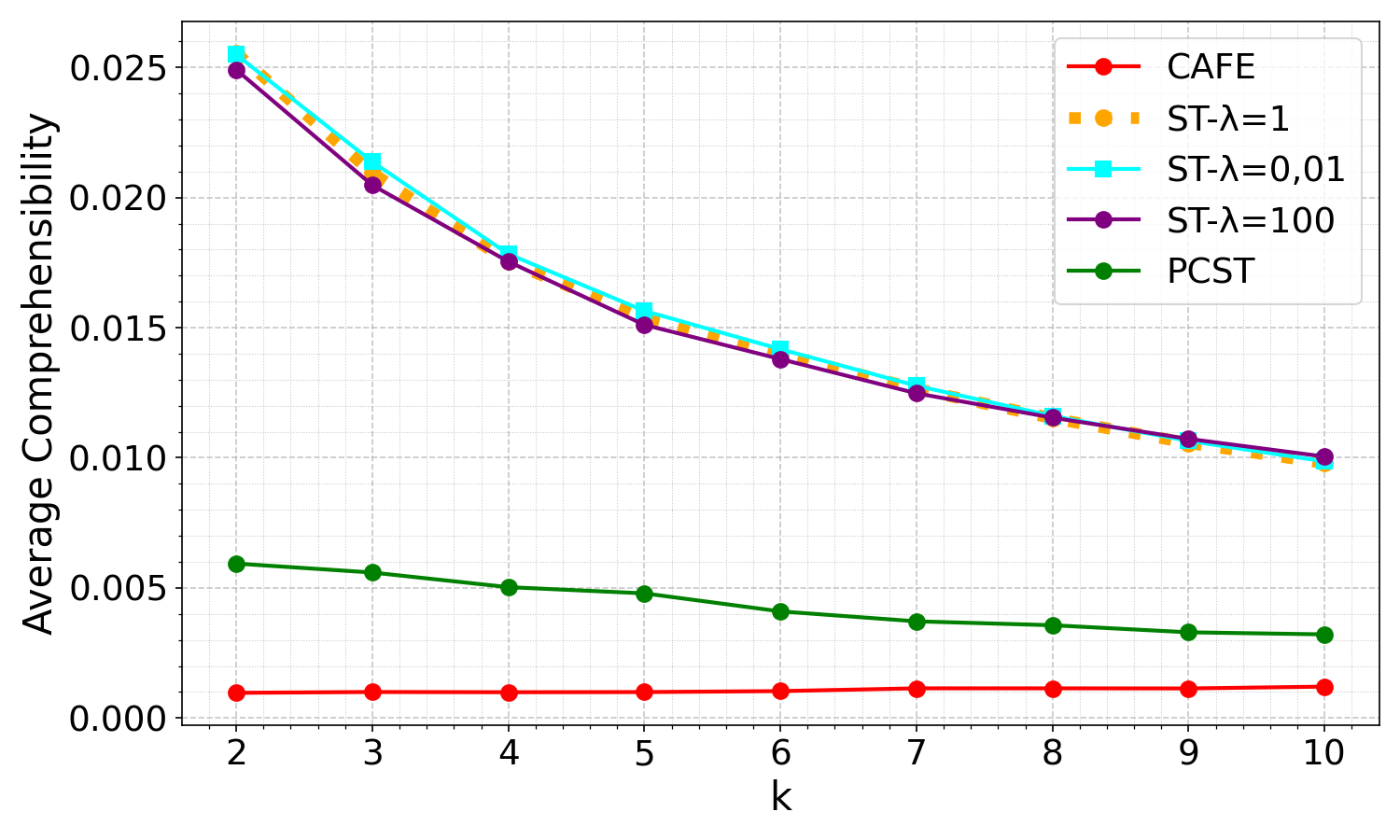}
        \caption{Unpopular items} 
        \label{fig:cmpr_cafe_item_group_unpop}
    \end{subfigure}
    \hfill
    \caption{Comprehensibility for popular and unpopular Items}
    \label{fig:fair}
\end{figure}

\textit{\textbf{Popularity bias.}}
A preliminary explanation fairness experiment indicated that the comprehensibility of less popular items was significantly worse in both baselines when compared to the popular items. However, our summarization methods did not exhibit this bias, suggesting a more fair explanation approach. This observation is shown in the comprehensibility metrics for popular and unpopular items shown in Figure \ref{fig:fair} for CAFE.

\section{User Study}

We conducted a user study with 30 participants, all undergraduate and graduate students engaged in machine learning-related projects, to assess two types of explanations used in recommendation systems: baseline path-based explanations (PGPR and CAFE) and summarized subgraph explanations (corresponding ST summaries).

The study was divided into two main parts:

\subsubsection{Comparison of Explanations (original Vs summarized)}
Participants were shown five pairs of explanations and the set of recommended items they refer to, with each pair containing one original path-based and one summarized subgraph explanation. The order was randomized to prevent bias. For each pair, participants answered the question: \textit{``Which explanation do you find more useful for decision-making?''}. Here is an example pair of explanations provided to participants:
\begin{itemize}
    \item \emph{Original Path-Based}: ``u94 watched item 612 related to external 81 related to item 2405, u94 watched item 1274 related to external u3824 related to item 1329, u94 watched item 1274 related to external u698 related to item 2069, u94 watched item 993 related to external u2318 related to item 2235, u94 watched item 1274 related to external u2270 related to item 1249, u94 watched item 682 related to external u1647 related to item 2215, u94 watched item 1274 related to external u2400 related to item 2386, u94 watched item 1274 related to external u2772 related to item 2261, u94 watched item 2600 related to external u2156 related to item 1694, u94 watched item 2016 related to external u602 related to item 2371''
    \item \emph{Summarized Subgraph}: ``u94 connects to 2215 via u2772, u8, u18, u7; connects to 2261, 1329, and 2069 via u2772; connects to 2371 via u8, 2386 via u18, and 2405 via u7; connects to 2235 via u2156, and through u2156 connects to 1694, 682, 1249, and 2069; u94 is also directly connected to 682.''
\end{itemize}

Results showed that 78.67\% of participants preferred the summarized subgraph explanations, suggesting that these summaries provided clearer insights into recommendation rationales.

\subsubsection{Usefulness Rating of Explanation Metrics}
After selecting their preferred explanations, participants rated the general usefulness of seven specific metrics for evaluating recommendation explanations, based on descriptions provided for each metric. Each metric was rated on a 1-5 scale, from 1 (Not Useful at All) to 5 (Extremely Useful). The metrics, descriptions, and average ratings were as follows:

\begin{itemize}
    \item Comprehensibility: Quantifies how easily users can understand explanation paths, with higher scores corresponding to briefer and more understandable explanations, and is inversely proportional to the length of the explanation paths or the size of the explanation subgraph.
        \begin{quote}
            \emph{Average Rating}: 4.52
        \end{quote}
    \item Actionability: Quantifies how useful explanation paths are in terms of actionable items, with a higher score indicating that users have more control over the recommendations; it is based on the proportion of actionable item nodes relative to the total number of nodes in the explanation path or subgraph, while non-actionable nodes include user nodes and external knowledge nodes.
        \begin{quote}
            \emph{Average Rating}: 3.79
        \end{quote}
    \item Diversity: Quantifies the variety among explanation paths provided by a recommendation system, with a higher score indicating that users receive a broader range of explanations; it measures the uniqueness of items or nodes among the edges of the explanation paths, and for summary subgraphs, diversity is assessed by comparing the variety of unique nodes across all pairs of edges using Jaccard similarity.
        \begin{quote}
            \emph{Average Rating}: 4.45
        \end{quote}
    \item Redundancy: Quantifies the extent of duplicate nodes in a set of explanation paths, with a higher score indicating more duplication, which can reduce the efficiency and informativeness of explanations; in summary subgraphs, redundancy measures the proportion of duplicated nodes within the subgraph.
        \begin{quote}
            \emph{Average Rating}: 4.14
        \end{quote}
    \item Consistency: Quantifies the stability of explanation paths across different values of k (number of top recommendations), with a higher score indicating more stable and reliable explanations; in summary explanations, consistency measures the similarity of consecutive summary subgraphs as k changes.
        \begin{quote}
            \emph{Average Rating}: 3.72
        \end{quote}
    \item Relevance: Quantifies how closely the explanation paths align with a user's preferences, based on the original edge weights (historical user-item interactions) in the knowledge graph; in summary explanations, relevance is measured as the total weight of the edges in the summary subgraph S, indicating the strength of connections based on interactions from historical user-item interactions.
        \begin{quote}
            \emph{Average Rating}: 4.38
        \end{quote}
    \item Privacy: Quantifies the extent to which user-specific information is exposed in explanation paths, with a higher score indicating better privacy protection by limiting the inclusion of user nodes; in summary explanations, privacy is measured by the proportion of user nodes within the subgraph.
        \begin{quote}
            \emph{Average Rating}: 3.69
        \end{quote}
\end{itemize}

Overall, participants favored the summarized subgraph explanations, with 78.67\% of participants selecting them over the original path-based explanations. Ratings on individual metrics indicate a strong preference for Comprehensibility (4.52) and Diversity (4.45), as these qualities made explanations more helpful for decision-making. Additionally, Relevance (4.38) and Redundancy (4.14) scored well, showing that participants valued explanations that were aligned with user interests and efficiently presented.

\section{Conclusion and future work}
\label{conclusion}
In this paper, we introduce summary explanations and provide efficient algorithms based on the Steiner Tree and Prize-Collecting Steiner Tree for graph-based recommenders. To the best of our knowledge, this is the first work to summarize path-based recommendation explanations for individual users, individual items, groups of users, and groups of items. Our results show that these summary explanations outperform baseline methods across multiple evaluation metrics.
Future work will explore explanation summaries to assess explanation fairness across user demographic and item category groups, as well as various refinements of our algorithms, including testing additional PCST prize assignment policies and considering incorporating node centrality measures. We also plan to study summaries to non-graph-based recommenders.

\section*{Acknowledgment}

This work has been partially supported by project MIS 5154714 of the National Recovery and Resilience Plan Greece 2.0 funded by the European Union under the NextGenerationEU Program.

\clearpage

\bibliographystyle{IEEEtran}
\bibliography{IEEEabrv,bibliography}

\begin{thebibliography}{10}
\providecommand{\url}[1]{#1}
\csname url@samestyle\endcsname
\providecommand{\newblock}{\relax}
\providecommand{\bibinfo}[2]{#2}
\providecommand{\BIBentrySTDinterwordspacing}{\spaceskip=0pt\relax}
\providecommand{\BIBentryALTinterwordstretchfactor}{4}
\providecommand{\BIBentryALTinterwordspacing}{\spaceskip=\fontdimen2\font plus
\BIBentryALTinterwordstretchfactor\fontdimen3\font minus \fontdimen4\font\relax}
\providecommand{\BIBforeignlanguage}[2]{{%
\expandafter\ifx\csname l@#1\endcsname\relax
\typeout{** WARNING: IEEEtran.bst: No hyphenation pattern has been}%
\typeout{** loaded for the language `#1'. Using the pattern for}%
\typeout{** the default language instead.}%
\else
\language=\csname l@#1\endcsname
\fi
#2}}
\providecommand{\BIBdecl}{\relax}
\BIBdecl

\bibitem{10.1145/3514221.3522564}
\BIBentryALTinterwordspacing
R.~Pradhan, A.~Lahiri, S.~Galhotra, and B.~Salimi, ``Explainable ai: Foundations, applications, opportunities for data management research,'' in \emph{Proceedings of the 2022 International Conference on Management of Data}, ser. SIGMOD '22.\hskip 1em plus 0.5em minus 0.4em\relax New York, NY, USA: Association for Computing Machinery, 2022, p. 2452–2457. [Online]. Available: \url{https://doi.org/10.1145/3514221.3522564}
\BIBentrySTDinterwordspacing

\bibitem{electronics8080832}
\BIBentryALTinterwordspacing
D.~V. Carvalho, E.~M. Pereira, and J.~S. Cardoso, ``Machine learning interpretability: A survey on methods and metrics,'' \emph{Electronics}, vol.~8, no.~8, 2019. [Online]. Available: \url{https://www.mdpi.com/2079-9292/8/8/832}
\BIBentrySTDinterwordspacing

\bibitem{zhang2020explainable}
Y.~Zhang, X.~Chen \emph{et~al.}, ``Explainable recommendation: A survey and new perspectives,'' \emph{Foundations and Trends{\textregistered} in Information Retrieval}, vol.~14, no.~1, pp. 1--101, 2020.

\bibitem{tintarev2012beyond}
N.~Tintarev and J.~Masthoff, ``Beyond explaining single item recommendations,'' in \emph{Recommender Systems Handbook}.\hskip 1em plus 0.5em minus 0.4em\relax Springer, 2012, pp. 711--756.

\bibitem{ricci2021recommender}
F.~Ricci, L.~Rokach, and B.~Shapira, ``Recommender systems: Techniques, applications, and challenges,'' \emph{Recommender systems handbook}, pp. 1--35, 2021.

\bibitem{deldjoo2024fairness}
Y.~Deldjoo, D.~Jannach, A.~Bellogin, A.~Difonzo, and D.~Zanzonelli, ``Fairness in recommender systems: research landscape and future directions,'' \emph{User Modeling and User-Adapted Interaction}, vol.~34, no.~1, pp. 59--108, 2024.

\bibitem{ghazimatin2020prince}
A.~Ghazimatin, O.~Balalau, R.~Saha~Roy, and G.~Weikum, ``Prince: Provider-side interpretability with counterfactual explanations in recommender systems,'' in \emph{Proceedings of the 13th International Conference on Web Search and Data Mining}, 2020, pp. 196--204.

\bibitem{wu2022graph}
S.~Wu, F.~Sun, W.~Zhang, X.~Xie, and B.~Cui, ``Graph neural networks in recommender systems: a survey,'' \emph{ACM Computing Surveys}, vol.~55, no.~5, pp. 1--37, 2022.

\bibitem{wang2020recommendation}
H.~Wang, Z.~Le, and X.~Gong, ``Recommendation system based on heterogeneous feature: A survey,'' \emph{IEEE Access}, vol.~8, pp. 170\,779--170\,793, 2020.

\bibitem{sharma2016graphjet}
A.~Sharma, J.~Jiang, P.~Bommannavar, B.~Larson, and J.~Lin, ``Graphjet: Real-time content recommendations at twitter,'' \emph{Proceedings of the VLDB Endowment}, vol.~9, no.~13, pp. 1281--1292, 2016.

\bibitem{de2024personalized}
M.~De~Nadai, F.~Fabbri, P.~Gigioli, A.~Wang, A.~Li, F.~Silvestri, L.~Kim, S.~Lin, V.~Radosavljevic, S.~Ghael \emph{et~al.}, ``Personalized audiobook recommendations at spotify through graph neural networks,'' in \emph{Companion Proceedings of the ACM on Web Conference 2024}, 2024, pp. 403--412.

\bibitem{amazon}
S.~Virinchi, A.~Saladi, and A.~Mondal, ``Recommending related products using graph neural networks in directed graphs,'' in \emph{Machine Learning and Knowledge Discovery in Databases}, M.-R. Amini, S.~Canu, A.~Fischer, T.~Guns, P.~Kralj~Novak, and G.~Tsoumakas, Eds.\hskip 1em plus 0.5em minus 0.4em\relax Cham: Springer International Publishing, 2023, pp. 541--557.

\bibitem{ugander2011anatomy}
J.~Ugander, B.~Karrer, L.~Backstrom, and C.~Marlow, ``The anatomy of the facebook social graph,'' \emph{arXiv preprint arXiv:1111.4503}, 2011.

\bibitem{ying2018graph}
R.~Ying, R.~He, K.~Chen, P.~Eksombatchai, W.~L. Hamilton, and J.~Leskovec, ``Graph convolutional neural networks for web-scale recommender systems,'' in \emph{Proceedings of the 24th ACM SIGKDD international conference on knowledge discovery \& data mining}, 2018, pp. 974--983.

\bibitem{geng2022path}
S.~Geng, Z.~Fu, J.~Tan, Y.~Ge, G.~De~Melo, and Y.~Zhang, ``Path language modeling over knowledge graphsfor explainable recommendation,'' in \emph{Proceedings of the ACM Web Conference 2022}, 2022, pp. 946--955.

\bibitem{xian2019reinforcement}
Y.~Xian, Z.~Fu, S.~Muthukrishnan, G.~De~Melo, and Y.~Zhang, ``Reinforcement knowledge graph reasoning for explainable recommendation,'' in \emph{Proceedings of the 42nd international ACM SIGIR conference on research and development in information retrieval}, 2019, pp. 285--294.

\bibitem{xian2020cafe}
Y.~Xian, Z.~Fu, H.~Zhao, Y.~Ge, X.~Chen, Q.~Huang, S.~Geng, Z.~Qin, G.~De~Melo, S.~Muthukrishnan \emph{et~al.}, ``Cafe: Coarse-to-fine neural symbolic reasoning for explainable recommendation,'' in \emph{Proceedings of the 29th ACM International Conference on Information \& Knowledge Management}, 2020, pp. 1645--1654.

\bibitem{guo2020survey}
Q.~Guo, F.~Zhuang, C.~Qin, H.~Zhu, X.~Xie, H.~Xiong, and Q.~He, ``A survey on knowledge graph-based recommender systems,'' \emph{IEEE Transactions on Knowledge and Data Engineering}, vol.~34, no.~8, pp. 3549--3568, 2020.

\bibitem{gao2023survey}
C.~Gao, Y.~Zheng, N.~Li, Y.~Li, Y.~Qin, J.~Piao, Y.~Quan, J.~Chang, D.~Jin, X.~He \emph{et~al.}, ``A survey of graph neural networks for recommender systems: Challenges, methods, and directions,'' \emph{ACM Transactions on Recommender Systems}, vol.~1, no.~1, pp. 1--51, 2023.

\bibitem{plm}
\BIBentryALTinterwordspacing
S.~Geng, Z.~Fu, J.~Tan, Y.~Ge, G.~de~Melo, and Y.~Zhang, ``Path language modeling over knowledge graphs for explainable recommendation,'' in \emph{Proceedings of the ACM Web Conference 2022}, ser. WWW '22.\hskip 1em plus 0.5em minus 0.4em\relax New York, NY, USA: Association for Computing Machinery, 2022, p. 946–955. [Online]. Available: \url{https://doi.org/10.1145/3485447.3511937}
\BIBentrySTDinterwordspacing

\bibitem{balloccu2023faithful}
G.~Balloccu, L.~Boratto, C.~Cancedda, G.~Fenu, and M.~Marras, ``Faithful path language modelling for explainable recommendation over knowledge graph,'' \emph{arXiv preprint arXiv:2310.16452}, 2023.

\bibitem{MovieLensData}
\BIBentryALTinterwordspacing
F.~M. Harper and J.~A. Konstan, ``The movielens datasets: History and context,'' \emph{ACM Trans. Interact. Intell. Syst.}, vol.~5, no.~4, dec 2015. [Online]. Available: \url{https://doi.org/10.1145/2827872}
\BIBentrySTDinterwordspacing

\bibitem{lfm1b}
\BIBentryALTinterwordspacing
M.~Schedl, ``The lfm-1b dataset for music retrieval and recommendation,'' in \emph{Proceedings of the 2016 ACM on International Conference on Multimedia Retrieval}, ser. ICMR '16.\hskip 1em plus 0.5em minus 0.4em\relax New York, NY, USA: Association for Computing Machinery, 2016, p. 103–110. [Online]. Available: \url{https://doi.org/10.1145/2911996.2912004}
\BIBentrySTDinterwordspacing

\bibitem{tao2019}
\BIBentryALTinterwordspacing
Y.~Tao, Y.~Jia, N.~Wang, and H.~Wang, ``The fact: Taming latent factor models for explainability with factorization trees,'' in \emph{Proceedings of the 42nd International ACM SIGIR Conference on Research and Development in Information Retrieval}, ser. SIGIR'19.\hskip 1em plus 0.5em minus 0.4em\relax New York, NY, USA: Association for Computing Machinery, 2019, p. 295–304. [Online]. Available: \url{https://doi.org/10.1145/3331184.3331244}
\BIBentrySTDinterwordspacing

\bibitem{wu2019}
\BIBentryALTinterwordspacing
L.~Wu, C.~Quan, C.~Li, Q.~Wang, B.~Zheng, and X.~Luo, ``A context-aware user-item representation learning for item recommendation,'' \emph{ACM Trans. Inf. Syst.}, vol.~37, no.~2, jan 2019. [Online]. Available: \url{https://doi.org/10.1145/3298988}
\BIBentrySTDinterwordspacing

\bibitem{talrec23}
\BIBentryALTinterwordspacing
K.~Bao, J.~Zhang, Y.~Zhang, W.~Wang, F.~Feng, and X.~He, ``Tallrec: An effective and efficient tuning framework to align large language model with recommendation,'' in \emph{Proceedings of the 17th ACM Conference on Recommender Systems}, ser. RecSys '23.\hskip 1em plus 0.5em minus 0.4em\relax New York, NY, USA: Association for Computing Machinery, 2023, p. 1007–1014. [Online]. Available: \url{https://doi.org/10.1145/3604915.3608857}
\BIBentrySTDinterwordspacing

\bibitem{gpt23}
\BIBentryALTinterwordspacing
S.~Dai, N.~Shao, H.~Zhao, W.~Yu, Z.~Si, C.~Xu, Z.~Sun, X.~Zhang, and J.~Xu, ``Uncovering chatgpt’s capabilities in recommender systems,'' in \emph{Proceedings of the 17th ACM Conference on Recommender Systems}, ser. RecSys '23.\hskip 1em plus 0.5em minus 0.4em\relax New York, NY, USA: Association for Computing Machinery, 2023, p. 1126–1132. [Online]. Available: \url{https://doi.org/10.1145/3604915.3610646}
\BIBentrySTDinterwordspacing

\bibitem{wang2023sequential}
Z.~Wang, X.~Chen, R.~Zhou, Q.~Dai, Z.~Dong, and J.-R. Wen, ``Sequential recommendation with user causal behavior discovery,'' in \emph{2023 IEEE 39th International Conference on Data Engineering (ICDE)}.\hskip 1em plus 0.5em minus 0.4em\relax IEEE, 2023, pp. 28--40.

\bibitem{wang2022tower}
S.~Wang, H.~Li, C.~C. Cao, X.-H. Li, N.~N. Fai, J.~Liu, X.~Xue, H.~Song, J.~Li, G.~Gu \emph{et~al.}, ``Tower bridge net (tb-net): Bidirectional knowledge graph aware embedding propagation for explainable recommender systems,'' in \emph{2022 IEEE 38th International Conference on Data Engineering (ICDE)}.\hskip 1em plus 0.5em minus 0.4em\relax IEEE, 2022, pp. 3268--3279.

\bibitem{bodria2023benchmarking}
F.~Bodria, F.~Giannotti, R.~Guidotti, F.~Naretto, D.~Pedreschi, and S.~Rinzivillo, ``Benchmarking and survey of explanation methods for black box models,'' \emph{Data Mining and Knowledge Discovery}, pp. 1--60, 2023.

\bibitem{adadi2018peeking}
A.~Adadi and M.~Berrada, ``Peeking inside the black-box: a survey on explainable artificial intelligence (xai),'' \emph{IEEE access}, vol.~6, pp. 52\,138--52\,160, 2018.

\bibitem{wu2023generic}
H.~Wu, H.~Fang, Z.~Sun, C.~Geng, X.~Kong, and Y.-S. Ong, ``A generic reinforced explainable framework with knowledge graph for session-based recommendation,'' in \emph{2023 IEEE 39th International Conference on Data Engineering (ICDE)}.\hskip 1em plus 0.5em minus 0.4em\relax IEEE, 2023, pp. 1260--1272.

\bibitem{wang2018ripplenet}
H.~Wang, F.~Zhang, J.~Wang, M.~Zhao, W.~Li, X.~Xie, and M.~Guo, ``Ripplenet: Propagating user preferences on the knowledge graph for recommender systems,'' in \emph{Proceedings of the 27th ACM international conference on information and knowledge management}, 2018, pp. 417--426.

\bibitem{10.1109/TKDE.2018.2833443}
\BIBentryALTinterwordspacing
C.~Shi, B.~Hu, W.~X. Zhao, and P.~S. Yu, ``Heterogeneous information network embedding for recommendation,'' \emph{IEEE Trans. on Knowl. and Data Eng.}, vol.~31, no.~2, p. 357–370, feb 2019. [Online]. Available: \url{https://doi.org/10.1109/TKDE.2018.2833443}
\BIBentrySTDinterwordspacing

\bibitem{10.1145/3219819.3219965}
\BIBentryALTinterwordspacing
B.~Hu, C.~Shi, W.~X. Zhao, and P.~S. Yu, ``Leveraging meta-path based context for top- n recommendation with a neural co-attention model,'' in \emph{Proceedings of the 24th ACM SIGKDD International Conference on Knowledge Discovery \& Data Mining}, ser. KDD '18.\hskip 1em plus 0.5em minus 0.4em\relax New York, NY, USA: Association for Computing Machinery, 2018, p. 1531–1540. [Online]. Available: \url{https://doi.org/10.1145/3219819.3219965}
\BIBentrySTDinterwordspacing

\bibitem{10.1609/aaai.v33i01.33015329}
\BIBentryALTinterwordspacing
X.~Wang, D.~Wang, C.~Xu, X.~He, Y.~Cao, and T.-S. Chua, ``Explainable reasoning over knowledge graphs for recommendation,'' in \emph{Proceedings of the Thirty-Third AAAI Conference on Artificial Intelligence and Thirty-First Innovative Applications of Artificial Intelligence Conference and Ninth AAAI Symposium on Educational Advances in Artificial Intelligence}, ser. AAAI'19/IAAI'19/EAAI'19.\hskip 1em plus 0.5em minus 0.4em\relax AAAI Press, 2019. [Online]. Available: \url{https://doi.org/10.1609/aaai.v33i01.33015329}
\BIBentrySTDinterwordspacing

\bibitem{seq2021}
\BIBentryALTinterwordspacing
X.~Dong, B.~Jin, W.~Zhuo, B.~Li, and T.~Xue, ``Improving sequential recommendation with attribute-augmented graph neural networks,'' in \emph{Advances in Knowledge Discovery and Data Mining: 25th Pacific-Asia Conference, PAKDD 2021, Virtual Event, May 11–14, 2021, Proceedings, Part II}.\hskip 1em plus 0.5em minus 0.4em\relax Berlin, Heidelberg: Springer-Verlag, 2021, p. 373–385. [Online]. Available: \url{https://doi.org/10.1007/978-3-030-75765-6_30}
\BIBentrySTDinterwordspacing

\bibitem{balloccu2022recency}
G.~Balloccu, L.~Boratto, G.~Fenu, M.~Marras \emph{et~al.}, ``Recency, popularity, and diversity of explanations in knowledge-based recommendation,'' in \emph{CEUR WORKSHOP PROCEEDINGS}, vol. 3177.\hskip 1em plus 0.5em minus 0.4em\relax CEUR-WS, 2022.

\bibitem{liu2018graph}
Y.~Liu, T.~Safavi, A.~Dighe, and D.~Koutra, ``Graph summarization methods and applications: A survey,'' \emph{ACM computing surveys (CSUR)}, vol.~51, no.~3, pp. 1--34, 2018.

\bibitem{lefevre2010grass}
K.~LeFevre and E.~Terzi, ``Grass: Graph structure summarization,'' in \emph{Proceedings of the 2010 SIAM International Conference on Data Mining}.\hskip 1em plus 0.5em minus 0.4em\relax SIAM, 2010, pp. 454--465.

\bibitem{toivonen2011compression}
H.~Toivonen, F.~Zhou, A.~Hartikainen, and A.~Hinkka, ``Compression of weighted graphs,'' in \emph{Proceedings of the 17th ACM SIGKDD international conference on Knowledge discovery and data mining}, 2011, pp. 965--973.

\bibitem{maccioni2016scalable}
A.~Maccioni and D.~J. Abadi, ``Scalable pattern matching over compressed graphs via dedensification,'' in \emph{Proceedings of the 22nd ACM SIGKDD International Conference on Knowledge Discovery and Data Mining}, 2016, pp. 1755--1764.

\bibitem{shen2006visual}
Z.~Shen, K.-L. Ma, and T.~Eliassi-Rad, ``Visual analysis of large heterogeneous social networks by semantic and structural abstraction,'' \emph{IEEE transactions on visualization and computer graphics}, vol.~12, no.~6, pp. 1427--1439, 2006.

\bibitem{mathioudakis2011sparsification}
M.~Mathioudakis, F.~Bonchi, C.~Castillo, A.~Gionis, and A.~Ukkonen, ``Sparsification of influence networks,'' in \emph{Proceedings of the 17th ACM SIGKDD international conference on Knowledge discovery and data mining}, 2011, pp. 529--537.

\bibitem{steiner_sum}
A.~Pappas, G.~Troullinou, G.~Roussakis, H.~Kondylakis, and D.~Plexousakis, ``Exploring importance measures for summarizing rdf/s kbs,'' in \emph{The Semantic Web}, E.~Blomqvist, D.~Maynard, A.~Gangemi, R.~Hoekstra, P.~Hitzler, and O.~Hartig, Eds.\hskip 1em plus 0.5em minus 0.4em\relax Cham: Springer International Publishing, 2017, pp. 387--403.

\bibitem{group_recs}
\BIBentryALTinterwordspacing
S.~Amer-Yahia, S.~B. Roy, A.~Chawlat, G.~Das, and C.~Yu, ``Group recommendation: semantics and efficiency,'' \emph{Proc. VLDB Endow.}, vol.~2, no.~1, p. 754–765, aug 2009. [Online]. Available: \url{https://doi.org/10.14778/1687627.1687713}
\BIBentrySTDinterwordspacing

\bibitem{fair_recs_18}
M.~Stratigi, H.~Kondylakis, and K.~Stefanidis, ``Fairgrecs: Fair group recommendations by exploiting personal health information,'' in \emph{Database and Expert Systems Applications}, S.~Hartmann, H.~Ma, A.~Hameurlain, G.~Pernul, and R.~R. Wagner, Eds.\hskip 1em plus 0.5em minus 0.4em\relax Cham: Springer International Publishing, 2018, pp. 147--155.

\bibitem{stratigi2017fairness}
------, ``Fairness in group recommendations in the health domain,'' in \emph{2017 IEEE 33rd International Conference on Data Engineering (ICDE)}.\hskip 1em plus 0.5em minus 0.4em\relax IEEE, 2017, pp. 1481--1488.

\bibitem{guo2020group}
L.~Guo, H.~Yin, Q.~Wang, B.~Cui, Z.~Huang, and L.~Cui, ``Group recommendation with latent voting mechanism,'' in \emph{2020 IEEE 36th International Conference on Data Engineering (ICDE)}.\hskip 1em plus 0.5em minus 0.4em\relax IEEE, 2020, pp. 121--132.

\bibitem{deng2021knowledge}
Z.~Deng, C.~Li, S.~Liu, W.~Ali, and J.~Shao, ``Knowledge-aware group representation learning for group recommendation,'' in \emph{2021 IEEE 37th International Conference on Data Engineering (ICDE)}.\hskip 1em plus 0.5em minus 0.4em\relax IEEE, 2021, pp. 1571--1582.

\bibitem{balloccu2023reinforcement}
G.~Balloccu, L.~Boratto, G.~Fenu, and M.~Marras, ``Reinforcement recommendation reasoning through knowledge graphs for explanation path quality,'' \emph{Knowledge-Based Systems}, vol. 260, p. 110098, 2023.

\bibitem{Zhou2021EvaluatingTQ}
\BIBentryALTinterwordspacing
J.~Zhou, A.~H. Gandomi, F.~Chen, and A.~Holzinger, ``Evaluating the quality of machine learning explanations: A survey on methods and metrics,'' \emph{Electronics}, 2021. [Online]. Available: \url{https://api.semanticscholar.org/CorpusID:233834400}
\BIBentrySTDinterwordspacing

\bibitem{zelikovsky1993faster}
A.~Z. Zelikovsky, ``A faster approximation algorithm for the steiner tree problem in graphs,'' \emph{Information Processing Letters}, vol.~46, no.~2, pp. 79--83, 1993.

\bibitem{goemans1995general}
M.~X. Goemans and D.~P. Williamson, ``A general approximation technique for constrained forest problems,'' \emph{SIAM Journal on Computing}, vol.~24, no.~2, pp. 296--317, 1995.

\bibitem{cao2018unifying}
Y.~Cao, X.~Wang, X.~He, Z.~Hu, and C.~Tat-seng, ``Unifying knowledge graph learning and recommendation: Towards a better understanding of user preference,'' in \emph{WWW}, 2019.

\bibitem{Zhao-DI-2019}
\BIBentryALTinterwordspacing
W.~X. Zhao, G.~He, K.~Yang, H.~Dou, J.~Huang, S.~Ouyang, and J.~Wen, ``Kb4rec: A data set for linking knowledge bases with recommender systems,'' \emph{Data Intelligence}, vol.~1, no.~2, pp. 121--136, 2019. [Online]. Available: \url{https://doi.org/10.1162/dint_a_00008}
\BIBentrySTDinterwordspacing

\bibitem{10.1007/978-3-031-20319-0_30}
L.~Coroama and A.~Groza, ``Evaluation metrics in explainable artificial intelligence (xai),'' in \emph{Advanced Research in Technologies, Information, Innovation and Sustainability}, T.~Guarda, F.~Portela, and M.~F. Augusto, Eds.\hskip 1em plus 0.5em minus 0.4em\relax Cham: Springer Nature Switzerland, 2022, pp. 401--413.

\bibitem{zytek2024llmsxaifuturedirections}
\BIBentryALTinterwordspacing
A.~Zytek, S.~Pidò, and K.~Veeramachaneni, ``Llms for xai: Future directions for explaining explanations,'' 2024. [Online]. Available: \url{https://arxiv.org/abs/2405.06064}
\BIBentrySTDinterwordspacing

\bibitem{rosenfeld2021better}
A.~Rosenfeld, ``Better metrics for evaluating explainable artificial intelligence,'' in \emph{Proceedings of the 20th international conference on autonomous agents and multiagent systems}, 2021, pp. 45--50.

\bibitem{10.1007/978-3-031-40878-6_12}
J.~Hulstijn, I.~Tchappi, A.~Najjar, and R.~Aydo{\u{g}}an, ``Metrics for evaluating explainable recommender systems,'' in \emph{Explainable and Transparent AI and Multi-Agent Systems}, D.~Calvaresi, A.~Najjar, A.~Omicini, R.~Aydogan, R.~Carli, G.~Ciatto, Y.~Mualla, and K.~Fr{\"a}mling, Eds.\hskip 1em plus 0.5em minus 0.4em\relax Cham: Springer Nature Switzerland, 2023, pp. 212--230.

\bibitem{nguyen2020quantitative}
A.-p. Nguyen and M.~R. Mart{\'\i}nez, ``On quantitative aspects of model interpretability,'' \emph{arXiv preprint arXiv:2007.07584}, 2020.

\bibitem{yang2018towards}
F.~Yang, N.~Liu, S.~Wang, and X.~Hu, ``Towards interpretation of recommender systems with sorted explanation paths,'' in \emph{2018 IEEE International Conference on Data Mining (ICDM)}.\hskip 1em plus 0.5em minus 0.4em\relax IEEE, 2018, pp. 667--676.

\bibitem{markchom2023explainable}
T.~Markchom, H.~Liang, and J.~Ferryman, ``Explainable meta-path based recommender systems,'' \emph{ACM Transactions on Recommender Systems}, 2023.

\bibitem{balloccu2022post}
G.~Balloccu, L.~Boratto, G.~Fenu, and M.~Marras, ``Post processing recommender systems with knowledge graphs for recency, popularity, and diversity of explanations,'' in \emph{Proceedings of the 45th International ACM SIGIR Conference on Research and Development in Information Retrieval}, 2022, pp. 646--656.

\bibitem{10.1007/978-3-031-28241-6_1}
G.~Balloccu, L.~Boratto, C.~Cancedda, G.~Fenu, and M.~Marras, ``Knowledge is power, understanding is impact: Utility and beyond goals, explanation quality, and fairness in path reasoning recommendation,'' in \emph{Advances in Information Retrieval}, J.~Kamps, L.~Goeuriot, F.~Crestani, M.~Maistro, H.~Joho, B.~Davis, C.~Gurrin, U.~Kruschwitz, and A.~Caputo, Eds.\hskip 1em plus 0.5em minus 0.4em\relax Cham: Springer Nature Switzerland, 2023, pp. 3--19.

\end{thebibliography}

\end{document}